\PassOptionsToPackage{table}{xcolor}

\documentclass[single-column]{fairmeta}
\usepackage{amsmath}
\crefname{section}{Sec.}{Secs.}\Crefname{section}{Sec.}{Secs.}
\crefname{subsection}{Sec.}{Secs.}\Crefname{subsection}{Sec.}{Secs.}
\crefname{subsubsection}{Sec.}{Secs.}\Crefname{subsubsection}{Sec.}{Secs.}
\crefname{appendix}{App.}{Apps.}\Crefname{appendix}{App.}{Apps.}
\crefname{figure}{Fig.}{Figs.}\Crefname{figure}{Fig.}{Figs.}
\crefname{table}{Tab.}{Tabs.}\Crefname{table}{Tab.}{Tabs.}
\crefname{equation}{Eq.}{Eqs.}\Crefname{equation}{Eq.}{Eqs.}
\usepackage{amssymb}
\usepackage[disable]{todonotes}
\usepackage{listings}
\usepackage{makecell}
\usepackage{wrapfig}
\usepackage{float}
\usepackage{placeins}
\usepackage{needspace}
\usepackage{tabularx}
\usepackage{textcomp}
\usepackage{stfloats}
\usepackage{url}
\usepackage{verbatim}
\usepackage{tocloft}
\usepackage{adjustbox}
\usepackage{pifont}
\usepackage{tikz}
\usepackage{fontawesome5}
\usepackage{forest}
\definecolor{cmphead}{HTML}{EFEAFB}
\definecolor{cmpstripe}{HTML}{F6F6F8}
\providecommand{\paperlink}[1]{\href{#1}{\faFile[regular]}}
\providecommand{\codelink}[1]{\hspace{4pt}\href{#1}{\faGithub}}
\providecommand{\nocode}{\hspace{4pt}\makebox[0.9em][c]{-}}
\providecommand{\yes}{\textcolor{black}{\ding{51}}}
\providecommand{\no}{\textcolor{black}{\ding{55}}}

\usepackage{comment}
\usepackage{colortbl}
\usepackage{diagbox}
\usepackage{xltabular}
\usepackage{array}
\usepackage{threeparttable}
\usepackage{xspace}
\usepackage{enumitem}

\makeatletter
\DeclareRobustCommand\onedot{\futurelet\@let@token\@onedot}
\def\@onedot{\ifx\@let@token.\else.\null\fi\xspace}
\makeatother

\definecolor{wmgreen}{RGB}{46,125,50}
\definecolor{wmorange}{RGB}{230,145,56}
\definecolor{wmgray}{RGB}{120,120,120}

\graphicspath{{./}{local/}}

\definecolor{myblue}{RGB}{233, 241, 249}
\definecolor{mygray}{RGB}{99, 110, 114}
\definecolor{myred}{RGB}{255, 118, 117}
\definecolor{myyellow}{RGB}{255, 234, 167}
\definecolor{mygreen}{RGB}{216, 226, 204}
\definecolor{mypurple}{RGB}{162, 155, 254}
\definecolor{mybrown}{RGB}{215, 190, 154}
\definecolor{myorange}{RGB}{255, 220, 190}

\definecolor{preC}{HTML}{B5651D}
\definecolor{preBg}{HTML}{FFF7EE}
\definecolor{runC}{HTML}{1E62C0}
\definecolor{runBg}{HTML}{F1F6FE}

\newcommand{\etal}{et al.\@\xspace}
\titleformat*{\paragraph}{\bfseries}
\renewcommand{\emph}[1]{\textbf{#1}}
\setlist{nosep,leftmargin=1.3em,topsep=2pt}
\newcolumntype{Y}{>{\raggedright\arraybackslash}X}
\newtcolorbox{scopebox}[1]{colback=metabg,colframe=metabg,coltitle=metafg,fonttitle=\sffamily\bfseries,
  title=#1,boxrule=0pt,arc=4pt,left=6pt,right=6pt,top=4pt,bottom=4pt,fontupper=\small}
\newtcolorbox{takeawaybox}{colback=metabg,colframe=metabg,coltitle=metafg,fonttitle=\sffamily\bfseries\small,
  title=Takeaway,boxrule=0pt,arc=4pt,left=6pt,right=6pt,top=2pt,bottom=4pt,fontupper=\small,before skip=6pt,after skip=8pt}

\title{Self-Improving Embodied Agents:\\ Making the Leap from Virtual to Real Worlds}

\renewcommand\authorlist{%
  {\normalfont\small {\sffamily Execution Team:}~Chang~Dong$^{*}$, Jiatong~Xia$^{*}$, Wenxuan~Song$^{*}$, Lishan~Yang, Jiajun~Liu, King~Hang~Wong, Fuhao~Li, Yikai~Qin, Wenze~Wang, Mingjian~Liang, Junlin~Hao, Yue~Chen, Yutao~Ouyang, Wenbo~Zhang$^{*,\ddag}$\par
  {\sffamily Advisor Team:}~Huazhe~Xu, Chao~Yu, Hao~Dong, Flood~Sung, Xionghui~Chen, Haoang~Li, Shiyu~Huang, Feras~Dayoub, Minh~Hoai, Lingqiao~Liu, Mengdi~Xu$^{\dag}$\par}}

\patchcmd{\mymaketitle}{\begin{tcolorbox}}{\begin{tcolorbox}[overlay={\node[anchor=south east, xshift=-0.5cm, yshift=0.45cm, inner sep=0pt]
  at (frame.south east) {\includegraphics[width=1.1cm]{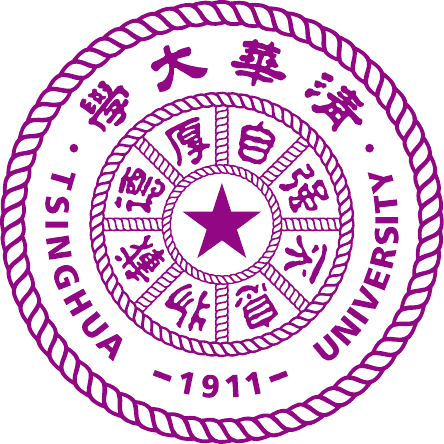}\hspace{0.3cm}\includegraphics[width=1.1cm]{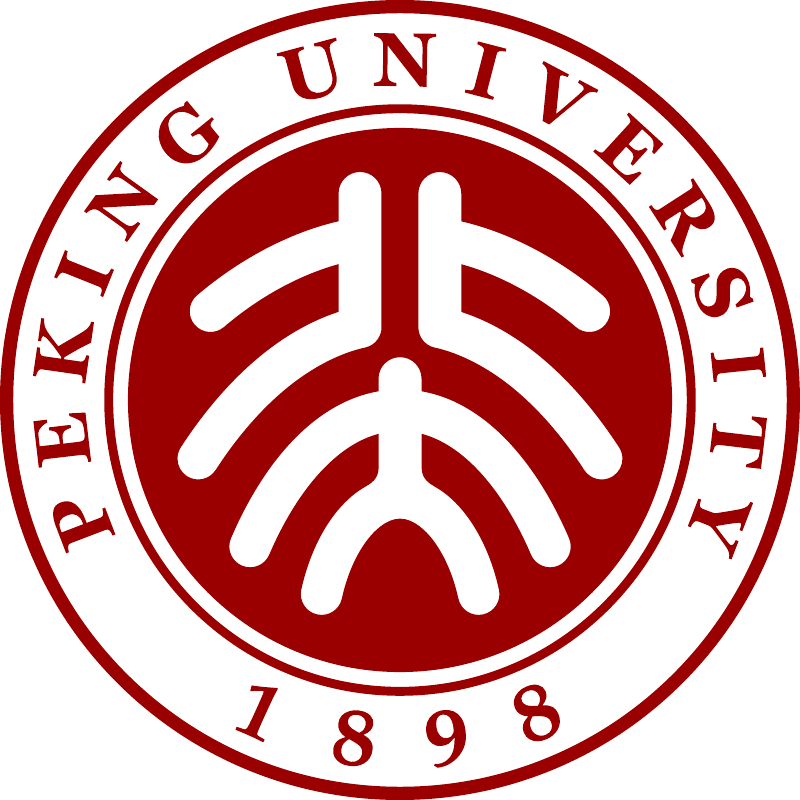}\hspace{0.3cm}\includegraphics[height=1.1cm]{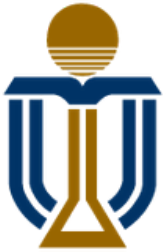}\hspace{0.3cm}\raisebox{0.15cm}{\includegraphics[height=0.8cm]{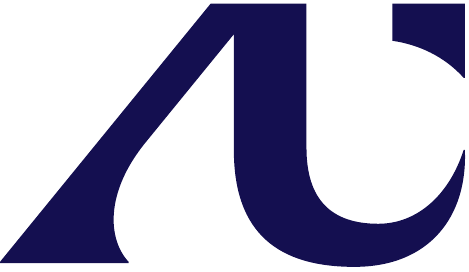}}\hspace{0.3cm}\raisebox{0.18cm}{\includegraphics[height=0.75cm]{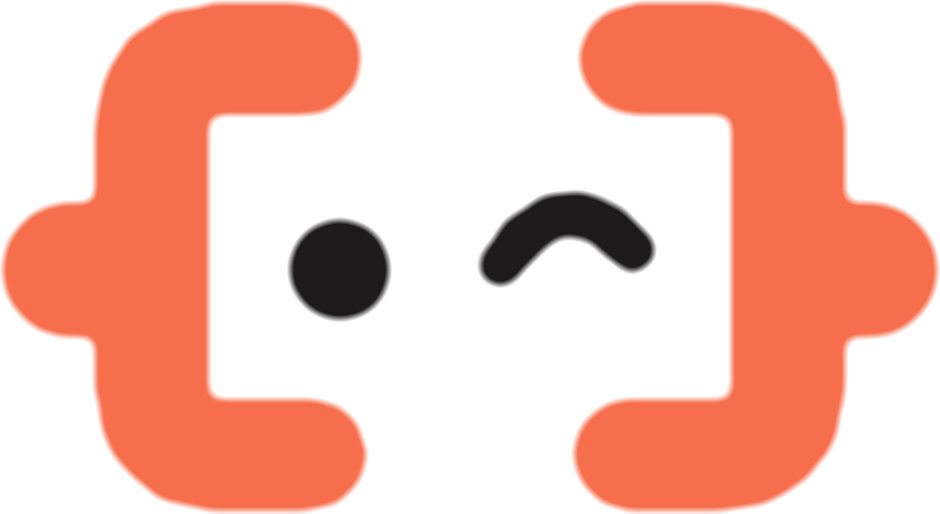}}};}]}{}{\PackageWarning{main}{logo not placed}}

\small{
\affiliation{Tsinghua University}
\affiliation{Peking University}
\affiliation{HKUST(GZ)}
\affiliation{Adelaide University}
\affiliation{RoboProbe}
}

\footnotesize{\contribution[*]{Core Contributor}
\contribution[\dag]{Corresponding Author}
\contribution[\ddag]{Project Lead}}

\abstract{
Embodied intelligence has long posed a formidable challenge and has often been regarded as distinct from the intelligence exhibited
by language models. Yet frontier models' growing competence as virtual agents and their emerging ability to interact with physical
environments suggest a closer connection between these forms of intelligence. We introduce Embodied Virtual-to-Real Agents (EVA), a
conceptual framework for a progressive transition from virtual agency to physical intelligence. Environments
and embodied agents are the basic elements of EVA, and recursive self-improvement is the route by which it would be realised: an
embodied agent first improves through repeated interaction in an initial environment; the environment is then made more complex and
closer to reality, and the agent continues to improve within it. This cycle repeats until the agent is capable enough to enter the
real world, where the same process continues.
 As EVA is an early-stage concept, we gather recent research across language models,
agents, and robotics into possible technical solutions for each step of this pathway, examining which components and
connections existing approaches establish and which remain unsolved. The survey outlines a potential pathway toward general-purpose physical intelligence and identifies open challenges in
sustaining self-improvement across simulated and real environments.
}

\correspondence{Mengdi Xu \email{xumd@mail.tsinghua.edu.cn}}
\date{November 2026}
\metadata[Project Page]{\href{http://roboprobe.org/blog/self-improving-embodied-agents/}{\nolinkurl{roboprobe.org}}}

\begin{document}

\maketitle

\begingroup
\setlength{\intextsep}{6pt}
\begin{figure}[H]
    \centering
    \includegraphics[width=1\linewidth]{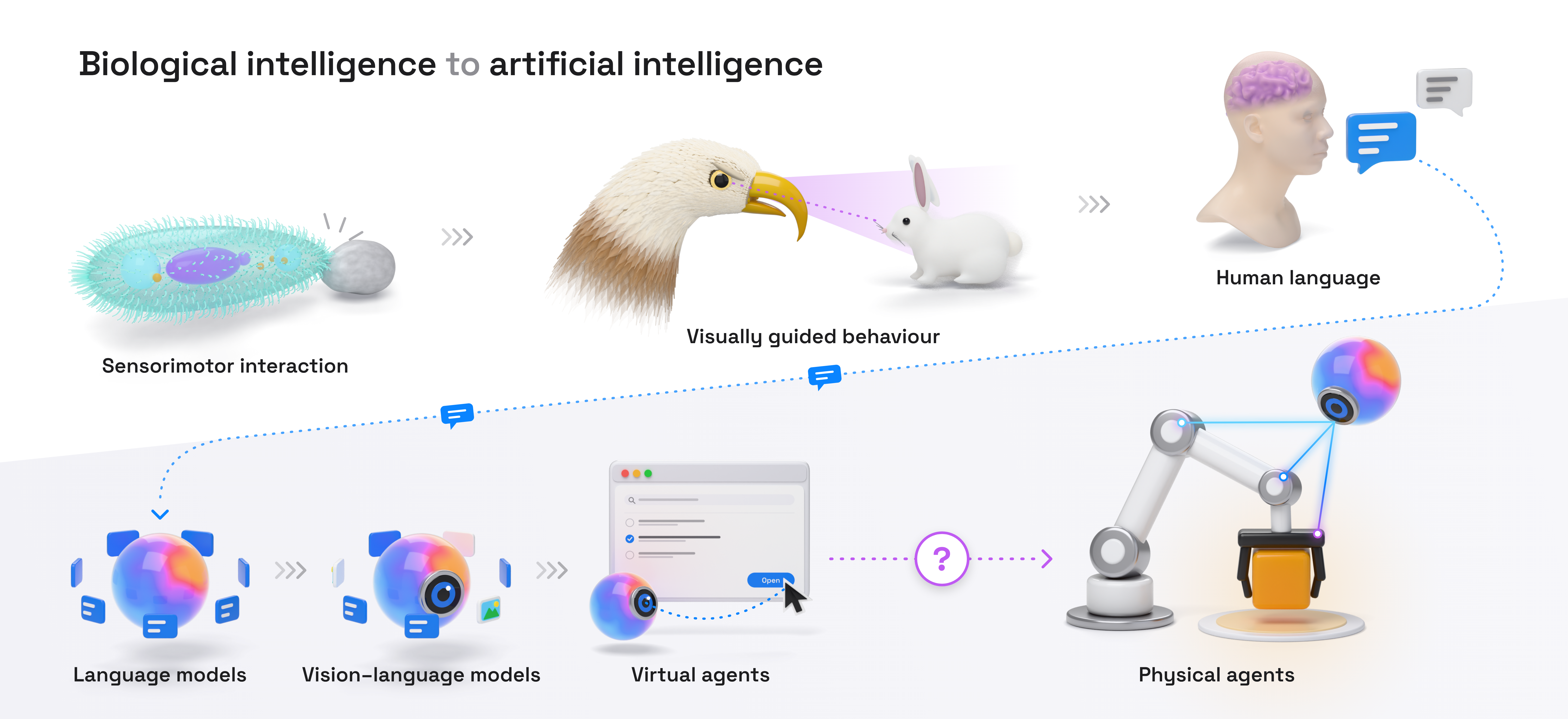}
    \captionsetup{font=small,skip=4pt}
    \caption{\textbf{Embodied Virtual-to-Real Agents (EVA).} A conceptual comparison of biological intelligence, which grew from sensorimotor interaction to visually guided behaviour and human language, with artificial intelligence, which has expanded from language and vision to GUI-based virtual agency. The dashed transition and question mark frame the step from virtual to physical-world agency as an open research direction, for which we propose EVA (\Cref{sec:vrea}).}
    \label{fig:conceptual-teaser}
\end{figure}
\endgroup

\clearpage
\begingroup
\setcounter{tocdepth}{3}
\renewcommand{\cfttoctitlefont}{\Large\sffamily\bfseries}
\setlength{\cftbeforetoctitleskip}{6pt}
\setlength{\cftaftertoctitleskip}{14pt}
\setlength{\cftbeforesecskip}{16pt}
\setlength{\cftbeforesubsecskip}{7pt}
\setlength{\cftbeforesubsubsecskip}{6pt}
\tableofcontents
\endgroup
\clearpage

\section{Introduction}
\label{sec:intro}

A long-standing goal of robotics is to give robots the physical intelligence to perceive, reason and act competently in the physical world.
Towards this goal, two approaches now stand out in research on robot foundation models.
\textbf{Vision--language--action (VLA) models}~\citep{brohan2023rt,kim2024openvla,black20240} build on vision--language pretraining but learn to act from large robot datasets such as Open X-Embodiment~\citep{collaboration2023open}.
Because robot demonstrations are costly to collect at scale, another line of work turns to video to learn how the world moves~\citep{nvidia2025cosmos,gao2026dreamdojo}, with \textbf{world--action models (WAMs)} coupling video prediction with the prediction of robot actions~\citep{li2025unified,nvidia2026cosmos}.
Both routes learn physical competence from data that capture physical interaction.

Compared with interaction in the physical world, interaction in virtual environments has advanced much faster. In software environments, language models can already use tools, write and revise code, and pursue goals through sequences of actions and feedback~\citep{yao2022react,yang2024swe,wang2023survey,team2025kimik2}, a capacity we refer to as \textbf{virtual agency}.
Work on computer use develops and evaluates systems that operate applications through graphical interfaces~\citep{hong2023cogagent,xie2024osworld,qin2025ui,wang2025opencua}.
In 3D virtual worlds, Voyager acquires skills through open-ended exploration in Minecraft~\citep{wang2023voyager}, while SIMA~2 follows instructions across diverse 3D games~\citep{team2025sima}.
Research on spatial understanding includes 3D-LLM, which answers questions about scenes from point clouds~\citep{hong20233d}, and SpatialVLM, which reasons about distances and relative sizes~\citep{chen2024spatialvlm}.
Virtual agents can also build the worlds themselves, as SceneSmith demonstrates by generating simulation-ready indoor scenes from natural-language prompts~\citep{pfaff2026scenesmith}.
These examples show the range of virtual agency.

\begin{figure}[t]
    \centering
    \includegraphics[width=\linewidth]{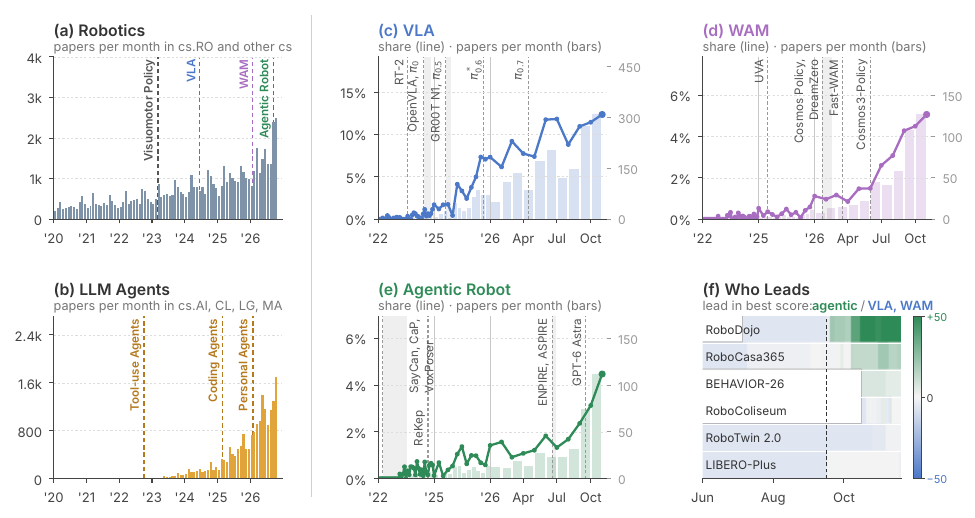}
    \caption{\textbf{Papers on robotics, LLM agents and robot policies by month, and who leads on robot benchmarks.} (a) Robotics papers on arXiv (cs.RO, or other cs categories when robotics is the main subject) and (b) LLM-agent papers (cs.AI, cs.CL, cs.LG, cs.MA), 2020 to October 2026. (c)--(e) Share of robotics papers on VLA models (policies initialised from vision--language weights), world--action models (WAM; one model that predicts video and actions) and agentic robots (a general-purpose agent operates the robot); lines give the share, bars the papers per month, and dashed lines mark representative works. Papers are labelled from their title and abstract by a language model. (f) Lead of the best agentic system over the best VLA or WAM policy on six benchmarks (green: agentic ahead; blue: VLA or WAM ahead; pale blue: no agentic result yet), on full evaluation sets only, as of 20 November 2026.}
    \label{fig:arxiv-trends}
\end{figure}

This agency has been steadily extending from the virtual to the physical world, with language models acting at progressively lower levels of robot control.
They first served as high-level planners, acting as System 2: they select pretrained skills, decompose tasks,
instruct learned low-level policies in language and supervise their execution~\citep{ahn2022i,huang2022inner}.
They then gradually came to generate actions analytically:
as code as policy, they write programs over robot APIs~\citep{liang2022code};
through spatial reasoning~\citep{hong20233d,chen2024spatialvlm} they specify value maps, keypoint constraints or image marks from which a motion planner or solver computes the motion~\citep{huang2023voxposer,huang2024rekep,liu2024moka};
and as coding agents, they write, call and evolve robot tools~\citep{fu2026cap,ge2026make}.
At this level the action is computed by the program, planner or solver, not by the model.
The third level arrived abruptly: in 2026, evaluations showed GPT-6 Astra and Fable 5 issuing low-level robot commands directly with fixed model weights~\citep{zhang2026unexpected,kim2026bridging} (\Cref{sec:llmpolicy}),
although frontier models' manipulation still trails their other abilities~\citep{huang2026frontiera,leet2026bitter}.
These evaluations leave open whether the models' general training included robotics data or whether they underwent dedicated robotics post-training.
As the same agent acts in software, in simulation and on robots, the boundary between virtual and physical agency is beginning to blur.
The robotics literature records the same shift (\Cref{fig:arxiv-trends}): after open vision--language--action and world--action models spread, papers on \textbf{agentic robots}, in which a general-purpose agent itself operates the robot through tools, code or direct actions rather than serving as one module of a pipeline designed by people, rose from about 1\% of monthly robotics papers in early 2026 to 4.5\% in October 2026, following the release of GPT-6 Astra.
Recent experiments show that general-purpose models can guide navigation in simulation~\citep{zhou2026embodied}, use spatial tools for plug insertion on a physical robot~\citep{wang2026spatialharness}, and apply skills learned in simulation to real-robot manipulation~\citep{he2026skill2real}.
We see this progressive extension of virtual agency as a feasible route towards general-purpose physical agency.
Recent surveys and roadmaps organise agentic robots by function or by system component, or review self-improvement in physical AI by mechanism or by the part that changes~\citep{hu2026agentic,qin2026towards,grid2026envisioning,zhou2026who,luo2026recursive,han2026recursive,feng2026self}. Fewer follow one agent along the route itself, from virtual to real environments, with environment construction, embodied agents and self-improvement as stages of one loop.

To frame this route, we put forward Embodied Virtual-to-Real Agents (EVA), a conceptual framework in which a general-purpose agent's competence, first developed in virtual settings, extends to the physical world.
Environments, virtual and real, and embodied agents are the two basic elements of EVA.
Self-improvement connects them: over repeated cycles, the agent keeps and reuses changes to itself and its environment, and we use this loop to organise the reviewed work.
Its stricter, recursive form, in which the improved agent also improves the process that improves it, is the route toward EVA that we examine.
In the pathway we envision, the agent first practises in virtual environments built from real-world data, where practice is cheaper than on real robots, until it can work in the real world without prior real-world experience.
From then on, what it experiences in the real world flows back to make the virtual environment more realistic~\citep{chen2026agentic,qiu2026vid2sid,li2026agentic}, and it keeps improving in both.

\begin{figure}[p]
    \centering
    \input{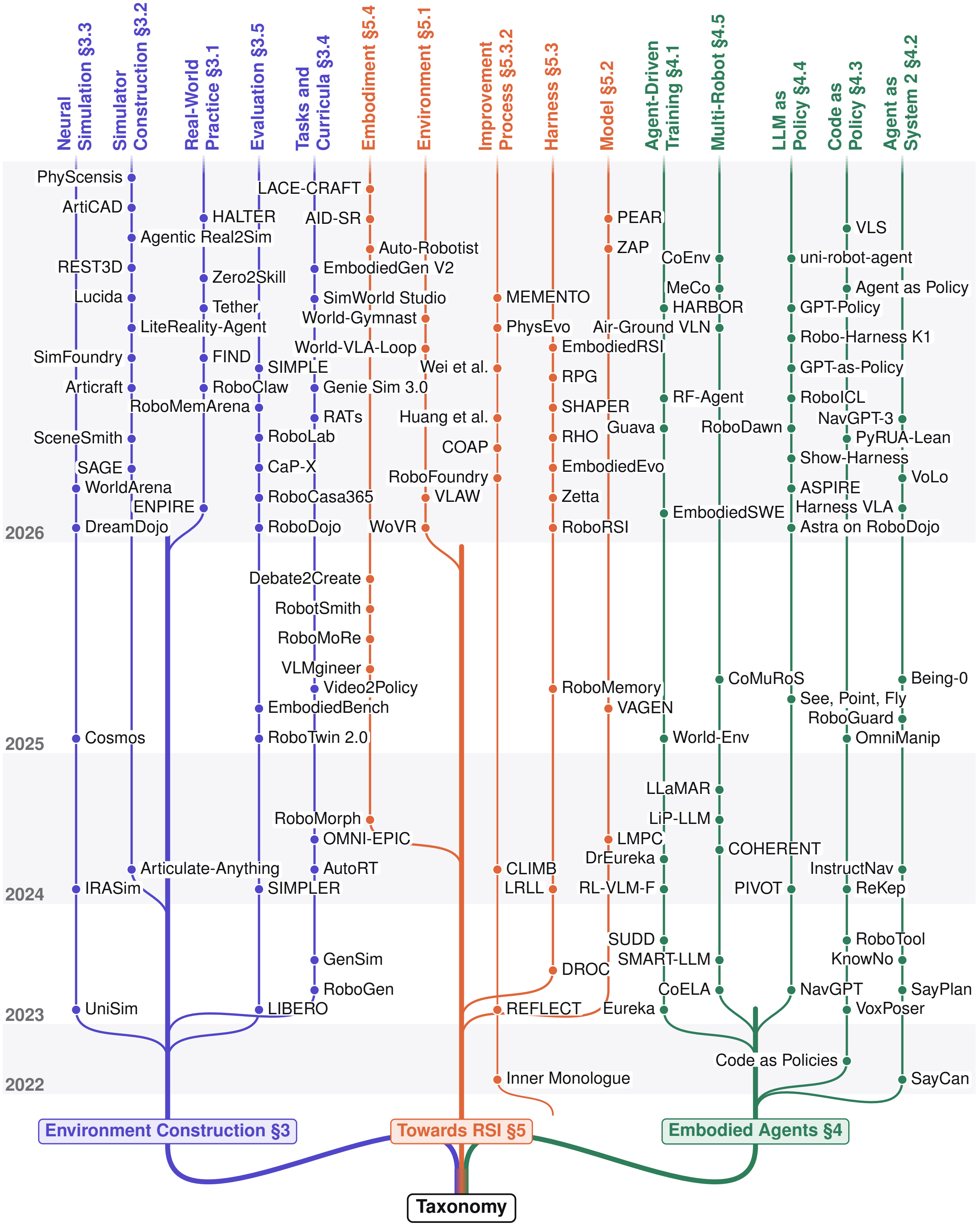}%
    \caption{\textbf{Taxonomy of the reviewed work as a growing tree.} Three trunks, one for each part of this review, split into
    branches, one for each subsection. Works sit on their branch in order of time, upward through the year bands. Each branch names its
    most cited works year by year, and for 2026 those with the most citations relative to their age; the comparison table at the end of
    each subsection lists all.}
    \label{fig:timeline}
\end{figure}

This survey reviews work in robotics, virtual agents, language models and computer vision, with robotics at its centre.
\Cref{sec:coevolution-framework} lays the foundations by defining the environment (virtual or real) and the embodied agent with its foundation model, harness and embodiment, introducing recursive self-improvement, and setting out the EVA framework.
We organise the reviewed work into three categories (\Cref{fig:timeline}).
Environment construction (\Cref{sec:sim}) covers real-world stations and resets, simulated and neural environments, task generation, evaluation and benchmarks~\citep{wang2023robogen,pfaff2026scenesmith}.
The review of embodied agents in learning and execution (\Cref{sec:embodied-agent}) covers policy training, the agent as System 2, code as policy, models acting as the policy, and multi-robot collaboration~\citep{ha2023scaling,chu2026openrua}.
Work towards recursive self-improvement (\Cref{sec:selfimprove}) is organised by the part that changes across cycles, whether the environment, the foundation model, the harness or the embodiment~\citep{xiao2026enpire,wang2023voyager}.
\Cref{sec:discussion} discusses the problems these categories share (latency, failure, cost, and safety and security), and \Cref{sec:conclusion} concludes.

\section{Foundations and Framework of Embodied Agents}
\label{sec:coevolution-framework}

This section develops the conceptual foundations of embodied agents in three parts. \Cref{sec:coevolution-definitions}
defines the environment, virtual or real, and the embodied agent: where its boundary lies, its foundation model, harness
and embodiment, and how these work as one loop while a task runs (\Cref{sec:def-runtime}). \Cref{sec:coevolution-roles} introduces recursive self-improvement, the route this survey examines, and then defines the three categories of the reviewed
research in the order in which the survey reviews them: environment construction (\Cref{sec:sim}), embodied agents (training and deployment, \Cref{sec:embodied-agent}) and self-improvement, the main line of the survey, which builds on both (\Cref{sec:selfimprove}). \Cref{sec:vrea} then sets out the
Embodied Virtual-to-Real Agents (EVA) framework.

\subsection{Definition}
\label{sec:coevolution-definitions}
\label{sec:coevolution-environment}

\subsubsection{Environment}
\label{sec:def-environment}

\paragraph{Definition.} The \textbf{environment} $\mathcal E$ is what lies outside the agent: the agent's actions change it, and the
agent receives its observations from it. It is of two kinds, virtual and real.

 A virtual environment $\mathcal E_V$ is an environment simulated by a computer. It takes one of two forms:
a physics simulation with rendering, whose state is explicit (\Cref{sec:sim-construct}), as in
RoboGen~\citep{wang2023robogen} and SceneSmith~\citep{pfaff2026scenesmith}; or a world model learned from data, whose state is
implicit (\Cref{sec:sim-neural}), as in UniSim~\citep{yang2023learning} and Ctrl-World~\citep{guo2025ctrl}. The agent acts in it
through a simulated or modelled body. Because it is computer-simulated, it can be configured, reset and run in parallel by program.

A real environment $\mathcal E_R$ is the physical environment in the real world, with its scene, objects
and people; an object that the robot manipulates belongs to it. The agent acts in it through the real robot and observes it through
real sensors. Its state cannot be set by program, so preparing or restoring it takes physical action (\Cref{sec:sim-reset}).

\subsubsection{Embodied Agents}
\label{sec:def-agent}

\paragraph{Embodied agent.} An embodied agent $\mathcal A=(F_\theta,H,B)$ is a foundation model in its harness together with a body,
through which it perceives and acts in a virtual or real environment (\Cref{sec:def-environment}). We set out the principle that
separates it from its environment and its three parts: the foundation model $F_\theta$, the harness $H$ and the embodiment $B$
(\Cref{fig:agent-parts}). Throughout, $k$ indexes development cycles. The agent comprises the model that decides, the harness that runs it and the body through which
it acts, wherever each of them runs: the model may be served in the cloud and the harness on a workstation, and in a virtual
environment the body is itself simulated. Everything the agent acts on and senses through its body, including the objects it
manipulates and the people around it, belongs to the environment. The body belongs to the agent because the agent can change it,
choosing or designing a configuration, such as an end-effector, for a task (\Cref{sec:si-embodiment}). An agent at development cycle
$k$, and the system that self-improvement can change (\Cref{sec:roles-self}), are therefore
\begin{equation}
\mathcal A_k=\big(F_{\theta_k},\,H_k,\,B_k\big),
\label{eq:agent-system}
\end{equation}
where $F_{\theta_k}$ is the foundation model with parameters $\theta_k$, $H_k$ the harness and $B_k$ the embodiment at cycle $k$.

\paragraph{General agent.} A general agent $\mathcal G=(F,H)$, such as a coding agent, is a foundation model in its harness with no
body; it works only in software, for instance building the embodied agent's environment (\Cref{sec:roles-environment}).

\paragraph{Foundation model.} The foundation model $F_\theta$ is the brain of the agent, the model that makes consequential decisions: what to do
next, which tool or policy to call, and which program, world, reward or design to write. Not every model in the
agent is a foundation model: a VLA is $F_\theta$ when it makes the high-level decisions, but when it is called, like a segmentation
model such as SAM or a value model, it is a tool in $\mathcal T_k$. This definition by role sets the framework;
\phantomsection\label{sec:scope-corpus}the scope of the review is narrower.  Systems whose $F_\theta$
is a robot-trained model fit the framework but lie outside the reviewed corpus, and appear only as tools or reference work. ``The
model'' means $F_\theta$ only unless specified.

\paragraph{Harness.} The harness $H$ is the supporting software system around the brain of  the embodied agent to finish tasks. Following Barbaste \etal, who define the harness of coding agents~\citep{barbaste2026harness}, we take the
harness $H_k$ to be all the software outside the foundation model, defined as
\begin{equation}
H_k=\big(\mathcal T_k,\,\mathcal O_k,\,\mathcal M_k\big),
\label{eq:harness}
\end{equation}
where $\mathcal T_k$ are the tools, $\mathcal O_k$ the orchestration and $\mathcal M_k$ the memory at cycle $k$. They are the three
sides of how the model is run: what it sees, what it can do, and what controls it.
\begin{itemize}
    \item \textbf{Memory} $\mathcal M_k$, the input side, is what the agent stores: short-term (working) memory within the current
    task, long-term memory kept across tasks, and logs of execution and outcomes. The \textbf{context} is the part of memory placed
    in the model's input at a decision.
    \item \textbf{Tools} $\mathcal T_k$, the action side, are the capabilities that the model can call: perception, evaluation and
    action tools. A \textbf{policy} $\pi_\phi\in\mathcal T_k$ is a learned action tool trained on robot or embodied data, either
    general, such as VLAs and WAMs, or specialised to one task or skill.
    \item \textbf{Orchestration} $\mathcal O_k$, the control side, decides when the model is called, how results are verified and
    when to stop; it covers the control loop, monitoring and recovery, safety rules, the reading and writing of memory and, where the
    system trains, the learning pipeline. The model decides what the next step is. The learning pipeline is used only when the agent
    improves, so it belongs to the improvement process $\Phi_k$ (\Cref{sec:roles-self}) rather than to the agent's execution.
\end{itemize}

\paragraph{Embodiment.} The embodiment $B_k$ is the hardware with which an embodied agent acts and perceives: the body, sensors, actuators and hand-held tools through which actions take
effect, together with the drivers and low-level controllers shipped with the robot; in a virtual environment it is simulated or
modelled. Its sensors supply the observations that the model and the harness work from, and its actuators turn their commands into
motion and contact, so it bounds what the agent can perceive and do. Like the harness, it can change across development cycles: the
agent can propose a new body or tool and evaluate it in its task (embodiment improvement, \Cref{sec:si-embodiment}).

\paragraph{Runtime.}\phantomsection\label{sec:def-runtime} The three parts work as one loop while a task runs. At each decision, orchestration $\mathcal O_k$ assembles the
context from memory $\mathcal M_k$: the long-term memory, fixed within the episode, and the short-term memory of the episode, namely
the goal $g$, the observations $o_{\le t}$ and the model's earlier outputs $y_{<t}$. The model $F_\theta$ returns an output $y_t$;
orchestration hands it to an action tool $\alpha\in\mathcal T_k$, which turns it into commands that the embodiment $B_k$ executes;
and orchestration checks the returned evidence against the current subgoal, writes the outcome to memory and calls the model again,
to continue, retry, replan, ask or stop. The forms this loop takes are set out in \Cref{sec:roles-execution}.

\subsection{Self-Improving Agent}
\label{sec:coevolution-roles}

\begin{figure}[!htb]
    \centering
    \def\catlink#1#2#3#4#5{%
      \pgfmathsetlengthmacro\catw{(#3-#1)*\catW}%
      \pgfmathsetlengthmacro\cath{(#4-#2)*\catH}%
      \node[inner sep=0pt, anchor=north west] at ({#1*\catW},{\catH-#2*\catH}) {\hyperref[#5]{\rule{0pt}{\cath}\rule{\catw}{0pt}}};}%
    \begin{tikzpicture}
      \node[inner sep=0pt, anchor=south west] (catimg) at (0,0) {\includegraphics[width=\linewidth]{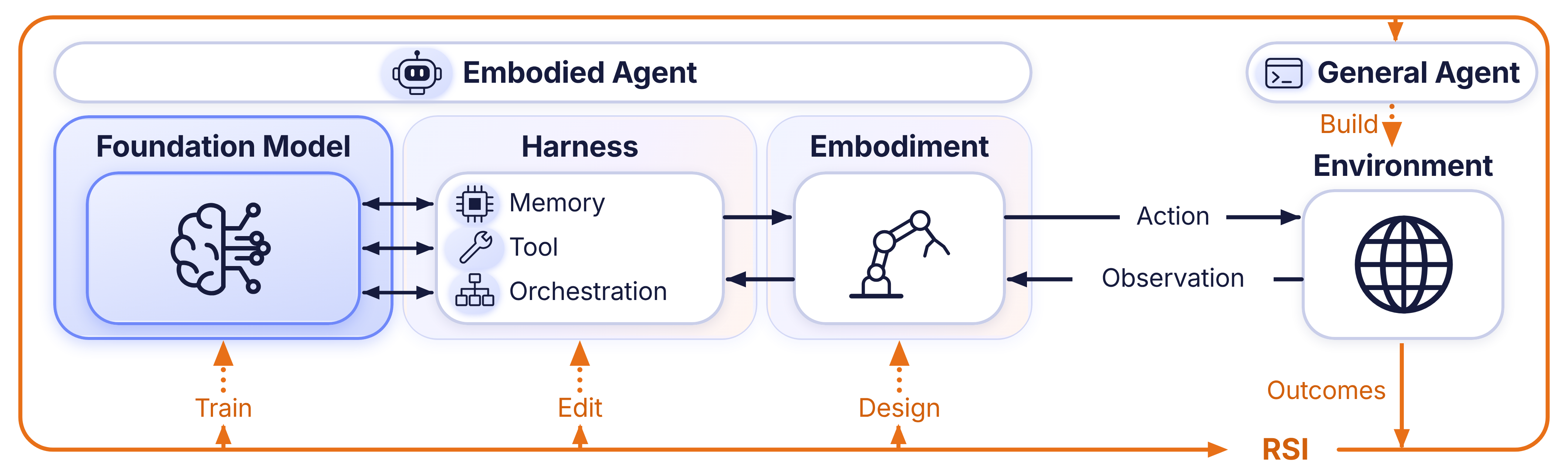}};
      \path (catimg.north east); \pgfgetlastxy{\catW}{\catH}%
      \catlink{0.095}{0.842}{0.190}{0.911}{sec:si-foundation}%
      \catlink{0.323}{0.842}{0.417}{0.911}{sec:si-harness}%
      \catlink{0.526}{0.842}{0.621}{0.911}{sec:si-embodiment}%
      \catlink{0.821}{0.234}{0.879}{0.304}{sec:sim}%
      \catlink{0.792}{0.924}{0.848}{1.000}{sec:selfimprove}%
    \end{tikzpicture}
    \caption{\textbf{An embodied agent and its environment.} The embodied agent $\mathcal A_k=(F_{\theta_k},H_k,B_k)$
    (\Cref{eq:agent-system}) consists of the foundation model, the harness, with its memory, tools and orchestration
    (\Cref{eq:harness}), and the embodiment; it acts in an environment, real or simulated (\Cref{sec:def-environment}).
    In the runtime loop (dark arrows), the harness passes prompts to the model and its responses on to the embodiment as intended
    actions, and the embodiment acts on the environment and returns what it perceives as observations. The improvement loop (orange)
    starts where a task ends: the outcomes of the agent's tasks, its experience $\xi_k$, leave the environment (arrow labelled
    Outcomes) and run clockwise round the frame, as its arrowhead at the bottom shows, to where the loop ends, the arrows that train
    the model, edit the harness, design the embodiment and, through the general agent, build or revise the virtual environment.
    The changed agent then runs its next tasks in the runtime loop, and their outcomes start the next cycle. This loop is
    self-improvement (\Cref{sec:roles-self}); the figure labels it RSI, which in the text names only its strict, recursive form,
    in which the change also reaches the improvement process. The parts are defined in \Cref{sec:def-agent}.}
    \label{fig:agent-parts}
\end{figure}

The route this survey examines is strict recursive self-improvement (RSI): the improved agent improves its improved self. The change made in
one cycle then reaches not only the agent but also the improvement process $\Phi_k$, the reward code, task generators, training
pipeline and verifiers used to form and accept changes, so that $\Phi_{k+1}$ is itself a product of the previous cycle and forms the
next change; it is shown only when the changed process makes the next improvement better than a frozen $\Phi_k$
(\Cref{sec:si-process}). The goal comes from software: a Gödel machine rewrites its own code once it has proved the rewrite
useful~\citep{schmidhuber2003goedel}, and the Darwin Gödel Machine, STOP and Gödel Agent validate changes empirically, so that the
improved code, being itself the improver, makes the next change~\citep{zhang2025darwin,zelikman2023self,yin2024godel,ren2026self}.
For the embodied agent of \Cref{sec:def-agent}, the loop in which such changes arise makes it a self-improving agent
(\Cref{fig:agent-parts}); the loop extends the act--observe loop of language-model agents such as ReAct~\citep{yao2022react} to a robot.
In the runtime loop, the harness assembles the model's context from its memory $\mathcal M_k$, an action tool
$\alpha\in\mathcal T_k$ turns the model's output $y_t$ and the observation $o_t$ into a command $u_t=\alpha(y_t,o_t)$, and the
embodiment $B_k$ acts and returns the next observation $o_{t+1}$; run on a task $z=(\rho_0,g,c)$, with initial condition $\rho_0$, goal
$g$ and success criterion $c$, the loop gives an episode. In the improvement loop (orange), the experience $\xi_k$ of development
cycle $k$ reaches the model, which uses it to revise the agent's parts or its virtual environment; this loop becomes recursive
when the change also reaches $\Phi_k$. Recursive self-improvement presupposes self-improvement (\Cref{sec:roles-self}), the main line
along which the survey reviews the work: most of the reviewed work reaches self-improvement, and few systems make a recursive update
(\Cref{sec:selfimprove}).

Work on the self-improving agent falls into three categories (\Cref{fig:timeline}): environment construction (\Cref{sec:roles-environment}) builds the environment in which the loop runs, the embodied
agent (\Cref{sec:roles-embodied}) trains the policy that the loop calls (agent learning) and takes part in the loop (agent
execution), and self-improvement (\Cref{sec:roles-self}) revises the agent itself across cycles. They are defined here, and reviewed, in this order:
environment construction (\Cref{sec:sim}), embodied agents (\Cref{sec:embodied-agent}) and then self-improvement, the main line of the survey, which builds on both (\Cref{sec:selfimprove}). A system
that combines several is reviewed where its main contribution lies.

\subsubsection{Environment Construction}
\label{sec:roles-environment}

\begin{figure}[!htb]
    \centering
    \def\catlink#1#2#3#4#5{%
      \pgfmathsetlengthmacro\catw{(#3-#1)*\catW}%
      \pgfmathsetlengthmacro\cath{(#4-#2)*\catH}%
      \node[inner sep=0pt, anchor=north west] at ({#1*\catW},{\catH-#2*\catH}) {\hyperref[#5]{\rule{0pt}{\cath}\rule{\catw}{0pt}}};}%
    \begin{tikzpicture}
      \node[inner sep=0pt, anchor=south west] (catimg) at (0,0) {\includegraphics[width=\linewidth]{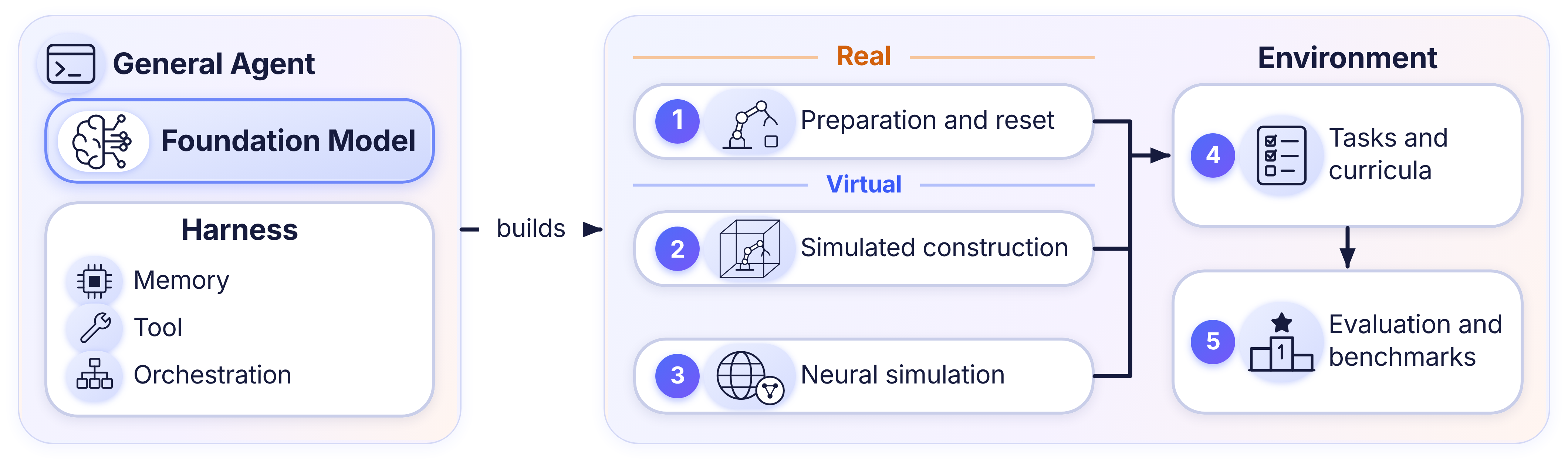}};
      \path (catimg.north east); \pgfgetlastxy{\catW}{\catH}%
      \catlink{0.404}{0.181}{0.698}{0.348}{sec:sim-reset}%
      \catlink{0.404}{0.458}{0.698}{0.626}{sec:sim-construct}%
      \catlink{0.404}{0.735}{0.698}{0.903}{sec:sim-neural}%
      \catlink{0.747}{0.181}{0.972}{0.497}{sec:learning-tasks}%
      \catlink{0.747}{0.587}{0.972}{0.903}{sec:sim-bench}%
    \end{tikzpicture}
    \caption{\textbf{Environment construction.} The agent, a foundation model in its harness, works on the environment: (1)~it prepares and resets a real station; (2)~it builds a simulated environment; (3)~it learns a world model as a neural simulation; (4)~on these environments it defines and builds the tasks and curricula that a policy practises; and (5)~it builds the benchmarks and evaluation methods that score the policy or the agent.}
    \label{fig:cat-environment}
\end{figure}

Environment construction (\Cref{sec:sim}, \Cref{fig:cat-environment}) is the work in which the model $F_\theta$ builds, prepares or
resets the environment, virtual or real, in which robots practise and are tested. Its five branches follow the order in which a world
is made ready and used.
Real-world preparation, reset and reset-free practice (\Cref{sec:sim-reset}) makes real-world practice repeatable: from the
observation $o$ of the station after a rollout, it restores and checks the starting state $\rho_0$, or chooses a task suited to the
current scene. Simulated environment construction (\Cref{sec:sim-construct}) writes $\mathcal E_V$ as a simulation program, and
neural simulation environment construction (\Cref{sec:sim-neural}) obtains it as a world model learned from interaction data. Task
generation and curriculum design (\Cref{sec:learning-tasks}) writes a goal and a success check, and may choose the next task from the
learner's record. Evaluation and benchmarks (\Cref{sec:sim-bench}) fixes how a policy or an agent is tested: with
$\tau_i=(o_0,y_0,u_0,\dots,o_T)$ the episode on task $z_i$ and $c$ the success criterion as implemented, $c(\tau_i)\in\{0,1\}$
records whether the episode succeeds, and the score $J$ is the success rate over $N$ tasks.

\subsubsection{Embodied Agents}
\label{sec:roles-embodied}
\label{sec:roles-learning}

\begin{figure}[!htb]
    \centering
    \def\catlink#1#2#3#4#5{%
      \pgfmathsetlengthmacro\catw{(#3-#1)*\catW}%
      \pgfmathsetlengthmacro\cath{(#4-#2)*\catH}%
      \node[inner sep=0pt, anchor=north west] at ({#1*\catW},{\catH-#2*\catH}) {\hyperref[#5]{\rule{0pt}{\cath}\rule{\catw}{0pt}}};}%
    \begin{tikzpicture}
      \node[inner sep=0pt, anchor=south west] (catimg) at (0,0) {\includegraphics[width=\linewidth]{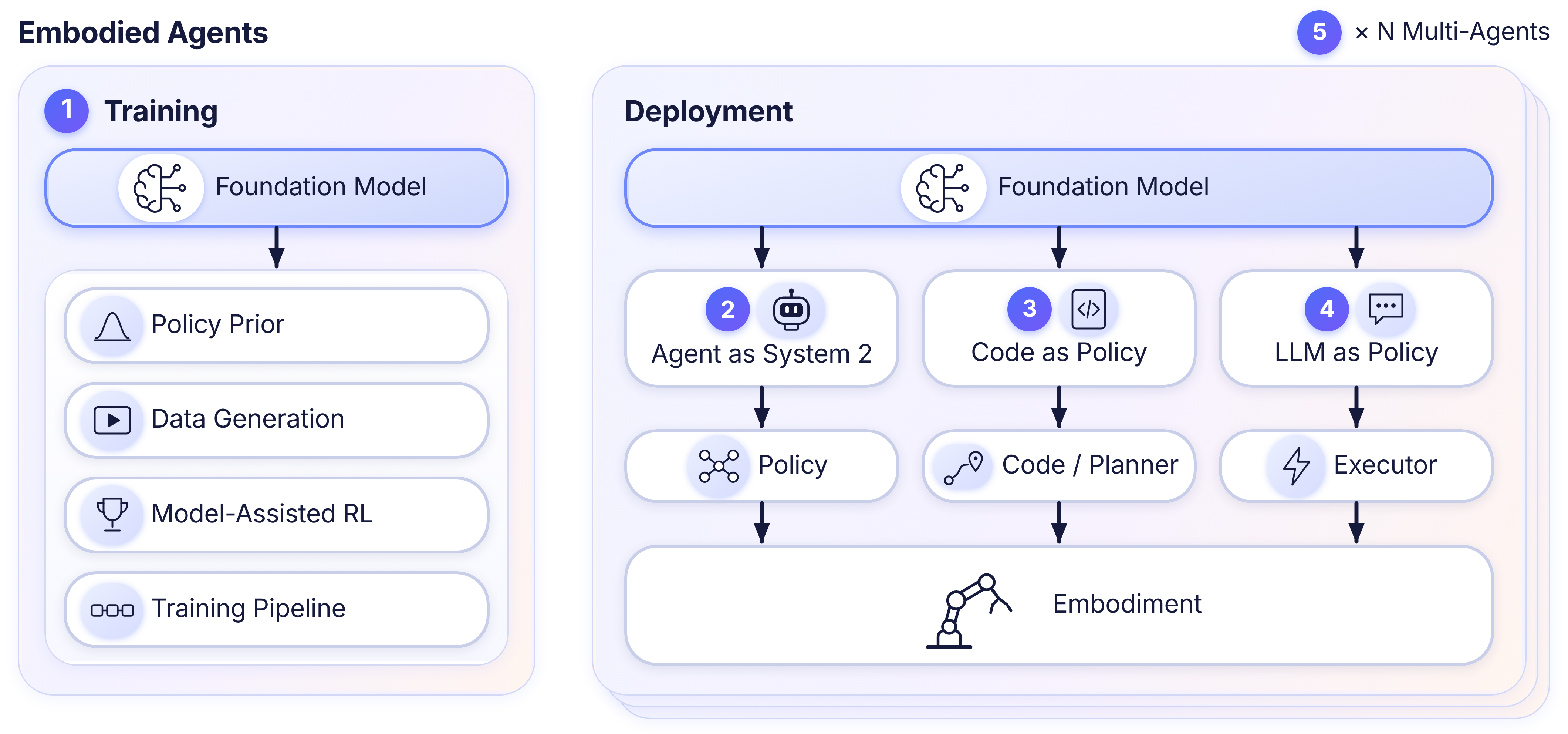}};
      \path (catimg.north east); \pgfgetlastxy{\catW}{\catH}%
      \catlink{0.028}{0.121}{0.057}{0.181}{sec:learning-frameworks}%
      \catlink{0.041}{0.391}{0.312}{0.496}{sec:learning-priors}%
      \catlink{0.041}{0.520}{0.312}{0.625}{sec:learning-demos}%
      \catlink{0.041}{0.649}{0.312}{0.754}{sec:learning-rl}%
      \catlink{0.041}{0.778}{0.312}{0.883}{sec:learning-pipeline}%
      \catlink{0.398}{0.367}{0.574}{0.528}{sec:slowfast}%
      \catlink{0.588}{0.367}{0.763}{0.528}{sec:code}%
      \catlink{0.777}{0.367}{0.953}{0.528}{sec:llmpolicy}%
      \catlink{0.827}{0.014}{0.856}{0.075}{sec:others}%
    \end{tikzpicture}
    \caption{\textbf{What the model does for the robot.} In training (\Cref{sec:learning-frameworks}) the foundation model supplies a policy prior, generates demonstrations, designs rewards, gives feedback or builds the training pipeline, from which the policy is learned. In deployment, while a task runs, the model's output $y_t$ reaches an action tool at one of three levels: a semantic instruction to the policy (agent as System 2, \Cref{sec:slowfast}), a program or specification run as code or by a planner (Code as Policy, \Cref{sec:code}), or a low-level action to an executor (LLM as Policy, \Cref{sec:llmpolicy}); the embodiment carries out the command. Several such agents may work together (multi-robot collaboration, \Cref{sec:others}).}
    \label{fig:cat-embodied}
\end{figure}

This category (\Cref{sec:embodied-agent}, \Cref{fig:cat-embodied}) concerns what the model $F_\theta$ does for the robot itself: it
helps a learning algorithm train the robot's policy (policy training, \Cref{sec:learning-frameworks}), and it takes part in the robot's behaviour while a task runs (agent execution),
by commanding actions, judging progress or deciding how to recover. \Cref{sec:embodied-agent} also reviews Multi-Robot Collaboration
(\Cref{sec:others}), which extends the agent to a team.

In policy training (\Cref{sec:learning-frameworks}), the model supplies one input of a learning algorithm, a prior
policy written as code, demonstrations, a reward, feedback on rollouts or the training pipeline itself, and the learner trains the
policy from it; the quality of the trained policy measures the work.

\phantomsection
\label{sec:roles-execution}

In agent execution (\Cref{sec:deployment}), the model takes part through the runtime loop of \Cref{sec:def-agent}. The three parts
(\Cref{sec:slowfast,sec:code,sec:llmpolicy}) differ in the level of the output $y_t$, which fixes the action tool $\alpha$ that turns
it into a command: a semantic instruction goes to a learned policy (Agent as System 2), a program or specification to an interpreter,
planner or solver (Code as Policy), and a low-level action such as a pose target or a waypoint to an executor (LLM as Policy). A
judgement in runtime monitoring, like a plan not yet executed, issues no command and only enters the history. The level also sets
when the model is called: an executor waits for the model at every step; an interpreter runs a program written once and returns to
the model only on failure or at the end; and a policy keeps acting while the model is woken by an event or a finished subgoal. How
much the model's latency costs the robot depends on this timing (\Cref{sec:latency}).

\subsubsection{Self-Improvement}
\label{sec:roles-self}

\begin{figure}[!htb]
    \centering
    \def\catlink#1#2#3#4#5{%
      \pgfmathsetlengthmacro\catw{(#3-#1)*\catW}%
      \pgfmathsetlengthmacro\cath{(#4-#2)*\catH}%
      \node[inner sep=0pt, anchor=north west] at ({#1*\catW},{\catH-#2*\catH}) {\hyperref[#5]{\rule{0pt}{\cath}\rule{\catw}{0pt}}};}%
    \begin{tikzpicture}
      \node[inner sep=0pt, anchor=south west] (catimg) at (0,0) {\includegraphics[width=\linewidth]{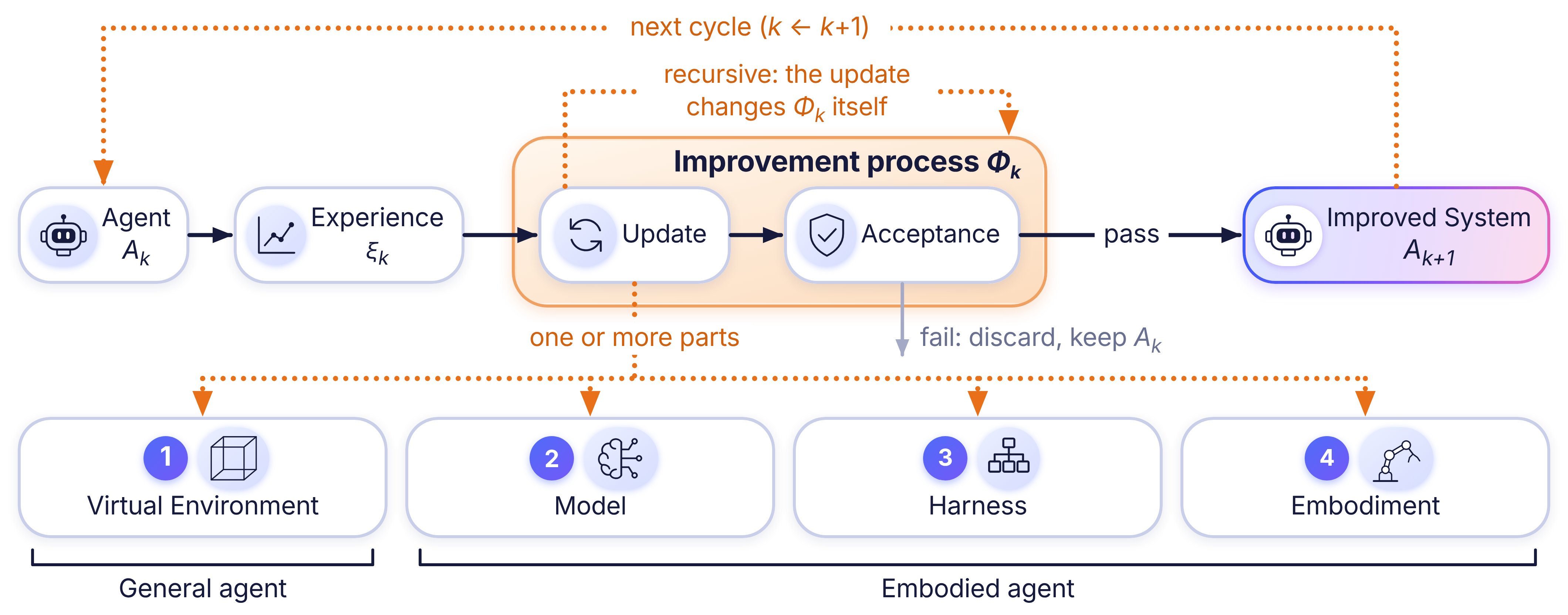}};
      \path (catimg.north east); \pgfgetlastxy{\catW}{\catH}%
      \catlink{0.011}{0.678}{0.247}{0.875}{sec:si-environment}%
      \catlink{0.258}{0.678}{0.494}{0.875}{sec:si-foundation}%
      \catlink{0.506}{0.678}{0.742}{0.875}{sec:si-harness}%
      \catlink{0.753}{0.678}{0.989}{0.875}{sec:si-embodiment}%
    \end{tikzpicture}
    \caption{\textbf{Self-improvement.} In one development cycle the agent $\mathcal A_k$ runs its tasks; the improvement process $\Phi_k$ forms an update from its experience $\xi_k$ and keeps the changed parts only if they pass the acceptance condition of \Cref{eq:roles-self}; many systems use a weaker gate. Repeated over cycles, with each change kept and reused in later tasks, this is self-improvement. An update is recursive when the change reaches $\Phi_k$ itself, such as the reward code, task generators, training pipeline or verifiers used to form and accept updates (the arrow labelled recursive). One or more parts change in each update: the virtual environment, the model, the harness or the embodiment (bottom, 1--4).}
    \label{fig:cat-rsi}
\end{figure}

Self-improvement (\Cref{sec:selfimprove}, \Cref{fig:cat-rsi}) is the work in which the agent changes its own parts from the
experience of its own tasks and keeps the changes that make it better. For an embodied agent, work is admitted to
\Cref{sec:selfimprove} when both conditions hold: (1)~the change acts on the agent $\mathcal A_k$ itself, its model, its
harness or its body, with the virtual environment in which it practises as the one exception; (2)~each change is
kept and reused in later episodes or tasks over repeated cycles. The stricter standard adds (3): each change passes an acceptance metric, so that it improves the agent's
performance on its later tasks. The comparison tables record the gate and the reuse of each work. Recursive self-improvement, the stricter form defined in \Cref{sec:coevolution-roles}, is the
route this survey examines; this subsection and \Cref{sec:selfimprove}, which works along that route, cover self-improvement.

The changed part gives the four routes of \Cref{sec:selfimprove}: environment improvement (\Cref{sec:si-environment}) revises the
virtual environment and keeps the revision when the agent's later tasks improve; foundation model improvement (\Cref{sec:si-foundation}) updates
the parameters $\theta$ of the foundation model; harness improvement
(\Cref{sec:si-harness}) keeps them fixed and revises the long-term memory that the model reads (context improvement), the
orchestration and the other tools (tool and orchestration improvement) or the policy, a learned tool, by training its weights $\phi$
(policy model improvement); and embodiment improvement (\Cref{sec:si-embodiment}) changes
the robot's body or the tool it holds. An update and its acceptance are
\begin{equation}
\begin{gathered}
x_{k+1}=\mathcal U_{\Phi_k}(x_k;\,\xi_k),\qquad
x_{k}=\begin{cases}
\mathcal E_{V,k} & \text{environment improvement},\\
\theta_{k}\ & \text{foundation model improvement},\\
\mathcal M_k,\ \mathcal O_k,\ \mathcal T_k & \text{harness improvement},\\
B_k & \text{embodiment improvement},
\end{cases}\\[2pt]
\text{kept only if}\quad J(x_{k+1})\ge J(x_k),
\end{gathered}
\label{eq:roles-self}
\end{equation}
where $x_k$ is the part that changes at cycle $k$, the virtual environment or a part of the agent (\Cref{eq:agent-system}), and
$\mathcal U_{\Phi_k}$ the update that $\Phi_k$ forms from the experience $\xi_k$. Each new version $x_{k+1}$ is thus obtained from
$x_k$ by one iteration and validated before it is kept: $J(x)$ is the success rate that the agent obtains with $x$ in place, on later
tasks not used to form the change (the evaluation score of \Cref{sec:roles-environment}), and $x_{k+1}$ replaces $x_k$ only if this
score does not fall. Work with a weaker gate, such as a compile check, is still
reviewed when the change is kept and reused, and its gate limits what it can show (\Cref{sec:si-limits}).

\subsection{Embodied Virtual-to-Real Agents (EVA)}
\label{sec:vrea}

\begin{figure}[!htb]
    \centering
    \includegraphics[width=\linewidth]{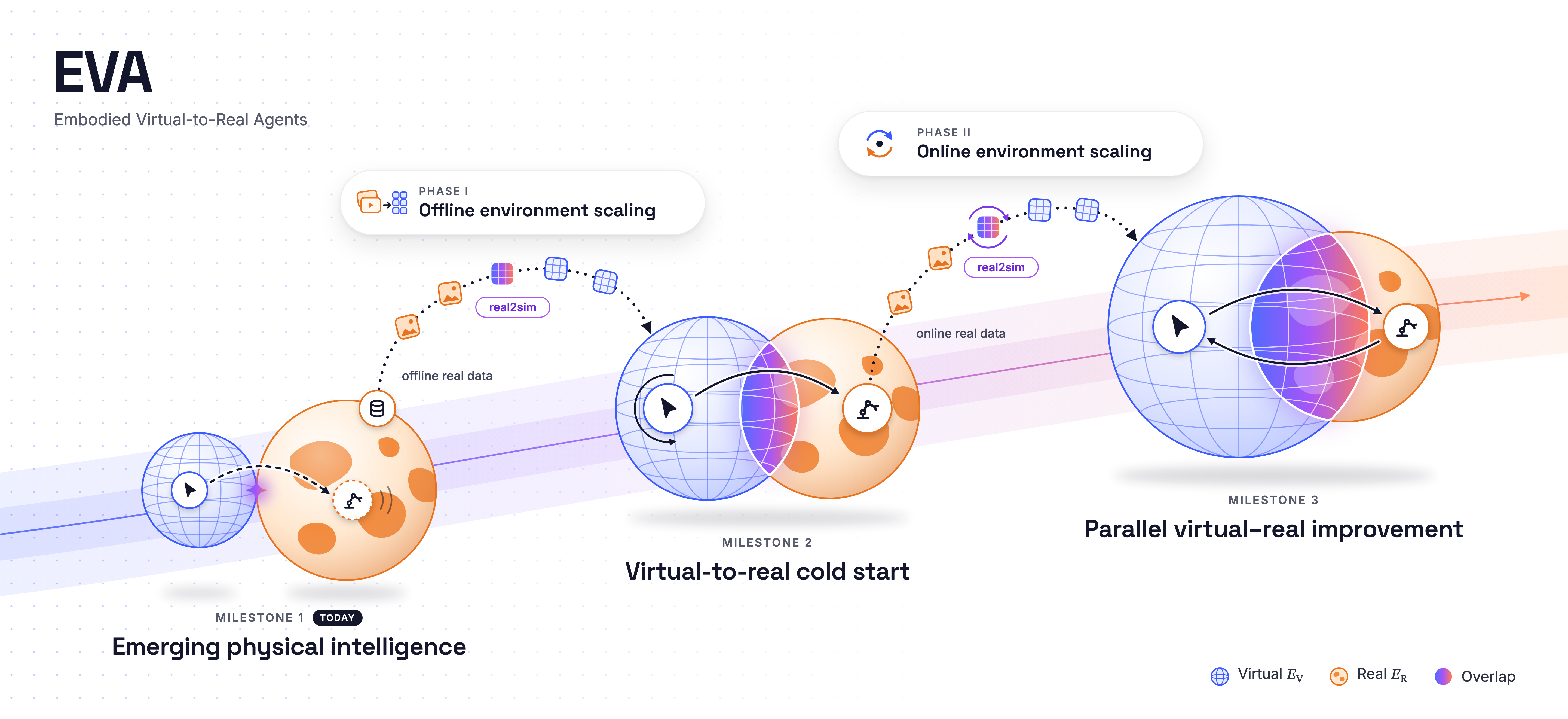}
    \caption{\textbf{The EVA pathway.} Three snapshots, left to right, show Milestone 1: emerging physical intelligence;
    Milestone 2: virtual-to-real cold start; and Milestone 3: parallel virtual--real improvement.
    The two arrows between them represent Phase I: offline environment scaling \& embodied agent self-improvement, and Phase II: online real-to-sim scaling.
    The two circles represent the virtual environment $\mathcal E_V$ and the real environment $\mathcal E_R$.
    The orange loop is the self-improvement loop of \Cref{fig:agent-parts}; the bottom arrow indicates growing capability.}
    \label{fig:eva-pathway}
\end{figure}

Embodied Virtual-to-Real Agents (EVA) is the conceptual framework that we propose for the step from virtual to robotic agents left open in \Cref{fig:conceptual-teaser}: a
general-purpose agent whose competence, first developed in virtual settings, extends to the physical world (\Cref{fig:eva-pathway}).

Two phases connect the three milestones: Phase I, offline environment scaling, and Phase II, online
real-to-sim scaling. \textbf{Milestone 1, emerging physical intelligence}: GPT-6 Astra and Fable 5 are the frontier models that first complete some physical tasks through their own low-level control, though unreliably, with weak precision and dynamic control as examples~\citep{zhang2026unexpected,kim2026bridging} (\Cref{sec:llmpolicy}); earlier frontier models were limited to high-level planning or code~\citep{ahn2022i,huang2022inner,liang2022code}; $\mathcal E_V$ and $\mathcal E_R$ just touch (\Cref{sec:def-environment}).
\textbf{Milestone 2, virtual-to-real cold start}: competence gained in $\mathcal E_V$ transfers stably enough to $\mathcal E_R$, without prior real-world experience, for the system to interact freely there.
\textbf{Milestone 3, parallel virtual--real improvement}: $\mathcal E_V$ and $\mathcal E_R$ broadly overlap, and the system improves in both at once.
We define \textbf{overlap} as the part of $\mathcal E_V$ that is sufficiently aligned with $\mathcal E_R$ in both visual appearance and dynamics.
Overlap arises as a by-product of scaling with extensive randomisation of scenes, appearance and physical parameters, rather than checking scenes one by one.
The loop starts in $\mathcal E_V$, which virtual agents can already operate through software interfaces~\citep{wang2023robogen,chu2026openrua}.

\paragraph{Phase I: offline environment scaling.} From Milestone 1, \textbf{emerging physical intelligence}, we propose continuing to
collect offline real-world data to scale $\mathcal E_V$ and letting the embodied system improve itself there until its cold-start
performance in $\mathcal E_R$ is stable enough to reach Milestone 2, \textbf{virtual-to-real cold start}.
Enough overlap for the cold start is indicated by agreement in aggregate performance between virtual and real trials; the intended transfer of reusable capabilities, learned structure,
memories, skills and learning procedures requires suitable observation and action interfaces and, where needed, further training.

\paragraph{Phase II: online real-to-sim scaling.} After Milestone 2, \textbf{virtual-to-real cold start}, we propose that the embodied system feeds real-world data back online.
Its development then becomes a loop of continual exploration in $\mathcal E_R$,
feeding the resulting data back into simulation in $\mathcal E_V$, improving itself faster through simulation, and thereby exploring
an ever more open real world. This phase leads to Milestone 3, \textbf{parallel virtual--real improvement}.

The rest of this survey reviews the parts of this framework:
\Cref{sec:sim} the virtual and real worlds in which the pathway
would run, \Cref{sec:embodied-agent} how the agent learns and acts in them, and \Cref{sec:selfimprove} the improvement loop that would drive it. Existing work provides mechanisms and partial loops
across these activities. EVA uses their connections to organise a potential pathway toward general-purpose physical agency, rather
than assuming that environmental expansion or repeated training automatically produces it, and \Cref{sec:discussion} returns to what
the pathway still needs.

\FloatBarrier

\needspace{16\baselineskip}
\section{Environment Construction}
\label{sec:sim}

In the pathway proposed by EVA (\Cref{sec:vrea}), an agent practises and improves in a virtual environment before gradually entering the real world and continuing there.
This section covers how the world in which a robot practises and is tested is prepared.
The real world is where the robot is ultimately meant to work, and \Cref{sec:sim-reset} covers automating its preparation and reset.
Building simulated environments is the main route to training models and letting them practise at scale (\Cref{sec:sim-construct}).
Neural simulators are a potential alternative to written simulation that may offer more diversity but are currently less controllable (\Cref{sec:sim-neural}).
\Cref{sec:learning-tasks} covers tasks and curricula in these worlds, and \Cref{sec:sim-bench} covers benchmarks that fix the world, tasks, interface, reset procedure and scoring rule for comparing robots.
Revising a virtual world from the agent's own experience is reviewed in \Cref{sec:si-environment}.

\subsection{Real-World Preparation, Reset and Reset-Free Practice}
\label{sec:sim-reset}

\begin{takeawaybox}
A general model can prepare the next real-robot trial by planning each reset from the current scene or by writing reset code
beforehand: the first can respond to unanticipated states, whereas the second is cheaper to run. Checking the initial state catches failed
resets, so that training or evaluation does not start from the wrong condition.
\end{takeawaybox}

\begin{figure}[H]
    \centering
    \includegraphics[width=\linewidth]{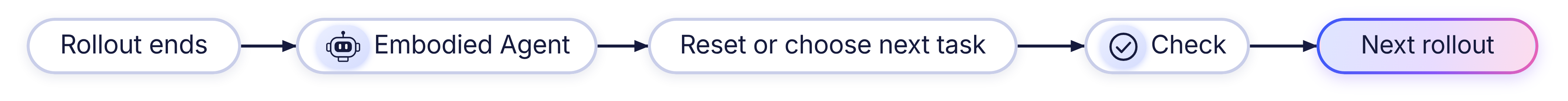}
    \caption{\textbf{Real-world preparation and reset.} After a rollout, the embodied agent either resets the scene, by planning each reset, by code written in advance or by a reset policy, or chooses a next task that the scene allows; a check of the starting state catches failures before the next rollout.}
    \label{fig:subfig-sim-reset}
\end{figure}

A reset restores a real robot station to the initial condition of the next rollout and confirms that this condition holds. Automating
it matters because real-robot training and evaluation need many repeated trials and resets done by people are costly, and because a
failed reset that goes unnoticed lets the next rollout start from an unintended state and enter training or evaluation under the wrong
label. A general model can take part in four ways, which differ in when the reset is decided: it can plan each reset from the current
scene, write the reset in advance, pair each task with its own reset policy, or avoid resets altogether. HALTER~\citep{jiang2026rollout}
plans each reset. While the rollout runs, it records the initial condition as a scene graph of relations between objects, keeping
objects that later become hidden; afterwards a language model reads the graph, chains skills from a library of reusable reset skills,
each learned as a policy, and checks after each skill and at the end that the scene matches, calling a person when a check fails.
Because the skills are reused, its resets carry over to held-out tasks where a per-task reset policy rarely succeeds, although every
reset still consists of skills already learned. ENPIRE~\citep{xiao2026enpire}, reviewed under policy training
(\Cref{sec:learning-frameworks}), and Zero2Skill~\citep{wang2026zero2skill} write the reset in advance. In ENPIRE, coding agents write
reset functions with safety limits and a reward before training starts, and the functions stay fixed while it runs; in Zero2Skill, the
model writes a reset routine of tool calls and a reset criterion once, and after each reset a vision--language model judges a fresh
image against the criterion, retrying before it calls an operator. Written code is cheap to run but covers only the scenes it
anticipates, whereas planning handles unanticipated scenes at the cost of a model call and a check at every reset.
RoboClaw~\citep{li2026roboclaw} instead pairs each subtask with a reset policy, trained on a reset task made easier than the forward
one, and runs it once the subtask is complete. FIND~\citep{fang2026find} and Tether~\citep{liang2026tether} avoid resets: after each
rollout, a vision--language or embodied-reasoning model reads the scene and chooses, from a fixed task library, a next task that the
scene allows. This works only when each end state starts another task, and people still recover objects that are dropped or
overturned.

\subsection{Simulated Environment Construction}
\label{sec:sim-construct}

\begin{takeawaybox}
The model builds assets and adds their physical properties, assembles scenes and adds the relations between objects, and pairs each
addition with a check, from a critic's judgement to a physics simulation. Invented worlds are accepted because they are plausible;
twins are checked against their capture, mostly in appearance and geometry, and agreement with recorded behaviour remains rare.
\end{takeawaybox}

\begin{figure}[H]
    \centering
    \includegraphics[width=\linewidth]{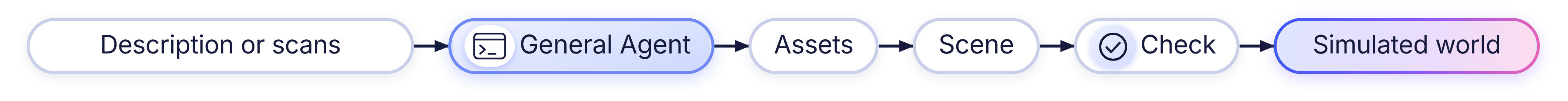}
    \caption{\textbf{Simulated environment construction.} From a description or scans, a general agent builds assets and places them in a scene, and the result is checked.}
    \label{fig:subfig-sim-construct}
\end{figure}

A simulated world can be configured, reset and run in parallel by program (\Cref{sec:def-environment}).
Building such a world has long been slow, expert work; agents now take over much of it, writing assets, scenes and tasks and checking
the result by rendering and running it.

Simulated environment construction writes such a world: from a description, scans or existing assets, the model produces $\mathcal E_V$ as a
simulator program that fixes the assets, their physical properties and the scene that relates them (\Cref{sec:roles-environment}); the
tasks set in it are reviewed in \Cref{sec:learning-tasks}. How a world is built and when it is accepted depend on what it is
answerable to. An invented world answers only to the task and is kept if critics judge it plausible and policies find it useful. A
digital twin answers to a particular real scene, object or robot and is accepted when simulation and recording agree, in appearance
and geometry and, where motion was recorded, in dynamics. Worlds in between, such as one completed from a single photograph or a
digital cousin of a real room, draw on real data without reproducing it and are judged, like invented worlds, by plausibility.
The subsection follows the order in which a world is assembled: \Cref{sec:sim-assets} covers assets, and \Cref{sec:sim-scenes}
covers scenes, invented and reconstructed.

\subsubsection{Asset Construction}
\label{sec:sim-assets}
Every simulated world, invented or twinned, is assembled from assets, and assets come in physical kinds that differ in what a simulator must be told about them. A rigid object needs geometry, a collision shape, mass and friction. Since 3D generators now supply most of the geometry for purely rigid objects~\citep{team2025sam,xiang2024structured}, as they do in SceneSmith and SAGE~\citep{pfaff2026scenesmith,xia2026sage}, the agent's task is to fit the result to observation; GATOR~\citep{wu2026gator} edits a generated object until it matches the input images. An articulated object adds parts and joints, which current generators do not produce reliably; SceneSmith, for example, generates its rigid objects but retrieves those with movable parts from a library. Most agentic work concentrates here, because such a structure is discrete and can be written as code: Articraft~\citep{zhou2026articraft} writes each asset as a program checked by tests, ArtiCAD~\citep{shui2026articad} fixes the joints before generating geometry, and Artisan~\citep{cai2026physically} adds physics checks in simulation. In these invented assets, physical values come from the model's prior knowledge or from a check against stated expectations. Twins take their structure from a particular object instead: Articulate-Anything~\citep{le2024articulate} fits parts and joints to images or video, retrieving the part meshes from a library, USDCraft~\citep{zhang2026usdcraft} fits them to scans, and FACT~\citep{shao2026fact} goes further, fitting the object's dynamics to videos of it moving. UnfoldArt~\citep{boudjoghra2026unfoldart}, working from a text prompt or a single view, fills in hidden parts with generative priors and so sits between twin and invention. A deformable object adds a material model, which describes how it bends or stretches under force and cannot be seen in a photo. DeformSmith~\citep{li2026deformsmith} and DiagGen~\citep{chen2026diaggen} therefore probe candidates in simulation and repair whatever responds wrongly; DeformSmith accepts an asset when its simulated probes behave as requested, and judges its plausibility only afterwards. Across all three kinds, geometry is routinely matched to observation, physical values are mostly assumed or checked for plausibility, and identification from a recording remains the exception.

\subsubsection{Scene Construction}
\label{sec:sim-scenes}
Assets combine into scenes, and a scene adds their relations: where each stands and what holds it up.
Invented scenes are usually built top-down. SceneSmith~\citep{pfaff2026scenesmith} passes each level, from layout to furniture to
small objects, through a designer and a critic, and MANSION~\citep{che2026mansion} extends the hierarchy to whole buildings with a
placement module that works in one pass, without a critic. SAGE~\citep{xia2026sage} checks physics inside its loop, simulating the
scene after every edit and rejecting unstable placements, and SceneSmith lets objects settle under gravity both inside some placement
tools and after each stage. PhyScensis~\citep{wang2026physcensis} instead states physics in what the model writes: spatial and physical
predicates, such as support, stacking and containment, which a solver backed by a physics engine realises as stable, densely packed
arrangements, returning failures and stability estimates for the model to revise. Scenes written as programs can also encode
behaviour, such as RoomWright's working light switches~\citep{xiao2026beyond} or the physical processes of
ChronoAgentic~\citep{wang2026chronoagentic} and Text2Sim~\citep{xiong2026text2sim}, but motion written into a program may be animation
rather than physics. Code2Worlds-2~\citep{zhang2026code2worlds} therefore adds the collision geometry, mass and joints an engine needs
and checks the scene by running it. An invented scene is kept because it is coherent and stable, not because it matches anything
real.

Reconstruction runs the other way: the capture is split into objects, an asset is generated for each and placed, and the result is
compared with the capture. LiteReality-Agent~\citep{huang2026litereality} edits an executable room script against RGB-D scans,
Lucida~\citep{qin2026lucida} places each asset with a trained vision--language policy that aligns its render with the input views, and
CoDimRecon~\citep{xie2026codimrecon} extends the procedure to deformable objects. With a single image, the hidden parts cannot be
checked, so REST3D~\citep{ma2026rest3d} and Fysiverse-3D-SimReady~\citep{wang2026fysiverse} fall back on physical stability and sit
between twin and invention, while SceneMosaic~\citep{ran2026scenemosaic} deliberately turns one image into many variants. Because they
are checked mainly against photos and scans, reconstructed scenes match how a room looks more often than how it behaves, and the two
can diverge: in clinic rooms rebuilt from photographs, shifting one device slightly can decide whether a robot arm reaches its screen,
which visual similarity does not reveal~\citep{wu2026operational}. Some systems therefore check the twin by using it. Agentic
RSR~\citep{li2026agentic} refines a scene rebuilt from a workspace video until a programmatic evaluator, not the agent, finds its depth
and colour errors against real views small enough, then has a coding agent develop the robot's policy in it, whose real success comes
close to its simulated success. DexAgent~\citep{wang2026dexagent} and SimFoundry~\citep{ranawaka2026simfoundry} rebuild twins from a
single video, letting a vision--language model choose reconstruction skills or set mass, friction and joints; policies trained in
DexAgent's twins transfer to real bimanual tasks, and SimFoundry's simulated evaluations track real success, though a person briefly
tuned its scenes. Others check behaviour against recordings directly. Video2World~\citep{tang2026video2world} compares rebuilt
environments with videos of real interaction, Agentic Real2Sim~\citep{chen2026agentic} refines a MuJoCo twin of a recorded robot
episode from replay evidence, with success in its twins tracking real success, and Vid2Sid~\citep{qiu2026vid2sid} has a
vision--language model compare simulated and real videos and keep the physics parameters that best match the real trajectory.
DexSim2Real~\citep{zeng2026dexsim2real} instead calibrates the randomisation of physics and rendering, using a vision--language model's
realism scores as the objective of an evolutionary search; replacing these scores with random ones lowers the real success of the
resulting policies. In all of these, the twin is kept because it agrees with a recording or looks real to a judge. Some environments
compute dynamics with a physics engine but render images with a neural method such as Gaussian splatting; since their dynamics come
from the engine, we treat them here rather than with neural simulation.

\begingroup\footnotesize\setlength{\tabcolsep}{5pt}\renewcommand{\arraystretch}{1.15}
\begin{xltabular}{\linewidth}{@{}>{\raggedright\arraybackslash}p{3.0cm} c >{\raggedright\arraybackslash}X >{\raggedright\arraybackslash}X >{\raggedright\arraybackslash}X c >{\raggedright\arraybackslash}X@{\hspace{2pt}}}
\caption{\textbf{Comparison of works on simulated environment construction} (\Cref{sec:sim-construct}). Builds: Asset denotes an individual asset; Scene (Invent.) denotes an invented or randomised scene; Scene (Recon.) denotes a scene reconstructed from real observations. Physics: the pipeline checks the physical validity of what it builds in a physics engine (\yes), by geometric collision or clearance tests only (Geom.), through physics checks used only to evaluate the finished output (Eval.), or not at all (\no). Checked against: a critic model, human raters, a reference asset, the appearance or the dynamics of a recording, or a policy trained or tested in the world.}\label{tab:cmp-3-2}\\
\toprule\rowcolor{cmphead} Work & Link & Input & Model writes & Builds & Physics & Checked against \\ \midrule \endfirsthead
\toprule\rowcolor{cmphead} Work & Link & Input & Model writes & Builds & Physics & Checked against \\ \midrule \endhead
\bottomrule \endlastfoot
GATOR~\citeyearpar{wu2026gator} & \paperlink{https://arxiv.org/abs/2610.11215}\nocode & Image & Edits & Asset & \no & Appearance \\
\rowcolor{cmpstripe}Articulate-Anything~\citeyearpar{le2024articulate} & \paperlink{https://arxiv.org/abs/2410.13882}\nocode & Image, video & Code & Asset & Geom. & Critic, appearance, policy \\
ArtiCAD~\citeyearpar{shui2026articad} & \paperlink{https://arxiv.org/abs/2604.10992}\nocode & Text & Code & Asset & \no & Critic, reference \\
\rowcolor{cmpstripe}Articraft~\citeyearpar{zhou2026articraft} & \paperlink{https://arxiv.org/abs/2605.15187}\nocode & Text & Code & Asset & Geom. & Human, downstream model \\
Artisan~\citeyearpar{cai2026physically} & \paperlink{https://arxiv.org/abs/2610.13947}\nocode & Text & Code & Asset & \yes & Critic, human, policy \\
\rowcolor{cmpstripe}FACT~\citeyearpar{shao2026fact} & \paperlink{https://arxiv.org/abs/2609.37067}\nocode & Image, video & Code & Asset & \yes & Appearance, dynamics \\
UnfoldArt~\citeyearpar{boudjoghra2026unfoldart} & \paperlink{https://arxiv.org/abs/2606.30608}\nocode & Text, image & Parameters & Asset & \no & Critic, reference \\
\rowcolor{cmpstripe}USDCraft~\citeyearpar{zhang2026usdcraft} & \paperlink{https://arxiv.org/abs/2610.11322}\nocode & Mesh & Code & Asset & Geom. & Appearance, policy \\
DeformSmith~\citeyearpar{li2026deformsmith} & \paperlink{https://arxiv.org/abs/2609.18620}\nocode & Text & Parameters & Asset & \yes & Critic, human \\
\rowcolor{cmpstripe}DiagGen~\citeyearpar{chen2026diaggen} & \paperlink{https://arxiv.org/abs/2609.23103}\nocode & Image & Parameters & Asset & \yes & Critic \\
MANSION~\citeyearpar{che2026mansion} & \paperlink{https://arxiv.org/abs/2603.11554}\nocode & Text & Predicates & Scene (Invent.) &\yes & Human \\
\rowcolor{cmpstripe}PhyScensis~\citeyearpar{wang2026physcensis} & \paperlink{https://arxiv.org/abs/2602.14968}\nocode & Text & Predicates & Scene (Invent.) &\yes & Critic, human, policy \\
SAGE~\citeyearpar{xia2026sage} & \paperlink{https://arxiv.org/abs/2602.10116}\nocode & Text & Tool calls & Scene (Invent.) &\yes & Critic, policy \\
\rowcolor{cmpstripe}SceneSmith~\citeyearpar{pfaff2026scenesmith} & \paperlink{https://arxiv.org/abs/2602.09153}\nocode & Text & Tool calls & Scene (Invent.) &\yes & Critic, human, policy \\
ChronoAgentic~\citeyearpar{wang2026chronoagentic} & \paperlink{https://arxiv.org/abs/2605.14398}\nocode & Text & Code & Scene (Invent.) &\yes & Critic \\
\rowcolor{cmpstripe}Code2Worlds-2~\citeyearpar{zhang2026code2worlds} & \paperlink{https://arxiv.org/abs/2610.17651}\codelink{https://github.com/AIGeeksGroup/Code2Worlds-2} & Text & Code & Scene (Invent.) &\yes & Simulator checks \\
RoomWright~\citeyearpar{xiao2026beyond} & \paperlink{https://arxiv.org/abs/2608.18840}\nocode & Text & Code & Scene (Invent.) &\yes & Critic \\
\rowcolor{cmpstripe}Text2Sim~\citeyearpar{xiong2026text2sim} & \paperlink{https://arxiv.org/abs/2609.36593}\nocode & Text & Code & Scene (Invent.) &\yes & Critic, human \\
Agentic RSR~\citeyearpar{li2026agentic} & \paperlink{https://arxiv.org/abs/2610.10479}\nocode & Video & Code & Scene (Recon.) &\yes & Appearance, policy \\
\rowcolor{cmpstripe}CoDimRecon~\citeyearpar{xie2026codimrecon} & \paperlink{https://arxiv.org/abs/2609.36024}\nocode & Scan & Code & Scene (Recon.) &\yes & Appearance \\
LiteReality-Agent~\citeyearpar{huang2026litereality} & \paperlink{https://arxiv.org/abs/2610.01863}\nocode & Scan & Code & Scene (Recon.) &Geom. & Appearance, critic, human \\
\rowcolor{cmpstripe}Lucida~\citeyearpar{qin2026lucida} & \paperlink{https://arxiv.org/abs/2608.30821}\nocode & Video & Placements & Scene (Recon.) &\no & Appearance \\
Video2World~\citeyearpar{tang2026video2world} & \paperlink{https://arxiv.org/abs/2610.04432}\codelink{https://github.com/AetherLabsAI/Video2World} & Video & Code & Scene (Recon.) &Eval. & Appearance, dynamics, human \\
\rowcolor{cmpstripe}Agentic Real2Sim~\citeyearpar{chen2026agentic} & \paperlink{https://arxiv.org/abs/2607.19190}\nocode & Video & Placements & Scene (Recon.) &\yes & Dynamics, critic, policy \\
Vid2Sid~\citeyearpar{qiu2026vid2sid} & \paperlink{https://arxiv.org/abs/2602.19359}\nocode & Video & Parameters & Scene (Recon.) &\yes & Dynamics \\
\rowcolor{cmpstripe}DexAgent~\citeyearpar{wang2026dexagent} & \paperlink{https://arxiv.org/abs/2609.35318}\nocode & Video & Tool calls & Scene (Recon.) &\yes & Appearance, policy \\
SimFoundry~\citeyearpar{ranawaka2026simfoundry} & \paperlink{https://arxiv.org/abs/2606.28276}\nocode & Video & Parameters & Scene (Recon.) &\yes & Critic, policy \\
\rowcolor{cmpstripe}DexSim2Real~\citeyearpar{zeng2026dexsim2real} & \paperlink{https://arxiv.org/abs/2605.05241}\nocode & Rendered frames & Scores & Scene (Invent.) &\no & Critic, policy \\
Fysiverse-3D-SimReady~\citeyearpar{wang2026fysiverse} & \paperlink{https://arxiv.org/abs/2609.31715}\codelink{https://github.com/Fysics-AI/Fysiverse-3D-SimReady} & Image & Predicates & Scene (Recon.) &\yes & Appearance, human \\
\rowcolor{cmpstripe}REST3D~\citeyearpar{ma2026rest3d} & \paperlink{https://arxiv.org/abs/2605.30338}\nocode & Image & Predicates & Scene (Recon.) &\yes & Appearance \\
SceneMosaic~\citeyearpar{ran2026scenemosaic} & \paperlink{https://arxiv.org/abs/2609.05594}\codelink{https://github.com/rxjfighting/SceneMosaic} & Image & Placements & Scene (Invent.) &\yes & Critic, human \\
\end{xltabular}
\endgroup

\subsection{Neural Simulation Environment Construction}
\label{sec:sim-neural}

\begin{takeawaybox}
A neural simulator is built mainly from data and training, so it can simulate only what its data covers. To serve policy learning,
it must also follow actions reliably, keep geometry and physics consistent over long rollouts and tell success from failure. Agents
have begun to collect interaction data and verify task progress, but choosing the interactions the model still lacks remains a
possibility rather than practice.
\end{takeawaybox}

\begin{figure}[H]
    \centering
    \includegraphics[width=\linewidth]{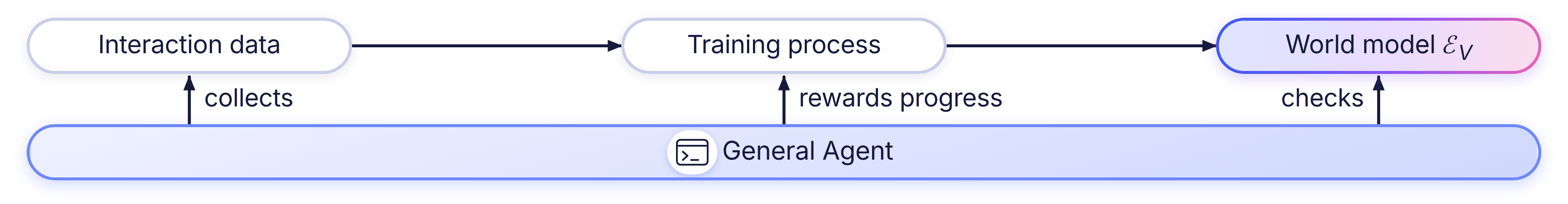}%
    \caption{\textbf{Neural simulation environment construction.} A world model is trained on interaction data. A general agent takes part at three points: it collects interaction data, checks whether a generated rollout shows task progress before it is rewarded in training, and checks whether rollouts in the world model succeed, a check the world model can also make itself.}
    \label{fig:subfig-sim-neural}
\end{figure}

A neural simulator, today usually realised as a world model, is a network that predicts the next observation from the current one and an action, so that a policy can act inside a world learned from data. It sits between a real station and a written simulation: its appearance and dynamics are learned from real recordings, yet it can be reset, replayed and run in parallel like a simulator. Its state, however, is implicit and its physics only approximate. Because such a world is learned rather than written, building it is mostly a matter of data and training, a step in which agents have so far played little part. A world model that an agent queries inside a decision to rehearse candidate actions, as in NovaPlan~\citep{fu2026novaplan}, is a tool rather than an environment and is not treated here. Typical action-conditioned world models, from UniSim~\citep{yang2023learning} and IRASim~\citep{zhu2024irasim} to Cosmos~\citep{nvidia2025cosmos}, Genie Envisioner~\citep{liao2025genie}, Ctrl-World~\citep{guo2025ctrl} and DreamDojo~\citep{gao2026dreamdojo}, are trained on large collections of robot and human video, and what they can simulate is bounded by what that data covers. Agents can widen this coverage by collecting the data themselves: in PlayWorld~\citep{yin2026playworld}, a vision--language model proposes tasks and keeps objects within reach while a policy carries them out, so that interaction data accumulate with only occasional human monitoring and resets. Beyond collecting data, an agent could also decide which interactions the world model still lacks. A world model that looks right, however, is not yet a simulator. It must follow actions beyond its training data, keep geometry and physics consistent over long rollouts and tell success from failure. Dedicated benchmarks now measure these properties~\citep{shang2026worldarena,co2026worldsimprobe}, and targeted retraining repairs them~\citep{bardhan2026persistent,zhang2026physisforcing}. Whether a task succeeds inside such a simulator is usually decided by the world model itself~\citep{tang2025roboscape} or by a vision--language judge~\citep{quevedo2025worldgym}. The same judgement is needed when the world model is trained with reinforcement learning, and there an agent has begun to take it over: in Reward as an Agent~\citep{li2026reward}, it checks whether a generated rollout shows real task progress before granting a reward.

\subsection{Task Generation and Curriculum Design}
\label{sec:learning-tasks}

\begin{takeawaybox}
Models can generate practice tasks and adapt curricula by selecting or revising tasks in response to the policy learner's record. Task validity must be checked separately from learning success: failure may reflect an invalid task or one the policy has not yet learned. Where a fixed learning goal is intended, existing checks do not establish whether evolving tasks and curricula remain aligned with it.
\end{takeawaybox}

\begin{figure}[H]
    \centering
    \includegraphics[width=\linewidth]{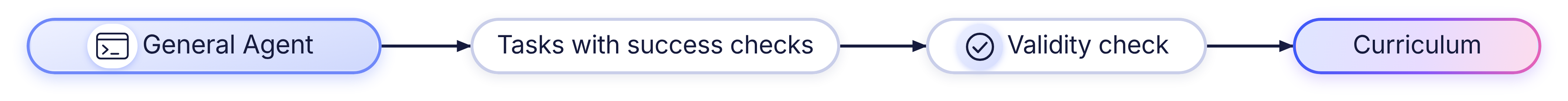}
    \caption{\textbf{Task generation and curriculum design.} A general agent writes tasks, each with a goal and a success check; a validity check tests them before any learning, and the agent orders the valid tasks into a curriculum, laid out in advance or revised as learning progresses.}
    \label{fig:subfig-sim-tasks}
\end{figure}

A task consists of a goal and a criterion for success within a given world: the task $z$ produced by the tasks and
curricula branch (\Cref{sec:roles-environment}). It is the step of environment construction that
follows the world: \Crefrange{sec:sim-reset}{sec:sim-neural} provide the world, and the task decides what the policy practises in
it, while its goal and success check determine what is learned and how learning is judged. In task generation, a general-purpose
vision--language model proposes new tasks, usually as simulator code that builds the scene and sets the goal and success check, and a
separate policy is trained on these tasks or on their demonstrations. In curriculum design, the model also orders the tasks, in stages
laid out in advance or by choosing or revising which task comes next as learning progresses. A generated task has to pass three tests: it must be valid (the scene can be built, the goal
can be reached and the success check measures the goal), learnable by the current policy, and relevant to the capability the system
is meant to acquire. Validity can be checked before any learning; learnability and relevance need the learner's record over repeated
attempts. This distinction structures the subsection: \Cref{sec:learning-tasks-generation} covers where tasks come from and how their
validity is checked, and \Cref{sec:learning-tasks-curriculum} covers learnability and relevance.

\subsubsection{Task Generation}
\label{sec:learning-tasks-generation}
Systems differ first in where the task comes from and how much of the world is built with it. GenSim~\citep{wang2023gensim} writes
new simulator tasks, with their scenes, goals and scripted demonstrations, from earlier tasks in a library. GenSim2~\citep{hua2024gensim2} extends this to articulated objects and long-horizon tasks, with a multimodal model writing the constraints from which a keypoint-based motion planner produces each demonstration. From
RoboGen~\citep{wang2023robogen} on, the task usually comes first and the world is built around it: from a sampled object, RoboGen
proposes a task, arranges the assets it requires, writes the supervision and learns the skill. SimWorld
Studio~\citep{kang2026simworld}, EmbodiedGen V2~\citep{wang2026embodiedgen} and Genie Sim 3.0~\citep{yin2026genie} invent the task
together with the world; EmbodiedGen V2 reports that most such worlds are usable without manual edits, and Genie Sim 3.0 builds them
from a request in natural language over a library of simulation-ready assets, with a vision--language model scoring the policies run
in its benchmark. Video2Policy~\citep{ye2025video2policy} and ARSTAG~\citep{li2026arstag}
instead take the task from reality, from a human video or from a photograph and an instruction, and rebuild only what the task
involves: the task is real, the world around it only plausible. Scene2Demo~\citep{liu2026scene2demo} instead reconstructs the whole scene from a single photograph. Real2Gym~\citep{ren2026real2gym} goes further, replaying the
demonstrated actions in physics and comparing them with the video frame by frame. HARBOR~\citep{li2026harbor}, which implements a task
given by the user inside an existing simulator, is reviewed under policy training (\Cref{sec:learning-frameworks}).
AutoRT~\citep{ahn2024autort}, the
distinctive real-world case, proposes tasks from each robot's scene across a fleet under a prompted robot constitution, yielding
episodes more diverse than those of earlier hand-designed datasets.

The first systems checked mainly validity, and mostly before learning. GenSim considers a task for its library only once its code
runs, its scripted demonstration completes the task, a policy trained on it alone succeeds at least occasionally and a person has
checked it, and the model then decides whether to add it; after adaptation on real data,
a policy pretrained on the generated tasks outperforms one pretrained on the original benchmark tasks. RoboGen's tasks still include
19 of 155 that fail because of errors in the scene or the reward. Later systems also test whether a task can be executed:
FATE~\citep{wei2026fate} checks each task first for static validity and then for executability with an embodied agent, repairing the
scene or the specification when it fails, and Scene2Demo's feedback agents revise failed action sequences. Such checks catch code
that does not run, a task that does not fit its scene or a goal that a scripted demonstration cannot reach. They do not guarantee that the implemented
success criterion matches the intended task: the GenSim authors note that a task whose behaviour does not match its language
instruction can still pass their syntax, runtime and task-completion checks. \citet{chen2024evaluation} therefore evaluate the generated tasks themselves: their quality, as judged by language and vision--language models from the rendered scene and trajectory, their diversity, and the generalisation of policies trained on them; applied to GenSim, RoboGen and BBSEA, no pipeline leads on every axis. Agentic Skill Discovery~\citep{zhao2024agentic} and
CurricuLLM~\citep{ryu2024curricullm} add checks on trained behaviour.

\subsubsection{Curriculum Design}
\label{sec:learning-tasks-curriculum}
A curriculum organises the sequence or distribution of tasks that a policy practises. It may be laid out in advance as training stages, or adapted to the
learner's record, by changing how existing tasks are sampled and ordered or by revising and generating tasks. Where the curriculum
adapts, a failure alone does not say whether a task is invalid or merely hard, and revising a task raises the question of whether its
objective survives. With the world fixed, CurricuLLM lays out its stages in advance: a language model first writes a sequence of
subtasks in natural language, then for each stage generates several candidate task codes with their rewards and goals, trains a policy
on each and, judging their rollouts, keeps the policy that best matches the subtask to start the next stage. RATs~\citep{zhang2026playful} lets a coding agent
propose tasks that are novel yet learnable. FIND~\citep{fang2026find} selects rather than writes, from a hand-designed library of tasks
that the scene left by the last rollout allows (\Cref{sec:sim-reset}); its main contribution, a policy that keeps improving on the real
robot, is reviewed with policy model improvement (\Cref{sec:si-model}). With the world regenerated, SimWorld Studio raises or lowers the difficulty of new environments according to the learner's
success, Eurekaverse~\citep{liang2024eurekaverse} evolves the terrain programs that trained the best policy from that policy's training statistics, and
WEDGE~\citep{zhang2026failures} builds environments around the current policy's failures, keeping those that are hard but still
solvable. OMNI-EPIC~\citep{faldor2024omni} has no fixed target: learned tasks become stepping stones, failed ones are revised and
retried, and a model of what people find interesting admits new ones. Starting each policy from an earlier one markedly raises success
in simulation, but its success checks, written as code by a language model, cannot tell a task made impossible by its environment from
one not yet learned, so both end in the same archive. A revision may also relax the objective: EmbodiedSmith~\citep{qin2026embodiedsmith}
revises a failed task's skill bindings or requests targeted scene edits after failures of its fixed executor rather than of a learner,
and without these revisions the success of its generated demonstrations
collapses, and although it keeps the success criteria fixed, a revised scene can turn the task into an easier variant, and the scene
edits are not ablated. The test thus moves from the world to the learner: a task is kept if it is feasible now and useful next, and
its target is judged by a model's notion of interest, fixed by hand, or fixed only in its success criteria.

\begingroup\footnotesize\setlength{\tabcolsep}{5pt}\renewcommand{\arraystretch}{1.15}
\begin{xltabular}{\linewidth}{@{}>{\raggedright\arraybackslash}p{3.1cm} c >{\raggedright\arraybackslash}X >{\raggedright\arraybackslash}X >{\raggedright\arraybackslash}X >{\raggedright\arraybackslash}X c c c@{\hspace{2pt}}}
\caption{\textbf{Comparison of works on task generation and curriculum design} (\Cref{sec:learning-tasks}). Valid, Learnable, Relevant: whether the system itself checks, for each generated task, the three tests of \Cref{sec:learning-tasks} (\yes), only part of one or only indirectly (Part.), or not at all (\no); checks left to prompting, construction or human evaluation count as \no.}\label{tab:cmp-3-4}\\
\toprule\rowcolor{cmphead} Work & Link & Task from & World built & Task accepted by & Curriculum driven by & Valid & Learn. & Relev. \\ \midrule \endfirsthead
\toprule\rowcolor{cmphead} Work & Link & Task from & World built & Task accepted by & Curriculum driven by & Valid & Learn. & Relev. \\ \midrule \endhead
\bottomrule \endlastfoot
GenSim~\citeyearpar{wang2023gensim} & \paperlink{https://arxiv.org/abs/2310.01361}\codelink{https://github.com/liruiw/GenSim} & Library & Task-relevant & Code checks, demo, training, human & -- & \yes & \yes & \no \\
\rowcolor{cmpstripe}RoboGen~\citeyearpar{wang2023robogen} & \paperlink{https://arxiv.org/abs/2311.01455}\nocode & Model & Full & Model judge (assets) & -- & Part. & \no & \no \\
EmbodiedGen V2~\citeyearpar{wang2026embodiedgen} & \paperlink{https://arxiv.org/abs/2607.07459}\codelink{https://github.com/HorizonRobotics/EmbodiedGen} & Model & Full & Model judge & -- & Part. & \no & \no \\
\rowcolor{cmpstripe}FATE~\citeyearpar{wei2026fate} & \paperlink{https://arxiv.org/abs/2603.01505}\nocode & Model & Full & Model judge, executor & Executor failures & \yes & \yes & \no \\
Genie Sim 3.0~\citeyearpar{yin2026genie} & \paperlink{https://arxiv.org/abs/2601.02078}\nocode & Request, model & Full & Demo succeeds & -- & \yes & \no & \no \\
\rowcolor{cmpstripe}GenSim2~\citeyearpar{hua2024gensim2} & \paperlink{https://arxiv.org/abs/2410.03645}\nocode & Library & Task-relevant & Demo succeeds & -- & \yes & \no & \no \\
Video2Policy~\citeyearpar{ye2025video2policy} & \paperlink{https://arxiv.org/abs/2502.09886}\nocode & Human video & Task-relevant & Model judge & -- & Part. & \no & \no \\
\rowcolor{cmpstripe}ARSTAG~\citeyearpar{li2026arstag} & \paperlink{https://arxiv.org/abs/2609.24563}\nocode & Image & Task-relevant & Demo succeeds & -- & \yes & \no & \no \\
Real2Gym~\citeyearpar{ren2026real2gym} & \paperlink{https://arxiv.org/abs/2609.37089}\nocode & Human video & Full & Demo succeeds & -- & \yes & \no & \no \\
\rowcolor{cmpstripe}Scene2Demo~\citeyearpar{liu2026scene2demo} & \paperlink{https://arxiv.org/abs/2602.12065}\nocode & Image & Full & Demo succeeds, model judge & -- & \yes & \no & \no \\
AutoRT~\citeyearpar{ahn2024autort} & \paperlink{https://arxiv.org/abs/2401.12963}\nocode & Robot scene & None & Model judge & -- & \yes & \no & \no \\
\rowcolor{cmpstripe}Agentic Skill Discovery~\citeyearpar{zhao2024agentic} & \paperlink{https://arxiv.org/abs/2405.15019}\nocode & Model & None & Trained behaviour & Success rate & \yes & \yes & \no \\
CurricuLLM~\citeyearpar{ryu2024curricullm} & \paperlink{https://arxiv.org/abs/2409.18382}\nocode & Model & None & Trained behaviour & Training stats & \yes & Part. & \no \\
\rowcolor{cmpstripe}RATs~\citeyearpar{zhang2026playful} & \paperlink{https://arxiv.org/abs/2606.19419}\nocode & Model & Task-relevant & Code checks & Success rate, novelty & \yes & \yes & \no \\
Eurekaverse~\citeyearpar{liang2024eurekaverse} & \paperlink{https://arxiv.org/abs/2411.01775}\nocode & Model & Full & Code checks & Training stats & \yes & \yes & \no \\
\rowcolor{cmpstripe}OMNI-EPIC~\citeyearpar{faldor2024omni} & \paperlink{https://arxiv.org/abs/2405.15568}\codelink{https://github.com/maxencefaldor/omni-epic} & Model & Full & Code checks, model judge & Success rate & Part. & \yes & \no \\
EmbodiedSmith~\citeyearpar{qin2026embodiedsmith} & \paperlink{https://arxiv.org/abs/2610.07969}\nocode & Request & Full & Code checks & Executor failures & \yes & \no & \no \\
\rowcolor{cmpstripe}SimWorld Studio~\citeyearpar{kang2026simworld} & \paperlink{https://arxiv.org/abs/2605.09423}\codelink{https://github.com/SimWorld-AI/SimWorld-Studio} & Model & Full & Code checks, model judge & Success rate & \yes & \yes & \no \\
WEDGE~\citeyearpar{zhang2026failures} & \paperlink{https://arxiv.org/abs/2610.19420}\codelink{https://github.com/zhangyuejoslin/WEDGE} & Library & Full & Code checks, learner rollouts & Failures & \yes & \yes & Part. \\
\end{xltabular}
\endgroup

\subsection{Evaluation and Benchmarks}
\label{sec:sim-bench}

\begin{takeawaybox}
A general model can turn an evaluation intent into executable tasks and verifiers, and benchmarks now test the decision layer
above a policy. A simulated or world-model score stands for real performance only once it agrees with real trials.
\end{takeawaybox}

A task in \Cref{sec:learning-tasks} is a goal with a success check that a policy practises; a benchmark fixes tasks together with the
interface, the reset procedure and the scoring rule under which a robot's ability is measured, so the model now sits the exam.
Rankings and claims of capability are decided by the test, so a flaw in it passes into every conclusion drawn from it. A real-world benchmark measures performance directly but needs hardware, resets and scoring for every trial. \Cref{sec:sim-bench-eval} covers
how tests are built and whether their scores can be trusted, and \Cref{sec:sim-bench-suites} what benchmarks test and how those of
systems that plan, write code or coordinate lower-level policies differ from those of learned policies.

\subsubsection{Evaluation}
\label{sec:sim-bench-eval}

\paragraph{Tests built by a general-purpose model.}
Building a test is environment construction applied to measurement, but whereas a practice task need only be feasible and useful next
(\Cref{sec:learning-tasks}), a test must also be fixed, comparable and discriminating. The first consideration is validity: the code
runs and the goals can be reached. RoboPlayground~\citep{wang2026roboplayground} compiles a user's plain-language description into a
validated task schema and executable MuJoCo code, admitting a task only after its goal state survives settling and a support check,
and Embodied-BenchForge~\citep{jiang2026embodied} turns a stated evaluation intent into a benchmark, with prompted language models
composing reusable skills that produce questions, tasks and executable verifiers. PARTNR's household tasks and their evaluation
functions were likewise written by language models and filtered by people~\citep{chang2024partnr}. The second is traceability and
repair: Embodied-BenchForge records each artefact in a dependency graph and re-runs or rolls back only the parts that fail
verification, and RoboPlayground turns later edits into new variants. The third is discrimination: Embodied-BenchForge produced six
offline question-answering benchmarks and one interactive AI2-THOR benchmark of 220 executable tasks that separate systems by
capability~\citep{jiang2026embodied}, and RoboPlayground's variants probe six learned policies~\citep{wang2026roboplayground}. The
fourth is agreement with real robots, which none of these tests yet checks. The fifth is independence: a model that builds a test
and resembles the one being tested may share its blind spots (\Cref{sec:sim-limits}).

\paragraph{Agreement with real trials.}
A simulated or world-model benchmark stands in for real trials and is trusted to the extent that its results match them.
SIMPLER~\citep{li2024evaluating} reduces control and visual gaps without building full twins, and its per-task success rates closely
follow real ones. PolaRiS~\citep{jain2025polaris}, X2Real~\citep{ruan2026x2real} and R2S-Eval~\citep{wang2026r2s} go further and
rebuild or calibrate the real setting: PolaRiS reconstructs evaluation scenes from short video scans, X2Real calibrates visual and
physical properties to the hardware, and R2S-Eval calibrates a simulator to a dual-arm setup and ranks policies by a vision--language
model's pairwise preferences over their rollouts. All three report close agreement with real results, as does
RoboLab~\citep{yang2026robolab}, which uses hand-curated tasks and orders policies as the real RoboArena does. These comparisons
differ in scope: PolaRiS fine-tunes each policy briefly in simulation before comparing, X2Real compares a single policy, and
RobotArena$\infty$~\citep{jangir2025robotarena}, which turns robot-dataset videos into simulated twins, compares one task. Agreement
may also be partial. The Colosseum~\citep{pumacay2024colosseum,morgan2026colosseum} finds its perturbation results partly aligned
with the same perturbations on real robots, and ManiGuard~\citep{peng2026maniguard} finds that simulated success follows real success
while violation rates do not. World models used as evaluators, such as WorldGym~\citep{quevedo2025worldgym} and
GigaWorld-1~\citep{team2026gigaworld}, face the same test. Per-task success inside WorldGym's action-conditioned video model
correlates with success on real robots, and RoboCoach~\citep{liu2026robocoach}, which runs skill experts in such a model to choose
which subtasks need new demonstrations, ranks task--policy pairs largely as real trials do.

\subsubsection{Benchmarks}
\label{sec:sim-bench-suites}

\paragraph{Benchmarks of learned policies.}
Policy benchmarks, the baseline for comparison, include real-world ones: RoboArena~\citep{atreya2025roboarena} aggregates double-blind pairwise comparisons from evaluators at several sites, and
RoboChallenge~\citep{yakefu2025robochallenge} runs many policies on a shared set of real tasks.
FurnitureBench~\citep{heo2023furniturebench}, HomeRobot~\citep{yenamandra2023homerobot}, GarmentLab~\citep{lu2024garmentlab},
RoboTwin~\citep{mu2025robotwin} and RoboDojo~\citep{chen2026robodojo} offer both a simulated and a real suite. Simulated benchmarks are mostly accepted on the strength of their design, from LIBERO~\citep{liu2023libero} and its
variants~\citep{zhou2025libero,cui2026libero,liu2026libero,qian2026libero} to BEHAVIOR-1K~\citep{li2024behavior},
RoboCasa~\citep{nasiriany2024robocasa,nasiriany2026robocasa365}, InternData-A1~\citep{tian2025interndata}, RoboTwin
2.0~\citep{chen2025robotwin}, MobileManiBench~\citep{wang2026mobilemanibench}, SIMPLE~\citep{wei2026simple},
VLA-Arena~\citep{zhang2025vla}, ActiveArena~\citep{li2026activearena}, RoboMemArena~\citep{lei2026robomemarena},
MEMOBench~\citep{sun2026memobench} and ForceBench~\citep{han2026forcebench}; several report real trials, or claim agreement, without
measuring it, and the few that measure it appear in \Cref{sec:sim-bench-eval}.

\paragraph{Benchmarks of agent-driven robots.}
A growing set of benchmarks tests not a policy but the decision layer above it. PARTNR poses household tasks that several agents must
plan together, and EmbodiedBench~\citep{yang2025embodiedbench} also tests planning; CaP-X~\citep{fu2026cap} and
EmbodiedSWE~\citep{shen2026embodiedswe} test programming, LADDER~\citep{petkov2026ladder} a planner's orchestration of frozen
policies, and KinDER~\citep{huang2026kinder} the physical reasoning these decisions depend on, against planning, reinforcement and
imitation baselines; RobotWorld~\citep{yang2026robotworld} spans manipulation, locomotion, driving and
aerial control with interaction budgets and executable success checks. These benchmarks differ from policy benchmarks first in their
protocol: a score depends on what the agent may observe, how abstract its actions are (EmbodiedBench) or at what level the robot API
sits (CaP-X, which also varies the readable state and the visual grounding), how often it receives feedback and how many interactions
it is allowed, so results are comparable only under the same protocol. They also test long horizons and iterative revision: LADDER
scales tasks from one to ten subgoals~\citep{petkov2026ladder}, and EmbodiedSWE gives coding agents contact-rich and long-horizon
dexterous tasks to solve by revising their programs. None reports agreement with the same systems on real robots, unlike the minority of policy benchmarks in
\Cref{sec:sim-bench-eval}.

\begingroup\small\setlength{\tabcolsep}{5pt}\renewcommand{\arraystretch}{1.15}
\begin{xltabular}{\linewidth}{@{}>{\raggedright\arraybackslash}p{3.7cm} c >{\hsize=.7\hsize\raggedright\arraybackslash}X >{\hsize=.75\hsize\raggedright\arraybackslash}X >{\hsize=.8\hsize\raggedright\arraybackslash}X >{\hsize=1.75\hsize\raggedright\arraybackslash}X@{\hspace{2pt}}}
\caption{\textbf{Comparison of works on evaluation and benchmarks} (\Cref{sec:sim-bench}): every evaluation work and benchmark introduced in the subsection, from tests built by an agent and works that measure agreement with real trials to benchmarks of learned policies and of agent-driven robots. Tests: a learned policy, an agent in which a general-purpose model decides, or both. Built by: who writes the tasks and checks, the first named doing most of the work. World: where the tested system is run; Sim + real also covers a real suite that is not matched task by task with the simulated one. Engines lists the simulator used for simulated tests, with its underlying platform in parentheses. Works that use more than two simulators are marked Multiple, and works that test only on real robots or in a world model are marked with a ``--''.}\label{tab:cmp-3-5}\\
\toprule\rowcolor{cmphead} Work & Link & Tests & Built by & World & Engines \\ \midrule \endfirsthead
\toprule\rowcolor{cmphead} Work & Link & Tests & Built by & World & Engines \\ \midrule \endhead
\bottomrule \endlastfoot
RoboPlayground~\citeyearpar{wang2026roboplayground} & \paperlink{https://arxiv.org/abs/2604.05226}\nocode & Policy & Agent & Sim & MuJoCo \\
\rowcolor{cmpstripe}Embodied-BenchForge~\citeyearpar{jiang2026embodied} & \paperlink{https://arxiv.org/abs/2609.13082}\nocode & Agent & Agent & Sim & Multiple \\
PARTNR~\citeyearpar{chang2024partnr} & \paperlink{https://arxiv.org/abs/2411.00081}\codelink{https://github.com/facebookresearch/partnr-planner} & Agent & Agent, people & Sim & Habitat 3.0 \\
\rowcolor{cmpstripe}SIMPLER~\citeyearpar{li2024evaluating} & \paperlink{https://arxiv.org/abs/2405.05941}\nocode & Policy & People & Sim & SAPIEN, Isaac Sim \\
The Colosseum~\citeyearpar{pumacay2024colosseum} & \paperlink{https://arxiv.org/abs/2402.08191}\nocode & Policy & People & Sim + real & CoppeliaSim \\
\rowcolor{cmpstripe}Colosseum V2~\citeyearpar{morgan2026colosseum} & \paperlink{https://arxiv.org/abs/2605.27759}\nocode & Policy & People & Sim + real & ManiSkill3 (SAPIEN) \\
PolaRiS~\citeyearpar{jain2025polaris} & \paperlink{https://arxiv.org/abs/2512.16881}\nocode & Policy & People & Sim + real & Isaac Sim \\
\rowcolor{cmpstripe}X2Real~\citeyearpar{ruan2026x2real} & \paperlink{https://arxiv.org/abs/2609.27449}\codelink{https://github.com/X-Square-Robot/x2real} & Both & People, agent & Sim + real & Mana (Isaac Lab-Arena) \\
R2S-Eval~\citeyearpar{wang2026r2s} & \paperlink{https://arxiv.org/abs/2609.03276}\nocode & Policy & People & Sim + real & Isaac Lab-Arena (Isaac Sim) \\
\rowcolor{cmpstripe}RoboLab~\citeyearpar{yang2026robolab} & \paperlink{https://arxiv.org/abs/2604.09860}\codelink{https://github.com/NVlabs/RoboLab/} & Policy & People & Sim & Isaac Lab (Isaac Sim) \\
ManiGuard~\citeyearpar{peng2026maniguard} & \paperlink{https://arxiv.org/abs/2608.17386}\codelink{https://github.com/NU-IDEAS-Lab/ManiGuard} & Policy & People & Sim + real & OmniGibson (PhysX) \\
\rowcolor{cmpstripe}RobotArena$\infty$~\citeyearpar{jangir2025robotarena} & \paperlink{https://arxiv.org/abs/2510.23571}\nocode & Policy & Agent, people & Sim & Genesis \\
WorldGym~\citeyearpar{quevedo2025worldgym} & \paperlink{https://arxiv.org/abs/2506.00613}\nocode & Policy & People & World model & -- \\
\rowcolor{cmpstripe}RoboCoach (Liu et al.)~\citeyearpar{liu2026robocoach} & \paperlink{https://arxiv.org/abs/2609.39685}\nocode & Policy & -- & World model & -- \\
GigaWorld-1~\citeyearpar{team2026gigaworld} & \paperlink{https://arxiv.org/abs/2607.02642}\nocode & World model & People & World model & -- \\
\rowcolor{cmpstripe}RoboArena~\citeyearpar{atreya2025roboarena} & \paperlink{https://arxiv.org/abs/2506.18123}\nocode & Policy & People & Real & -- \\
RoboChallenge~\citeyearpar{yakefu2025robochallenge} & \paperlink{https://arxiv.org/abs/2510.17950}\nocode & Policy & People & Real & -- \\
\rowcolor{cmpstripe}FurnitureBench~\citeyearpar{heo2023furniturebench} & \paperlink{https://arxiv.org/abs/2305.12821}\nocode & Policy & People & Sim + real & FurnitureSim (Isaac Gym) \\
HomeRobot~\citeyearpar{yenamandra2023homerobot} & \paperlink{https://arxiv.org/abs/2306.11565}\nocode & Policy & People & Sim + real & Habitat \\
\rowcolor{cmpstripe}GarmentLab~\citeyearpar{lu2024garmentlab} & \paperlink{https://arxiv.org/abs/2411.01200}\nocode & Policy & People & Sim + real & GarmentLab Engine (Isaac Sim) \\
RoboTwin~\citeyearpar{mu2025robotwin} & \paperlink{https://arxiv.org/abs/2504.13059}\nocode & Policy & People & Sim + real & ManiSkill3 (SAPIEN) \\
\rowcolor{cmpstripe}RoboDojo~\citeyearpar{chen2026robodojo} & \paperlink{https://arxiv.org/abs/2607.04434}\nocode & Policy & People & Sim + real & Isaac Lab (Isaac Sim) \\
LIBERO~\citeyearpar{liu2023libero} & \paperlink{https://arxiv.org/abs/2306.03310}\nocode & Policy & People & Sim & robosuite (MuJoCo) \\
\rowcolor{cmpstripe}LIBERO-PRO~\citeyearpar{zhou2025libero} & \paperlink{https://arxiv.org/abs/2510.03827}\nocode & Policy & -- & Sim & robosuite (MuJoCo) \\
LIBERO-Safety~\citeyearpar{cui2026libero} & \paperlink{https://arxiv.org/abs/2606.23686}\nocode & Policy & People, agent & Sim (one real case) & robosuite (MuJoCo) \\
\rowcolor{cmpstripe}LIBERO-RECOVER~\citeyearpar{liu2026libero} & \paperlink{https://arxiv.org/abs/2609.05178}\nocode & Policy & Agent, people & Sim & robosuite (MuJoCo) \\
LIBERO-VIFO~\citeyearpar{qian2026libero} & \paperlink{https://arxiv.org/abs/2608.17600}\nocode & Policy & People & Sim + real & robosuite (MuJoCo) \\
\rowcolor{cmpstripe}BEHAVIOR-1K~\citeyearpar{li2024behavior} & \paperlink{https://arxiv.org/abs/2403.09227}\nocode & Policy & People & Sim + real & OmniGibson (PhysX) \\
RoboCasa~\citeyearpar{nasiriany2024robocasa} & \paperlink{https://arxiv.org/abs/2406.02523}\nocode & Policy & Agent, people & Sim + real & robosuite (MuJoCo) \\
\rowcolor{cmpstripe}RoboCasa365~\citeyearpar{nasiriany2026robocasa365} & \paperlink{https://arxiv.org/abs/2603.04356}\nocode & Policy & Agent, people & Sim + real & robosuite (MuJoCo) \\
InternData-A1~\citeyearpar{tian2025interndata} & \paperlink{https://arxiv.org/abs/2511.16651}\nocode & Policy & People & Sim + real & Isaac Sim, SAPIEN \\
\rowcolor{cmpstripe}RoboTwin 2.0~\citeyearpar{chen2025robotwin} & \paperlink{https://arxiv.org/abs/2506.18088}\nocode & Policy & -- & Sim + real & SAPIEN \\
MobileManiBench~\citeyearpar{wang2026mobilemanibench} & \paperlink{https://arxiv.org/abs/2602.05233}\nocode & Policy & People & Sim & Isaac Sim \\
\rowcolor{cmpstripe}SIMPLE~\citeyearpar{wei2026simple} & \paperlink{https://arxiv.org/abs/2606.08278}\nocode & Policy & People & Sim + real & MuJoCo, Isaac Sim (rendering) \\
VLA-Arena~\citeyearpar{zhang2025vla} & \paperlink{https://arxiv.org/abs/2512.22539}\nocode & Policy & People & Sim + real & robosuite (MuJoCo) \\
\rowcolor{cmpstripe}ActiveArena~\citeyearpar{li2026activearena} & \paperlink{https://arxiv.org/abs/2609.24124}\nocode & Policy & People & Sim + real & RoboTwin 2.0 (SAPIEN) \\
RoboMemArena~\citeyearpar{lei2026robomemarena} & \paperlink{https://arxiv.org/abs/2605.10921}\nocode & Both & Agent, people & Sim + real & robosuite (MuJoCo) \\
\rowcolor{cmpstripe}MEMOBench~\citeyearpar{sun2026memobench} & \paperlink{https://arxiv.org/abs/2609.07047}\nocode & Policy & People & Sim & robosuite (MuJoCo) \\
ForceBench~\citeyearpar{han2026forcebench} & \paperlink{https://arxiv.org/abs/2610.19604}\nocode & Policy & People & Sim + real & Isaac Lab (Isaac Sim) \\
\rowcolor{cmpstripe}EmbodiedBench~\citeyearpar{yang2025embodiedbench} & \paperlink{https://arxiv.org/abs/2502.09560}\nocode & Agent & People & Sim & Multiple \\
CaP-X~\citeyearpar{fu2026cap} & \paperlink{https://arxiv.org/abs/2603.22435}\nocode & Agent & People & Sim + real & robosuite (MuJoCo), OmniGibson (PhysX) \\
\rowcolor{cmpstripe}EmbodiedSWE~\citeyearpar{shen2026embodiedswe} & \paperlink{https://arxiv.org/abs/2609.27308}\codelink{https://github.com/EmbodiedSWE/EmbodiedSWE} & Agent & People & Sim & Isaac Lab (PhysX, Newton) \\
LADDER~\citeyearpar{petkov2026ladder} & \paperlink{https://arxiv.org/abs/2610.18053}\nocode & Both & People & Sim + real & Multiple \\
\rowcolor{cmpstripe}KinDER~\citeyearpar{huang2026kinder} & \paperlink{https://arxiv.org/abs/2604.25788}\nocode & Both & People & Sim & Multiple \\
RobotWorld~\citeyearpar{yang2026robotworld} & \paperlink{https://arxiv.org/abs/2610.10409}\nocode & Agent & People & Sim & Multiple \\
\end{xltabular}
\endgroup

\subsection{Discussion}
\label{sec:sim-limits}

\paragraph{Limitations and open questions.} Real-world practice continues to depend on human intervention for resets, task specification, and recovery from failures, while predefined reset strategies and reset-free task sequences offer limited flexibility. In simulation construction, generated environments are often evaluated for visual or physical plausibility rather than their actual utility for policy learning, with limited validation of behavioral fidelity and real-world effectiveness. Neural simulators further struggle to generalize beyond their training data and maintain physical consistency over long horizons, while agents rarely take an active role in identifying and addressing missing interactions. Similarly, generated tasks are typically assessed for validity and, to a lesser extent, learnability, but rarely for their relevance to the capabilities that policies are intended to acquire. Finally, evaluation remains insufficiently grounded in real-world outcomes, while shared biases between generation and evaluation can allow systematic errors to go undetected. Together, these limitations highlight the significant challenges that remain in establishing autonomous and reliable simulation-driven learning pipelines.

\paragraph{EVA pathway.} Read against the pathway of \Cref{sec:vrea}, the branches of this section supply different parts of the overlap between $\mathcal E_V$ and $\mathcal E_R$. Invented scenes and assets, generated tasks and curricula, and neural simulators widen its coverage, so that more of the tasks and conditions of $\mathcal E_R$ exist in $\mathcal E_V$; twins and calibration raise its fidelity, so that where they exist, $\mathcal E_V$ predicts the real outcome; real-world preparation and reset make practice on real stations repeatable, as Milestone 2 (virtual-to-real cold start) and Phase II (online real-to-sim scaling) require; and evaluation checked against real trials supplies the agreement by which the cold start would be timed. As these branches advance, the pathway advances with them. Wider coverage lets more of the agent’s practice take place in $\mathcal E_V$, where trials are cheap, safe and resettable at scale; higher fidelity lets the competence gained there carry over to $\mathcal E_R$ with less further training; repeatable real stations let the agent keep improving after the cold start and return real outcomes that scale $\mathcal E_V$ further; and reliable agreement turns the decision of when to enter $\mathcal E_R$ from a guess into a measurement. Better environments thus enlarge the overlap on which EVA relies, bring the cold start earlier and keep the parallel improvement in both environments going.

\needspace{16\baselineskip}
\section{Embodied Agents}
\label{sec:embodied-agent}
\label{sec:learning}

\label{sec:deployment}
In the pathway that EVA proposes (\Cref{sec:vrea}), the embodied agent makes the step from virtual to real: its foundation model already does well on virtual tasks, drawing on the language and reasoning it gained in general pretraining, and must now act in the physical world through a body. This section asks how the embodied agent acquires physical competence (training) and how it uses that competence during a task (deployment), with the environments of \Cref{sec:sim} supplying the demonstrations, rewards and rollouts used for training. \Cref{sec:learning-frameworks} reviews classic policy-learning ideas and methods: VLAs and WAMs as policy priors, data generation and distillation for imitation learning, model-assisted reinforcement learning, and automation of the training pipeline. These classic approaches are not yet methods for training self-improving embodied agents, but their techniques could potentially be used for that training. \Crefrange{sec:slowfast}{sec:llmpolicy} cover the LLM's execution of embodied tasks, from the most abstract model output to the most concrete: the LLM acts as a high-level planner that instructs low-level policies (\Cref{sec:slowfast}), writes code that generates trajectories (\Cref{sec:code}), or directly outputs the executed trajectories (\Cref{sec:llmpolicy}). \Cref{sec:others} covers multi-robot collaboration, and \Cref{sec:deployment-limits} discusses limitations and open questions.

\subsection{Policy Training}
\label{sec:learning-frameworks}

\begin{takeawaybox}
Classic policy-learning methods offer priors, demonstrations and curated data, exploration guidance, rewards and judging signals,
and automated training with acceptance checks. Transferring them to embodied systems built around a general-purpose model risks
carrying errors in data, guidance or judgement into learned behaviour. Better training scores and passing automated checks do not
guarantee the intended behaviour; evaluation must assess the resulting behaviour and account for the limits of the judge or success criterion.
\end{takeawaybox}

This subsection reviews classic policy-learning routes: VLA and WAM policy priors, data generation and distillation for imitation
learning, model-assisted reinforcement learning, and automation of the training pipeline. Directly training an embodied system
built around a general-purpose model remains at an early stage, so we examine techniques from these routes that could transfer to it.
When transferred, policy priors supply an initialisation (\Cref{sec:learning-priors}); data generation and distillation supply
demonstrations and curated data (\Cref{sec:learning-demos}); model-assisted reinforcement learning supplies exploration guidance,
rewards and judging signals (\Cref{sec:learning-rl}); and pipeline automation supplies automated training with acceptance checks
(\Cref{sec:learning-pipeline}). This subsection covers these techniques and their origins; a closed loop in which the system
organises the training of its own policy from its own feedback is reviewed in \Cref{sec:si-model}.

\subsubsection{Policy Priors}
\label{sec:learning-priors}

\begin{figure}[H]
    \centering
    \includegraphics[width=\linewidth]{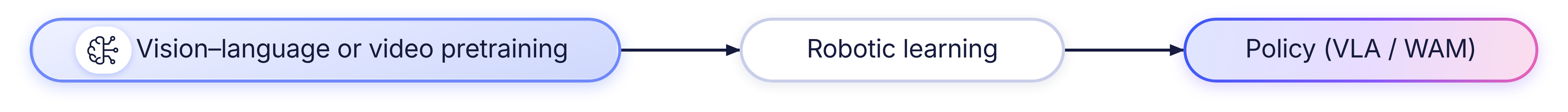}%
    \caption{\textbf{Policy priors.} Vision--language or video pretraining gives the policy its starting point, and robotic learning turns it into a vision--language--action model (VLA) or a world--action model (WAM).}
    \label{fig:learning-priors}
\end{figure}

Most work on policy priors starts the policy from the weights of a pretrained vision--language or video model. Vision--language--action models initialise the policy from a
web-pretrained VLM and train it on robot data, writing actions as text tokens, as RT-2 and OpenVLA do~\citep{brohan2023rt,kim2024openvla},
or through an attached action expert, as in $\pi_0$~\citep{black20240}; later systems co-train on data beyond robot actions so that
what the backbone knew is kept~\citep{intelligence20250,team2025geminia,team2026xiaomi}. Video generation models serve as plans from
which an inverse dynamics model extracts actions, as pretraining, or as representations that an action decoder
reads~\citep{du2023learning,wu2023unleashing,cheang2024gr,hu2024video,liao2025genie}, and world--action models train video and action
prediction together~\citep{li2025unified,song2026uniwam,nvidia2026cosmos}. The controlled comparisons of this line show which part of the
prior helps: with the action expert, robot data and training budget fixed, GroundingPI changes only the backbone and finds a grounding
model best in every out-of-distribution setting it tests~\citep{yu2026groundingpi}, and OpenWAM finds that world and action learning
help each other only with dedicated action capacity and an explicit flow of information from world to action~\citep{wang2026openwam}.

\subsubsection{Data Generation and Distillation}
\label{sec:learning-demos}

\begin{figure}[H]
    \centering
    \includegraphics[width=\linewidth]{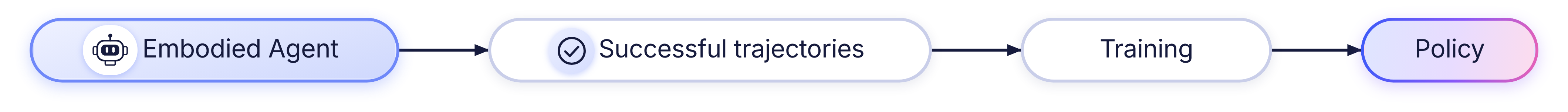}%
    \caption{\textbf{Data generation and distillation.} An embodied agent attempts the task, its successful trajectories are kept, and a policy is trained on them.}
    \label{fig:learning-demos}
\end{figure}

In data generation and distillation, the model supplies or shapes the data from which a student policy learns. Most often it is the
demonstrator. For a given task, it either writes a program that solves it, as in Code as
Policy (\Cref{sec:code}), often with access to the simulator state, or acts on the robot itself, as in LLM as Policy
(\Cref{sec:llmpolicy}), and a check keeps the runs that succeed. A smaller student policy, usually trained by behaviour cloning or
supervised fine-tuning, learns from these verified demonstrations and then runs without the model. The model can also judge data that
it did not produce, selecting or weighting demonstrations and rollouts by their estimated progress or success.

Each system is a pipeline in which a teacher produces runs, a verifier keeps those that succeed, an augmentation step multiplies them
and a student is trained on the result. SUDD established this pipeline: an LLM grounds the steps of a task in sampling-based
primitives that use privileged simulator state, writes its own success check and retries failures. A visuomotor student
distilled from the successful runs outperforms its own data-collection policy~\citep{ha2023scaling}. EmbodiedSWE~\citep{shen2026embodiedswe}, both a benchmark
and a data engine, has coding agents write and debug a solution program for each of 28 long-horizon tasks, graded where they cannot
write simulator state. Their edits of scenes and strategies give students a much higher held-out score, at a fixed data budget,
than variation by randomisation and noise alone. Further generators vary the teacher, the check
or what is distilled: SkillWeaver~\citep{zhu2026skillweaver} searches over reinforcement-learned skills, RE-0~\citep{zhang2026re} fine-tunes a
code-writing student on teacher patches kept only when a replay shows that they help, and MimicAgent~\citep{nayak2026mimicagent} writes
reference motions for a quadruped that reinforcement learning then tracks.

Filtering for success raises the question of whether the retained trajectories contain enough recovery behaviour for the student
to learn. The answer depends on what counts as a run. SUDD~\citep{ha2023scaling} keeps whole successful attempts,
including the failed tries that preceded success. Without retries, the distilled policy's average success falls from 67.0\% to
32.2\%, and it shows no recovery behaviour. EmbodiedSWE~\citep{shen2026embodiedswe} discards failed trajectories but writes recovery
branches into the teacher's strategies and injects noise to trigger them, without testing whether the student
recovers. Recova makes recovery a separate target: in a reconstructed digital twin, an agent diagnoses failures, tests
corrective programs and collects recovery rollouts for a dedicated recovery policy, which it invokes during deployment once it
detects a failure; on four real-robot workstations, these recovery skills raise mean success from 77.5\% after DAgger fine-tuning
to 87.5\%~\citep{liu2026recova}.

A student outperforming its teacher does not by itself establish the language model's contribution: the gain may come from the retries, the
student's training or the success filter. SUDD~\citep{ha2023scaling} reports no data-collection baseline without a language
model. RobotGPT~\citep{jin2023robotgpt} compares reinforcement learning from demonstrations produced by
self-corrected code with direct execution of the code. This comparison likewise shows what learning adds to the code, not what the
model adds over another demonstrator. The closest controls compare other sources of demonstrations. The VLM teacher of Manipulate-Anything~\citep{duan2024manipulate} acts from camera images without privileged state, and students trained on its demonstrations match those trained on hand-scripted
ones on simulated benchmark tasks. On a single real arm controlled by a frontier model~\citep{kim2026bridging}, a policy trained on
equal numbers of human and generated demonstrations succeeds as often as one trained on the same total of human
demonstrations. Robo-COP~\citep{zhang2026co}, compared with policy model improvement (\Cref{sec:si-model}), retrains its VLA on data collected during deployment; its comparison with a frozen
policy isolates the retraining but not the orchestrator.

Guava~\citep{liu2026guava} distils the model's decisions rather than its programs, fine-tuning a small student on validated decisions of a frontier
model inside a shared harness of semantic tools, where the student comes close to its teacher.
LocalNav~\citep{baumann2026localnav} applies the same idea to object-goal navigation. GLIDE~\citep{song2026learning} keeps people as the demonstrators: for
tasks that operators fail to teleoperate, a coding agent writes and revises guardrail code that filters their commands by task phase,
which, on three real tasks, turns mostly failed data collection into mostly successful collection.

Other systems curate data by judging it for supervised learning. GVL~\citep{ma2024vision} estimates progress to weight
demonstrations. ARS~\citep{hu2026ars} uses two prompted agents to propose and verify a timeline of task events, then estimates
per-frame progress to select and weight action chunks for behaviour cloning. SOAR~\citep{zhou2024autonomous} uses a frozen VLM to
propose feasible tasks from a fixed list and to judge the success of rollouts on five robots operating without resets; retraining on
the labelled successes improves the policy despite a success-detector precision of 0.63. Such curation helps when the model
distinguishes task-relevant states reliably, but suffers from inaccurate progress estimates, and what the judge mislabels the student
learns.

\subsubsection{Model-Assisted Reinforcement Learning}
\label{sec:learning-rl}
\label{sec:learning-reward}

\begin{figure}[H]
    \centering
    \includegraphics[width=\linewidth]{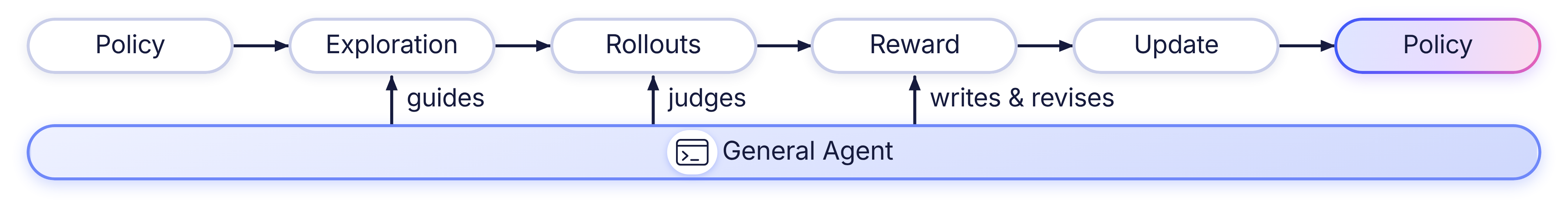}
    \caption{\textbf{Model-assisted reinforcement learning.} A reinforcement learning algorithm updates the policy, while a general agent guides exploration, judges rollouts, and writes and revises the reward.}
    \label{fig:learning-reward}
\end{figure}

In model-assisted reinforcement learning, a reinforcement learning algorithm trains the policy, while a general-purpose model guides
exploration and provides learning signals. Before or alongside learning, the model can write a control program that guides
exploration and behaviour. Between training runs, it can write and revise reward functions using training feedback. Inside the
training loop, it can act as a critic, evaluating observations or rollouts to provide rewards or preferences for learning.

\paragraph{Exploration and behaviour guidance.}
The model uses text descriptions of the task and the robot and object states to write an initial control program. This program
guides exploration towards states that random exploration would rarely reach, helping reinforcement learning when rewards are sparse.
Unlike in data generation (\Cref{sec:learning-demos}), the output is a prior policy rather than a set of verified demonstrations;
the learner keeps optimising a reward and can surpass the prior. The controller can serve in three distinct ways: to guide exploration, to constrain the policy through imitation, or as a base that the learner corrects.
RLingua~\citep{chen2024rlingua} uses a model-written controller both to guide exploration and as a target for imitation, so that the
policy learns tasks on which the same reinforcement learning algorithm alone fails. ExploRLLM~\citep{ma2024explorllm} uses
model-generated policy code to guide exploration, while a residual reinforcement learning policy learns to correct the base
controller. This combination improves sample efficiency in tabletop manipulation and supports zero-shot transfer to a real robot.
LAMARL~\citep{zhu2025lamarl} extends the approach to robot swarms, pulling the learned actions towards the prior's and adding a
generated reward, which improves sample efficiency and assembles shapes on real robots, although people still set the environment
parameters and correct generated functions that fail the model's review. The three uses differ in how far the prior's errors reach:
exploration guidance changes which states the learner visits, and residual correction learns what to add to the controller, whereas
where the learner is pulled towards the prior through imitation, the same guidance can carry the prior's errors into the policy and
hold back further improvement. None of these papers systematically tests robustness to a deliberately incorrect prior.

\paragraph{Reward design and revision.}
Here the task is given, and the model writes the reward function that reinforcement learning optimises. It reads the task description
and a description or the code of the environment, then revises its reward code using human comments, training
statistics~\citep{song2023self}, tests under varied physics or model judgements of rollout videos.
Text2Reward and Eureka established the training loop. Text2Reward~\citep{xie2023text2reward} describes the environment to the model as compact classes, writes
a dense reward zero- or few-shot and regenerates it from a person's comments on rollout videos, matching or beating expert-written
rewards on most of its simulated manipulation tasks. 
Eureka~\citep{ma2023eureka} removed the hand-made abstraction and the person: it gives the
model the environment's source code, trains a policy on each of several sampled rewards, keeps the one that scores best on the
benchmark's ground-truth success metric and rewrites it from a text summary of how each reward component evolved during training.
Eureka optimises a benchmark metric that may not fully reflect human intent, though human comments help stabilise gait; later systems widen the loop without changing this criterion. DrEureka~\citep{ma2024dreureka} carries it to the real robot,
adding tests under varied physics to choose domain randomisation, and teaches a real quadruped to walk on a yoga
ball. RF-Agent~\citep{gao2026rf} searches a tree over all past rewards instead of continuing from the best one, thus gaining in
optimisation rather than in fidelity to the instruction.

Behaviour-level critique targets a policy that optimises the reward but violates the instruction. RDA~\citep{lee2026rda} gives the clearest controlled
comparison of this line: against Eureka with the same model, instructions and compute, it has a VLM score each subtask of the
instruction in videos of the trained policies and rewrites the reward against the failures it names.  In GROVE~\citep{cui2025grove}, the CLIP-based pose score proposed as part of the reward also serves as the
judge, so the policy is trained on the signal that judges it. Its language model writes the reward and is queried again only when a fixed rule detects a sustained fall in that
score, without analysing the failures, so GROVE's contribution lies more in the combined reward than in the model's decisions. ROOT~\citep{sadikoglu2026root} separates the roles: a panel of VLMs that sees neither the method nor the reward code
scores the result. Yet the panel also rated one baseline favourably despite its low gait accuracy, so the authors read its scores
only alongside state-based metrics and human preferences on part of the tasks. In the systems reviewed, the
judge of behaviour is a VLM or a person, and human checks cover only part of the tasks.

\phantomsection
\label{sec:learning-feedback}
\paragraph{Evaluation and feedback.}
Inside the training loop, the model acts as a judge of a separate policy. It reads the task in language and the camera observations,
then returns the signal that the learner optimises: a score, a preference between two observations or a judgement of success, used
directly as a reward or to train a reward model. In the most recent systems, it also decides when an episode ends or what the policy
practises next, allowing the loop to run without a person labelling outcomes or resetting the environment. Unlike in reward design
above, the model does not write reward code between training runs but supplies the signal itself. Early work used pretrained models
in this way: VLM-RMs~\citep{rocamonde2023vision} uses CLIP similarity as a reward, while RL-VLM-F~\citep{wang2024rl} learns a reward
model from VLM preferences between observations. These methods improve policy learning when the model distinguishes task-relevant
states reliably, but struggle with visually similar states.

World-Env~\citep{xiao2025world} trains a VLA, from five demonstrations per task, inside a learned video world model
(\Cref{sec:sim-neural}), where a VLM-based reflector judges success and ends each imagined rollout; in its ablation, a pretrained VLM
prompted for success gives little gain and can impair learning on harder tasks, whereas a reward head trained on labelled expert and
policy rollouts makes the feedback useful. On real robots, FIND (\Cref{sec:sim-reset}) uses a VLM to identify which tasks of a fixed
library the scene allows and to judge each outcome as the sparse reward for residual RL on a frozen VLA, with its continued
improvement compared under policy model improvement (\Cref{sec:si-model}).

Reward models trained on robot data such as RoboReward's fine-tuned reward model~\citep{lee2026roboreward}, and progress
estimators that do not guide policy training such as ROVER~\citep{schroeder2025rover}, fall outside the scope of this survey.

Across the three groups, what the model supplies enters what the learner optimises, so its errors are optimised as well. Together,
these systems show that a model-written controller can help where the reward is sparse and that model judgements can serve as the
training signal when task progress or success is visible in camera images, although real-robot evidence remains limited to a few
tasks. How far a prior's errors survive in the policy, how reliably judges distinguish states near completion, how much task-specific
supervision they require, and whether policies exploit their errors remain open questions. ENPIRE takes model involvement further by
having agents run the whole training loop against a fixed verifier (\Cref{sec:learning-pipeline}).

\subsubsection{Training Pipeline}
\label{sec:learning-pipeline}

\begin{figure}[H]
    \centering
    \includegraphics[width=\linewidth]{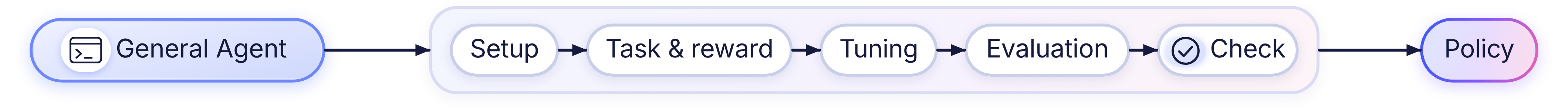}
    \caption{\textbf{Training pipeline.} A general agent runs the training pipeline, from setup, task and reward design and tuning to evaluation, with checks deciding what advances, and produces the policy. What performs the check differs across systems.}
    \label{fig:learning-pipeline}
\end{figure}

General agents can run the training infrastructure and the experiments on it themselves, for imitation or reinforcement learning,
rather than supplying one input to it. They start from an existing
codebase and a task in language, and their work stays checkable because each system tests a change before it advances, by different
means. The three systems that run the whole pipeline by editing its code are best defined by who proposes a change, who lets a stage advance and what
rejects a change or rolls it back. In HARBOR~\citep{li2026harbor}, sub-agents work through a sequence of executable gates, ranging from import checks to short training runs, that decide whether a stage may advance; a failed gate sends a summary back for repair, and once a fixed retry budget is used up, the user intervenes and the run resumes from the last state that passed its gates. The harness greatly raises stage pass rates in simulation, but its gates verify execution, not the final policy’s semantic correctness. RoboCoach~\citep{lei2026digital} separates proposal from decision: a coding agent revises the training code while a feedback
agent chooses from a tree of past experiments whether to extend, roll back or restart. With the simulator’s binary success fixed as an immutable metric, this loop clearly beats one-shot generation on simulated tasks and the platforms’ expert-written reference policies on real-robot tasks,
although some gains on simpler tasks may come
from ordinary code repair or hyperparameter search. ENPIRE~\citep{xiao2026enpire} shows that the agents can also build the
verification machinery on real bimanual stations: a coding agent first writes safety limits, reset code (\Cref{sec:sim-reset}) and a
reward built from a few minutes of demonstrations, refined under human assessment. Only for one task is the reward also checked
against a held-out labelled set that the agent can study but cannot train on or change. With these components held fixed, the agents edit the training code and select ideas by reported success, reaching near-perfect success on dexterous tasks.
An earlier and more constrained design, AURA~\citep{zhu2025aura}, generates a staged curriculum with its rewards, domain randomisation and training configuration as YAML specifications in a fixed schema for an existing simulation environment. Unlike the three systems above, it does not edit training code freely and only checks that a specification is well formed.

Each way of fixing the acceptance criterion has a limit: a frozen verifier (ENPIRE) passes its own errors on to every change it
judges, and a protected simulator metric (RoboCoach) says nothing about real-world behaviour beyond the transferred tasks. These systems support iterative improvement by retaining changes that pass their tests,
but whether they achieve recursive self-improvement (\Cref{sec:selfimprove}) depends on whether those changes also improve the
system's ability to make further improvements.

\begingroup\footnotesize\setlength{\tabcolsep}{4pt}\renewcommand{\arraystretch}{1.15}
\begin{xltabular}{\linewidth}{@{}>{\raggedright\arraybackslash}p{2.2cm} >{\raggedright\arraybackslash}X@{\hspace{2pt}}}
\caption{\textbf{Works on policy training} (\Cref{sec:learning-frameworks}), grouped by its four subsubsections and, within each, by what the general model supplies. Policy priors start the policy from a pretrained model's weights: vision--language--action models train a web-pretrained vision--language model on robot data, and video-based policies and world--action models build on video generation. Model-assisted reinforcement learning trains by reinforcement learning; the other groups use reinforcement or supervised learning. In data generation and distillation, reward design and revision and the training pipeline, a check, a metric, a vision--language model or a person decides what is kept. Robo-COP is compared in \Cref{tab:cmp-5-2}.}\label{tab:cmp-4-2}\\
\toprule\rowcolor{cmphead} Model supplies & Works \\ \midrule \endfirsthead
\toprule\rowcolor{cmphead} Model supplies & Works \\ \midrule \endhead
\bottomrule \endlastfoot
Policy priors & \begin{tabular}[t]{@{}>{\raggedright\arraybackslash}p{4.43cm}|>{\raggedright\arraybackslash}p{4.43cm} >{\raggedright\arraybackslash}p{4.43cm}@{}}\textit{Robot-trained VLM} \newline \mbox{RT-2}~\citeyearpar{brohan2023rt}, OpenVLA~\citeyearpar{kim2024openvla}, $\pi_0$~\citeyearpar{black20240}, Gemini Robotics 1.5~\citeyearpar{team2025geminia}, HybridVLA~\citeyearpar{liu2025hybridvla}, \mbox{PD-VLA}~\citeyearpar{song2025pd}, $\pi_{0.5}$~\citeyearpar{intelligence20250}, ReconVLA~\citeyearpar{song2025reconvla}, Spatial Forcing~\citeyearpar{li2025spatial}, \mbox{VLA-Adapter}~\citeyearpar{wang2025vla}, GroundingPI~\citeyearpar{yu2026groundingpi}, \mbox{Xiaomi-Robotics-1}~\citeyearpar{team2026xiaomi}, UniWAM~\citeyearpar{song2026uniwam} & \textit{Robot-trained video model} \newline \mbox{GR-1}~\citeyearpar{wu2023unleashing}, UniPi~\citeyearpar{du2023learning}, \mbox{GR-2}~\citeyearpar{cheang2024gr}, Video Prediction Policy~\citeyearpar{hu2024video}, Genie Envisioner~\citeyearpar{liao2025genie}, UVA~\citeyearpar{li2025unified}, \mbox{4D-WAM}~\citeyearpar{yang20264d}, \mbox{Cosmos 3}~\citeyearpar{nvidia2026cosmos}, OpenWAM~\citeyearpar{wang2026openwam} & \end{tabular} \\
\rowcolor{cmpstripe}Data generation and distillation & \begin{tabular}[t]{@{}>{\raggedright\arraybackslash}p{4.43cm}|>{\raggedright\arraybackslash}p{4.43cm}|>{\raggedright\arraybackslash}p{4.43cm}@{}}\textit{Program} \newline RobotGPT~\citeyearpar{jin2023robotgpt}, SUDD~\citeyearpar{ha2023scaling}, EmbodiedSWE~\citeyearpar{shen2026embodiedswe}, MimicAgent~\citeyearpar{nayak2026mimicagent}, \mbox{RE-0}~\citeyearpar{zhang2026re}, Recova~\citeyearpar{liu2026recova} & \textit{Actions} \newline \mbox{Manipulate-Anything}~\citeyearpar{duan2024manipulate}, Guava~\citeyearpar{liu2026guava}, \citet{kim2026bridging}, LocalNav~\citeyearpar{baumann2026localnav}, SkillWeaver~\citeyearpar{zhu2026skillweaver} & \textit{Filtered or weighted data} \newline GLIDE~\citeyearpar{song2026learning} (guardrail code), GVL~\citeyearpar{ma2024vision}, SOAR~\citeyearpar{zhou2024autonomous}, ARS~\citeyearpar{hu2026ars}\end{tabular} \\
Model-assisted reinforcement learning & \begin{tabular}[t]{@{}>{\raggedright\arraybackslash}p{4.43cm}|>{\raggedright\arraybackslash}p{4.43cm}|>{\raggedright\arraybackslash}p{4.43cm}@{}}\textit{Exploration and behaviour guidance} \newline RLingua~\citeyearpar{chen2024rlingua}, ExploRLLM~\citeyearpar{ma2024explorllm}, LAMARL~\citeyearpar{zhu2025lamarl} & \textit{Reward design and revision} \newline Kept by a ground-truth metric: Eureka~\citeyearpar{ma2023eureka}, DrEureka~\citeyearpar{ma2024dreureka}, \mbox{RF-Agent}~\citeyearpar{gao2026rf}; by a VLM: GROVE~\citeyearpar{cui2025grove}, RDA~\citeyearpar{lee2026rda}, ROOT~\citeyearpar{sadikoglu2026root}; by a check or a person: \mbox{Self-Refined LLM}~\citeyearpar{song2023self}, Text2Reward~\citeyearpar{xie2023text2reward} & \textit{Evaluation and feedback} \newline \mbox{VLM-RMs}~\citeyearpar{rocamonde2023vision}, \mbox{RL-VLM-F}~\citeyearpar{wang2024rl}, \mbox{World-Env}~\citeyearpar{xiao2025world}\end{tabular} \\
\rowcolor{cmpstripe}Training pipeline & \begin{tabular}[t]{@{}>{\raggedright\arraybackslash}p{4.43cm}|>{\raggedright\arraybackslash}p{4.43cm} >{\raggedright\arraybackslash}p{4.43cm}@{}}\textit{Training code} \newline ENPIRE~\citeyearpar{xiao2026enpire}, HARBOR~\citeyearpar{li2026harbor}, RoboCoach (Lei et al.)~\citeyearpar{lei2026digital} & \textit{YAML specification} \newline AURA~\citeyearpar{zhu2025aura} & \end{tabular} \\
\end{xltabular}
\endgroup

\needspace{16\baselineskip}
\subsection{Agent as System 2}
\label{sec:slowfast}

\begin{takeawaybox}
The model decides at a high level what a learned policy should do next, or judges a plan, its execution or its outcome, while the
policy produces the grounded motion. This division lets policies trained on short segments serve long-horizon tasks and recover from
failed steps. Its weakest point is judging whether a step has succeeded.
\end{takeawaybox}

\begin{figure}[H]
    \centering
    \includegraphics[width=\linewidth]{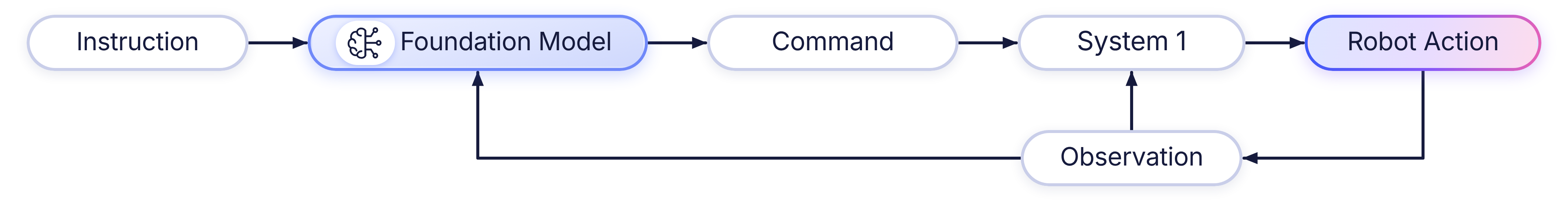}
    \caption{\textbf{Agent as System 2.} The foundation model, as the slow layer, reasons and judges, and passes a subgoal, skill choice or instruction to a learned policy (System 1) that produces actions in short segments. The policy acts on each observation, while the slow layer monitors execution and lets it continue, stop or pass to replanning.}
    \label{fig:subfig-slowfast}
\end{figure}

Policies trained with a model's help (Policy Training, \Cref{sec:learning-frameworks}) can serve as low-level controllers in larger systems. In this split, a learned low-level policy, such as a skill, VLA or navigation policy, forms the fast layer (System 1), producing grounded motion, while a large language model or vision--language model forms the slow layer (System 2), reasoning over environmental observations and robot state. On the one hand, the slow layer carries long, open-ended instructions, while the fast layer produces grounded motion on short task segments; the slow layer therefore \textbf{instructs} execution through a subgoal, skill choice or semantic instruction to the policy. On the other hand, because the slow layer answers more slowly than the robot moves and execution can fail silently, the fast layer keeps running while the slow layer \textbf{monitors} a proposed plan, ongoing execution or its outcome, allowing execution to continue, stop or pass to replanning and recovery. A model whose command a backend only converts and executes belongs instead to LLM as
Policy (\Cref{sec:llmpolicy}), and code that drives the robot to Code as Policy (\Cref{sec:code}); code or a formal specification
written only to check another decision maker belongs here. What the general model contributes has to be read against the trained
components it relies on. In Being-0, for instance, a lightweight vision--language connector trained on navigation data turns the
frontier model's plans into skill commands for a humanoid, and the same system without this connector completes far fewer
long-horizon tasks~\citep{yuan2025being}.

\subsubsection{Instructing Low-Level Policies}
\label{sec:slowfast-instruct}
Instruction draws on the complementarity of the two layers: the slow layer states what the low-level policy should achieve next and
leaves the motion that achieves it to the policy. In the first paradigm, mostly before 2026, the model chooses the next step, and
the systems form a sequence of mechanisms, each filling a gap that the previous one leaves. Language Models as Zero-Shot Planners,
Socratic Models and LM-Nav establish the division of labour, mapping a language model's steps to admissible actions, detected objects
or navigation landmarks, but without knowing whether a step can succeed~\citep{huang2022language,zeng2022socratic,shah2022lm}.
SayCan adds feasibility by weighting each skill with a value function that scores whether it can succeed from the current state;
removing the value functions clearly lowers planning success~\citep{ahn2022i}. Inner Monologue returns feedback after
every step, so that a skill that fails no longer goes unnoticed~\citep{huang2022inner}; ViLa reads the current image instead of
text reports~\citep{hu2023look}; and Text2Motion checks the whole skill sequence for feasibility before the robot
moves~\citep{lin2023text2motion}. Later systems extend this scheme rather than change it: SayPlan, SayNav, InstructNav, RoboMemory,
BUMBLE and MessyMem add scene graphs and memory~\citep{rana2023sayplan,rajvanshi2023saynav,long2024instructnav,lei2025robomemory,shah2024bumble,banwasi2026messymem},
with RoboMemory and MessyMem keeping memories across tasks, as in context improvement (\Cref{sec:si-icl}); CA-Nav calls the model
far less often~\citep{chen2024constraint}; and TidyBot, AquaChat, Manual2Skill and BiBo carry the division to new inputs and
bodies~\citep{wu2023tidybot,akram2025aquachat,tie2025manual2skill,jian2025endowing}. All of these systems treat a skill as finished
when it returns.

In 2026, frozen VLAs are called as tools in an agent's loop, and the model's work shifts from choosing the next step to managing a
policy's execution: deciding when to invoke, stop or switch the policy, verifying the outcome and recovering. A controlled study in
which only the stop decision differs shows why this is hard: letting the planner decide when a subgoal ends gives clearly
lower long-horizon task success than a separate success detector, although the detector sees privileged simulator
state~\citep{hu2026matters}. The systems of this paradigm divide these responsibilities in different ways. Pigey verifies each short subgoal
before replanning~\citep{galanti2026addressing}. Harness VLA calls the VLA with a chunk budget and a stop
predicate~\citep{zhang2026harness}, and NavMCP issues budgeted sub-goals to a frozen navigation foundation model and decides when
to stop~\citep{lei2026scaffolding}; NavGPT-3 runs reasoning, acting and monitoring as separate threads in an OS-like runtime that
decides which thread controls the robot, so that the navigation VLA can be interrupted and control switched when a sudden event
occurs~\citep{zhou2026navgpt}; Thea and SafeHarness add an independent evaluator and a check against routes too close to
obstacles~\citep{wang2026harness,xu2026coding}. DynaHarness moves stopping, switching and recovery out of the model into a
deterministic layer, and its ablations show that the low-level capabilities carry the gain: removing the analytic contact skills
sharply lowers success~\citep{deng2026dynaharness}. NovaPlan owes much of its long-horizon success to local recovery
from failed steps~\citep{fu2026novaplan}. Context that outlasts one task, such as BUMBLE's failures and Harness VLA's traces, leads to Self-Improvement (\Cref{sec:selfimprove}), where
AdaHVLA revises the orchestration code around a frozen VLA~\citep{shah2024bumble,zhang2026harness,tang2026adahvla}. Managing execution
thus helps when the failure is recognised, an intervention can be selected, and a capable low-level component carries it out.

\subsubsection{Runtime Monitoring}
\label{sec:slowfast-monitor}
Monitoring addresses the second concern: execution may fail while the slow layer is inactive. The slow layer judges another decision
maker instead of issuing the motion itself, and on its verdict execution continues or stops, or passes to
a replanner, a fallback or a human. \textbf{Supervision} judges task progress and deviation from the expected execution, whereas
\textbf{guardrails} judge whether an action or plan violates a safety constraint and hold a veto over the proposer. Supervision
differs first in when the model judges: KnowNo and UPS calibrate uncertainty before acting and ask for help only when the choice is
ambiguous~\citep{ren2023robots,yuan2026act}, while REFLECT explains a failure after the episode from multisensory
recordings~\citep{liu2023reflect}. During execution, supervision separates condition generation from frequent checking, so three
things can fail independently: the condition may miss the failure, a violation may be seen late, or no recovery may remain once it
is seen. In DoReMi the planner states with each skill what must remain true and a vision--language model checks it at short
intervals, succeeding far more often than Inner Monologue, which replans only when a skill ends~\citep{guo2023doremi}.
Code-as-Monitor compiles the condition into code that halts an unmodified policy without further model calls, but misses failures
without visible displacement~\citep{zhou2024code}. In AESOP an on-board scorer flags anomalies and the slow model picks a
fallback that model-predictive control keeps feasible during the wait, reaching the recovery region in every test
scenario, far more often than a planner that ignores the latency~\citep{sinha2024real}. Other supervisors differ in who judges: in
VoLo, COME-robot and FRAMES the model judges the running skill or step from images, whereas Cybo-Waiter checks predicates
geometrically and consults a VLM only when that evidence is uncertain~\citep{chen2026volo,zhi2024closed,periasami2026frames,ren2026cybo}.
The checks that must run at control frequency are therefore those that are cheap and fixed before execution; the slow model's judgements can be deferred where the fast layer absorbs
the delay.

Guardrails block an unsafe action even when it advances the task: a large model turns safety requirements into rules that a
separate component enforces. Safety Chip, RoboSafe and RoboGuard write these rules as temporal-logic specifications or executable
predicates~\citep{yang2023plug,wang2025robosafe,ravichandran2025safety}. On a Spot robot, with formulas checked by experts, Safety
Chip stays safe as the number of constraints grows, whereas the same planner given the constraints in its prompt does
not~\citep{yang2023plug}; RoboGuard grounds its rules with a trusted model that never sees user
prompts~\citep{ravichandran2025safety}. FORTRESS and CORE turn semantic hazards into regions of space, enforced by a reach-avoid
planner or a control-barrier-function filter~\citep{ganai2025real,ravichandran2026contextual}; CORE prevents most of the
violations towards which simulated controllers are steered, far more than generic rules written before
deployment~\citep{ravichandran2026contextual}. Guardrails are thus a form of monitoring in which the slow layer writes the rules and
a separate component enforces them, and safety holds only as far as the rules are right: formulas must be translated correctly, and
the specification must cover the hazard, for instance by recognising a hammer as a potential
weapon~\citep{yang2023plug,ravichandran2025safety}.

\begingroup\small\setlength{\tabcolsep}{5pt}\renewcommand{\arraystretch}{1.15}
\begin{xltabular}{\linewidth}{@{}>{\raggedright\arraybackslash}p{3.7cm} c >{\raggedright\arraybackslash}X >{\raggedright\arraybackslash}X >{\raggedright\arraybackslash}X >{\raggedright\arraybackslash}X >{\raggedright\arraybackslash}X@{\hspace{2pt}}}
\caption{\textbf{Comparison of works on agents as System 2} (\Cref{sec:slowfast}). The upper layer is the general model; the lower layer carries out what it passes down; Checked by: what judges the execution (--: no check). RoboMemory and MessyMem, which keep their memory across tasks, are compared in \Cref{tab:cmp-5-3}.}\label{tab:cmp-4-3}\\
\toprule\rowcolor{cmphead} Work & Link & Upper layer & Lower layer & Passed down & Update timing & Checked by \\ \midrule \endfirsthead
\toprule\rowcolor{cmphead} Work & Link & Upper layer & Lower layer & Passed down & Update timing & Checked by \\ \midrule \endhead
\bottomrule \endlastfoot
\citet{huang2022language} & \paperlink{https://arxiv.org/abs/2201.07207}\nocode & GPT-3 & Skills & Plan & Once & -- \\
\rowcolor{cmpstripe}Inner Monologue~\citeyearpar{huang2022inner} & \paperlink{https://arxiv.org/abs/2207.05608}\nocode & InstructGPT & Skills & Skill & Per step & Detector \\
LM-Nav~\citeyearpar{shah2022lm} & \paperlink{https://arxiv.org/abs/2207.04429}\nocode & GPT-3 & Nav policy & Subgoal & Once & -- \\
\rowcolor{cmpstripe}SayCan~\citeyearpar{ahn2022i} & \paperlink{https://arxiv.org/abs/2204.01691}\nocode & PaLM & Skills & Skill & Per step & Value fn \\
Socratic Models~\citeyearpar{zeng2022socratic} & \paperlink{https://arxiv.org/abs/2204.00598}\nocode & GPT-3 & Policy & Plan & Once & -- \\
\rowcolor{cmpstripe}Text2Motion~\citeyearpar{lin2023text2motion} & \paperlink{https://arxiv.org/abs/2303.12153}\nocode & Codex & Skills & Plan & Once & Value fn \\
ViLa~\citeyearpar{hu2023look} & \paperlink{https://arxiv.org/abs/2311.17842}\nocode & GPT-4V & Skills & Skill & Per step & Model \\
\rowcolor{cmpstripe}SayNav~\citeyearpar{rajvanshi2023saynav} & \paperlink{https://arxiv.org/abs/2309.04077}\nocode & GPT-3.5 & Nav policy & Plan & Periodic & -- \\
SayPlan~\citeyearpar{rana2023sayplan} & \paperlink{https://arxiv.org/abs/2307.06135}\nocode & GPT-4 & Skills & Plan & Once & Formal \\
\rowcolor{cmpstripe}TidyBot~\citeyearpar{wu2023tidybot} & \paperlink{https://arxiv.org/abs/2305.05658}\nocode & GPT-3.5 & Skills & Skill & Per step & -- \\
BUMBLE~\citeyearpar{shah2024bumble} & \paperlink{https://arxiv.org/abs/2410.06237}\nocode & GPT-4o & Skills & Skill & Per step & Code \\
\rowcolor{cmpstripe}CA-Nav~\citeyearpar{chen2024constraint} & \paperlink{https://arxiv.org/abs/2412.10137}\nocode & GPT-4 & Nav policy & Subgoal & Once & Detector \\
InstructNav~\citeyearpar{long2024instructnav} & \paperlink{https://arxiv.org/abs/2406.04882}\nocode & GPT-4 & Nav policy & Subgoal & Per step & Model \\
\rowcolor{cmpstripe}AquaChat~\citeyearpar{akram2025aquachat} & \paperlink{https://arxiv.org/abs/2507.16841}\nocode & GPT-4 & Controller & Plan & Once & Formal \\
Being-0~\citeyearpar{yuan2025being} & \paperlink{https://arxiv.org/abs/2503.12533}\nocode & GPT-4o & Skills & Skill & Per step & Model \\
\rowcolor{cmpstripe}BiBo~\citeyearpar{jian2025endowing} & \paperlink{https://arxiv.org/abs/2511.00041}\nocode & GPT-4o & Policy & Subgoal & Per step & Code \\
Manual2Skill~\citeyearpar{tie2025manual2skill} & \paperlink{https://arxiv.org/abs/2502.10090}\nocode & GPT-4o & Controller & Plan & Once & -- \\
\rowcolor{cmpstripe}DynaHarness~\citeyearpar{deng2026dynaharness} & \paperlink{https://arxiv.org/abs/2609.40306}\nocode & Qwen3-VL-4B & Skills & Skill & Per step & Code \\
Harness VLA~\citeyearpar{zhang2026harness} & \paperlink{https://arxiv.org/abs/2607.08448}\codelink{https://github.com/RLinf/RPent} & GPT-5.5 & VLA & Skill & Per step & Code \\
\rowcolor{cmpstripe}\citet{hu2026matters} & \paperlink{https://arxiv.org/abs/2606.10267}\nocode & Gemini 2.5 Flash & VLA & Subgoal & Per step & Model \\
NovaPlan~\citeyearpar{fu2026novaplan} & \paperlink{https://arxiv.org/abs/2602.20119}\nocode & GPT-5.2 & Video model & Subgoal & Per step & Model \\
\rowcolor{cmpstripe}Pigey~\citeyearpar{galanti2026addressing} & \paperlink{https://arxiv.org/abs/2607.21725}\codelink{https://github.com/lianegalanti/Pigey} & Claude Opus 4.7 & VLA & Skill & Per step & Model \\
SafeHarness~\citeyearpar{xu2026coding} & \paperlink{https://arxiv.org/abs/2609.20822}\nocode & GPT-5.5 & Controller & Plan & Per step & Geometry \\
\rowcolor{cmpstripe}Thea~\citeyearpar{wang2026harness} & \paperlink{https://arxiv.org/abs/2608.11246}\codelink{https://github.com/EIT-HAI/Thea} & GPT-5.5 & Skills & Skill & Per step & Model \\
NavMCP~\citeyearpar{lei2026scaffolding} & \paperlink{https://arxiv.org/abs/2608.30396}\nocode & Qwen3.6-Plus & Nav policy & Subgoal & On event & Model \\
\rowcolor{cmpstripe}NavGPT-3~\citeyearpar{zhou2026navgpt} & \paperlink{https://arxiv.org/abs/2610.10787}\nocode & GPT-6 Astra & VLA & Instruction & On event & Model \\
DoReMi~\citeyearpar{guo2023doremi} & \paperlink{https://arxiv.org/abs/2307.00329}\nocode & Vicuna-13B & Skills & Skill & Per step & Model \\
\rowcolor{cmpstripe}KnowNo~\citeyearpar{ren2023robots} & \paperlink{https://arxiv.org/abs/2307.01928}\nocode & PaLM-2L & Skills & Skill & Per step & Model \\
REFLECT~\citeyearpar{liu2023reflect} & \paperlink{https://arxiv.org/abs/2306.15724}\nocode & GPT-4 & Skills & Verdict & Once & Model \\
\rowcolor{cmpstripe}AESOP~\citeyearpar{sinha2024real} & \paperlink{https://arxiv.org/abs/2407.08735}\nocode & GPT-3.5 & Controller & Verdict & On event & Detector \\
Code-as-Monitor~\citeyearpar{zhou2024code} & \paperlink{https://arxiv.org/abs/2412.04455}\nocode & GPT-4o & Policy & Subgoal & Per step & Code \\
\rowcolor{cmpstripe}COME-robot~\citeyearpar{zhi2024closed} & \paperlink{https://arxiv.org/abs/2404.10220}\nocode & GPT-4V & Skills & Skill & Per step & Model \\
Cybo-Waiter~\citeyearpar{ren2026cybo} & \paperlink{https://arxiv.org/abs/2603.10675}\nocode & n.r. & Policy & Plan & Once & Geometry \\
\rowcolor{cmpstripe}FRAMES~\citeyearpar{periasami2026frames} & \paperlink{https://arxiv.org/abs/2609.22538}\nocode & GPT-5.4-mini & Controller & Skill & Per step & Model \\
UPS~\citeyearpar{yuan2026act} & \paperlink{https://arxiv.org/abs/2602.22474}\nocode & Gemini 2.0 Flash & Policy & Verdict & Per step & Model \\
\rowcolor{cmpstripe}VoLo~\citeyearpar{chen2026volo} & \paperlink{https://arxiv.org/abs/2606.07723}\nocode & Claude Opus 4.6 & VLA & Subgoal & Per step & Model \\
Safety Chip~\citeyearpar{yang2023plug} & \paperlink{https://arxiv.org/abs/2309.09919}\nocode & GPT-4 & Skills & Rule & Once & Formal \\
\rowcolor{cmpstripe}FORTRESS~\citeyearpar{ganai2025real} & \paperlink{https://arxiv.org/abs/2505.10547}\nocode & Molmo & Controller & Rule & Periodic & Detector \\
RoboGuard~\citeyearpar{ravichandran2025safety} & \paperlink{https://arxiv.org/abs/2503.07885}\nocode & GPT-4o & Skills & Rule & n.r. & Formal \\
\rowcolor{cmpstripe}RoboSafe~\citeyearpar{wang2025robosafe} & \paperlink{https://arxiv.org/abs/2512.21220}\nocode & Gemini 2.5 Flash & Skills & Rule & Per step & Code \\
CORE~\citeyearpar{ravichandran2026contextual} & \paperlink{https://arxiv.org/abs/2602.19983}\nocode & Gemma 3 27B & Controller & Rule & Periodic & Formal \\
\end{xltabular}
\endgroup

\needspace{16\baselineskip}
\subsection{Code as Policy}
\label{sec:code}

\begin{takeawaybox}
Code is a representation, not a guarantee: it carries out a wrong goal or assumption as faithfully as a right one. The reviewed evidence locates the remaining gap in the specification, the perception that grounds it, and the choice of which changes are handled locally and which return to the model.
\end{takeawaybox}

\begin{figure}[H]
    \centering
    \includegraphics[width=\linewidth]{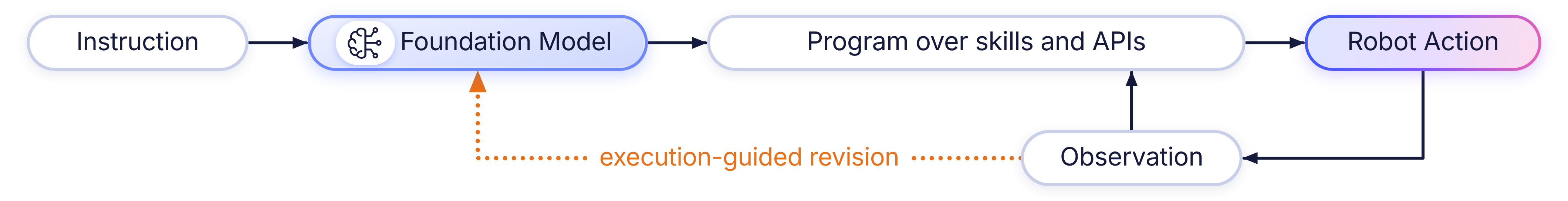}
    \caption{\textbf{Code as Policy.} The model writes a program over existing skills and APIs, and running it produces the action. Observations return to the program, and execution feedback can guide the model to revise it.}
    \label{fig:subfig-code}
\end{figure}

Code as Policy uses generated programs or computable specifications to connect a general model’s task reasoning with robot execution. Programs combine existing perception and motion functions through computation and control flow. Specifications express objectives or constraints that a planner or controller turns into motion. Execution feedback can also guide the model to revise either representation. We review these three complementary mechanisms, focusing on how they divide responsibility between the model and the robot’s execution stack. This division determines which changes the system can handle locally and which require further model reasoning.  

\subsubsection{Programs over Robot Interfaces}
Here, the model generates executable programs that compose existing perception and
motion interfaces. Computation and control flow express task-specific behaviour through reusable robot
capabilities~\citep{liang2022code,vemprala2023chatgpt,seker2026gta}. RoboTool~\citep{xu2023creative} divides program writing among an analyser, a planner, a calculator and a coder, all GPT-4, to solve creative tool-use tasks on an arm and a quadruped, and RoboScript~\citep{chen2024roboscript} runs the same generated script in simulation and on real arms through a unified ROS interface, with a benchmark in which GPT-4 clearly outperforms GPT-3.5 and Gemini. Their scope depends on the capabilities and abstraction of
those interfaces. Multimodal demonstrations can supply additional task information, including force-related control
parameters~\citep{wang2025chain}. Checks address reliability before and during execution: contract validation checks a generated
behaviour tree against a robot-side contract that defines the exposed skills, the allowed behaviour-tree operators and optional
composition templates~\citep{salfity2026contract}, while precondition checks guide runtime recovery in ProgPrompt's simulated
setting~\citep{singh2022progprompt}. Some functions or checks query the model at run
time~\citep{liang2022code,singh2022progprompt}. The returned value or recovery decision can change, while the surrounding program
and its control flow remain fixed. Recovery depends on the feedback available through the interfaces
and the responses encoded in the program.

\subsubsection{Constraints and Objectives for Solvers}
Here the model specifies constraints or objectives that the motion should
satisfy, and downstream planning, control or filtering computes or adjusts the motion. Because the specification is
kept apart from the motion, online perception can let the motion adapt to a changing scene while the specification stays the same.
VoxPoser and ReKep are representative implementations~\citep{huang2023voxposer,huang2024rekep}: VoxPoser's planner re-evaluates
3D affordance and avoidance maps on the latest observation, whereas ReKep re-solves keypoint-based costs from tracked keypoints and
backtracks to an earlier stage when a constraint is violated. What the robot achieves therefore depends jointly on the
specification, the perception that grounds it and the capability of the solver.

Three mechanisms recur. Geometric relations between object parts, or interaction points and axes in an object's frame, are solved
for poses~\citep{huang2024copa,pan2025omnimanip}. Rewards or objectives guide a controller or steer a frozen pretrained
policy~\citep{yu2023language,liu2026vls}. Formal or structured constraints go to existing planners or to safety
filters~\citep{chen2023autotamp,huo2026norm,darvish2024organa,wu2026embodiment}; in E$^2$-CARE the filter enforces predefined hard
safety constraints under its control assumptions, while violations of the model-generated soft constraints are penalised rather
than ruled out.

Systems can also combine mechanisms. ORGANA's model passes structured specifications to a planner and reads the results of
completed experiments to propose the next~\citep{darvish2024organa}. AgentChord compiles recovery branches for anticipated failures
into its task graph, realised by constraint-based optimisation and selected by compiled monitors without a new model call, and
finishes disturbed real-world tasks sooner than Code-as-Monitor~\citep{zhou2024code}, which calls the model again during
execution~\citep{xu2026reaction}. A failure that no branch covers may still require replanning, revision of the specification or
additional capabilities, not necessarily a rewrite by the model.

\subsubsection{Execution-Guided Revision}
In this branch, results of running the code on the robot or in simulation return to the model, which
revises the program or specification that follows. The reviewed systems draw on complementary sources of evidence for this
revision. Errors and fresh observations show whether an attempt produced its intended change, from syntax errors and faults in
flight to failed skills and camera or depth checks of the result~\citep{chen2023typefly,ouyang2024long,jia2026agent,chu2026openrua}.
ASPIRE makes this evidence finer-grained: for each perception, planning, grasping and control call it records the observations,
inputs, outputs and, where possible, visual evidence, so that its coding agent can localise the failing primitive, repair the
program and search for further candidate programs based on the remaining failure traces~\citep{lu2026aspire}. Validated repairs are kept in a skill library shared across tasks; this library is reviewed in \Cref{sec:selfimprove}.
Critics assess the plan or the executed behaviour~\citep{huo2026abot,ji2025genswarm}. Since an outcome shows that an attempt failed but not why, other systems actively probe, for instance to estimate mass or stiffness, or retain diagnosed failures and earlier episodes as experience~\citep{lin2026physcap,li2026aquacap,he2026robot}. In these systems, experience serves the current task: AquaCap's failure memory starts empty in each trial; He et al.'s experiments reuse an episode chosen by the experimenters; and ABot-Claw writes successful executions back into memory by design, without measuring their reuse. Accordingly, none is reviewed as self-improvement (\Cref{sec:selfimprove}). Refinement in simulation before deployment, as
in GenSwarm, is distinct from feedback obtained during physical execution, as in TypeFly, ABot-Claw and Agent as Policy, and the
systems also differ in what they observe and how they revise. Effective revision therefore depends both on the model's ability to
revise programs or specifications and on whether the returned evidence suffices to diagnose the discrepancy.

Revision by the model need not carry the fast physical response, because the two can run on different timescales. In KPI, a
vision-language agent writes stage-wise motion references and interaction contracts that state per direction whether to track,
comply or hold a force. Within a stage an adaptive impedance kernel handles contact, and between stages the agent revises the
remaining reference and contract from fresh observations. Its real-robot comparisons support this combined interface and control
design. The prompts are specific to each interface and the baselines also differ in their underlying control modules, so the
comparisons do not attribute the gain to either component alone~\citep{wang2026kpi}.

Beyond such local responses, a system must also decide when to call the model again. Frequent calls let it respond sooner to new
observations but cost more inference, and how often a call is needed depends on how much of the task each program runs before
control returns to the model. The reviewed systems place this return at different points: some let the model decide again between
tool calls, others revise the program at intervals~\citep{ge2026make,naouali2026vlcp}. How primitives are composed, whether program
state persists and how observations are delivered affect the same trade-off~\citep{si2026fewer}. What any schedule achieves also depends on the feedback
available, so results obtained in simulation with privileged object state, an explicit success signal and no preset cap on
attempts leave open how they carry over to physical execution~\citep{tsui2026demonstration}.

\begingroup\small\setlength{\tabcolsep}{5pt}\renewcommand{\arraystretch}{1.15}
\begin{xltabular}{\linewidth}{@{}>{\raggedright\arraybackslash}p{3.7cm} c >{\raggedright\arraybackslash}X >{\raggedright\arraybackslash}X >{\raggedright\arraybackslash}X >{\raggedright\arraybackslash}X@{\hspace{2pt}}}
\caption{\textbf{Comparison of works on code as policy} (\Cref{sec:code}). Update timing: when the model is called again.}\label{tab:cmp-4-4}\\
\toprule\rowcolor{cmphead} Work & Link & Upper layer & Writes & Run by & Update timing \\ \midrule \endfirsthead
\toprule\rowcolor{cmphead} Work & Link & Upper layer & Writes & Run by & Update timing \\ \midrule \endhead
\bottomrule \endlastfoot
Code as Policies~\citeyearpar{liang2022code} & \paperlink{https://arxiv.org/abs/2209.07753}\nocode & Codex & Program & APIs & In program \\
\rowcolor{cmpstripe}ProgPrompt~\citeyearpar{singh2022progprompt} & \paperlink{https://arxiv.org/abs/2209.11302}\nocode & GPT-3 & Program & Skills & In program \\
ChatGPT for Robotics~\citeyearpar{vemprala2023chatgpt} & \paperlink{https://arxiv.org/abs/2306.17582}\codelink{https://github.com/microsoft/PromptCraft-Robotics} & ChatGPT & Program & APIs & User turn \\
\rowcolor{cmpstripe}Chain-of-Modality~\citeyearpar{wang2025chain} & \paperlink{https://arxiv.org/abs/2504.13351}\nocode & Gemini 1.5 Pro & Program & APIs & Once \\
GTA-2~\citeyearpar{seker2026gta} & \paperlink{https://arxiv.org/abs/2609.09808}\nocode & Gemini 3.1 Pro & Program & Controller & User turn \\
\rowcolor{cmpstripe}\citet{salfity2026contract} & \paperlink{https://arxiv.org/abs/2607.12220}\nocode & Claude Sonnet 4.6 & Behaviour tree & Skills & On failure \\
RoboTool~\citeyearpar{xu2023creative} & \paperlink{https://arxiv.org/abs/2310.13065}\nocode & GPT-4 & Program & Skills & Once \\
\rowcolor{cmpstripe}RoboScript~\citeyearpar{chen2024roboscript} & \paperlink{https://arxiv.org/abs/2402.14623}\nocode & GPT-4 & Program & APIs & Once \\
AutoTAMP~\citeyearpar{chen2023autotamp} & \paperlink{https://arxiv.org/abs/2306.06531}\nocode & GPT-4 & Formal spec & Planner & On failure \\
\rowcolor{cmpstripe}Language to Rewards~\citeyearpar{yu2023language} & \paperlink{https://arxiv.org/abs/2306.08647}\nocode & GPT-4 & Objective & Controller & User turn \\
VoxPoser~\citeyearpar{huang2023voxposer} & \paperlink{https://arxiv.org/abs/2307.05973}\codelink{https://github.com/huangwl18/VoxPoser} & GPT-4 & Objective & Planner & Once \\
\rowcolor{cmpstripe}CoPa~\citeyearpar{huang2024copa} & \paperlink{https://arxiv.org/abs/2403.08248}\nocode & GPT-4V & Constraints & Planner & Once \\
ORGANA~\citeyearpar{darvish2024organa} & \paperlink{https://arxiv.org/abs/2401.06949}\codelink{https://github.com/ac-rad/organa} & GPT-4 & Formal spec & Planner & Per stage \\
\rowcolor{cmpstripe}ReKep~\citeyearpar{huang2024rekep} & \paperlink{https://arxiv.org/abs/2409.01652}\nocode & GPT-4o & Constraints & Planner & Once \\
OmniManip~\citeyearpar{pan2025omnimanip} & \paperlink{https://arxiv.org/abs/2501.03841}\nocode & GPT-4o & Constraints & Planner & On failure \\
\rowcolor{cmpstripe}AgentChord~\citeyearpar{xu2026reaction} & \paperlink{https://arxiv.org/abs/2605.11951}\codelink{https://github.com/Jasonxu1225/AgentChord} & GPT-5 & Constraints & Planner & Once \\
E$^2$-CARE~\citeyearpar{wu2026embodiment} & \paperlink{https://arxiv.org/abs/2606.28592}\nocode & n.r. & Constraints & Skills & Per stage \\
\rowcolor{cmpstripe}NORM-Nav~\citeyearpar{huo2026norm} & \paperlink{https://arxiv.org/abs/2605.16979}\nocode & Qwen-VL-Max & Constraints & Planner & Once \\
VLS~\citeyearpar{liu2026vls} & \paperlink{https://arxiv.org/abs/2602.03973}\nocode & n.r. & Objective & Policy & Per stage \\
\rowcolor{cmpstripe}TypeFly~\citeyearpar{chen2023typefly} & \paperlink{https://arxiv.org/abs/2312.14950}\nocode & GPT-4 & Program & Skills & In program \\
\citet{ouyang2024long} & \paperlink{https://arxiv.org/abs/2404.05291}\nocode & GPT-4 Turbo & Program & Skills & On failure \\
\rowcolor{cmpstripe}GenSwarm~\citeyearpar{ji2025genswarm} & \paperlink{https://arxiv.org/abs/2503.23875}\codelink{https://github.com/WindyLab/GenSwarm} & o1-mini & Program & APIs & On failure \\
ABot-Claw~\citeyearpar{huo2026abot} & \paperlink{https://arxiv.org/abs/2604.10096}\codelink{https://github.com/amap-cvlab/ABot-Claw} & n.r. & Program & Skills & On failure \\
\rowcolor{cmpstripe}Agent as Policy~\citeyearpar{jia2026agent} & \paperlink{https://arxiv.org/abs/2609.12541}\nocode & GPT-6 Astra & Program & APIs & Per decision \\
AquaCap~\citeyearpar{li2026aquacap} & \paperlink{https://arxiv.org/abs/2609.23133}\nocode & DeepSeek-V4-Pro & Program & APIs & On failure \\
\rowcolor{cmpstripe}FAEA~\citeyearpar{tsui2026demonstration} & \paperlink{https://arxiv.org/abs/2601.20334}\nocode & Claude Opus 4.5 & Program & APIs & On failure \\
\citet{he2026robot} & \paperlink{https://arxiv.org/abs/2609.31770}\codelink{https://github.com/hesd10/astra-robot-sim2real} & GPT-6 Astra & Program & APIs & Per decision \\
\rowcolor{cmpstripe}KPI~\citeyearpar{wang2026kpi} & \paperlink{https://arxiv.org/abs/2609.36151}\nocode & GPT-6 Astra & Motion reference & Controller & Per stage \\
PhysCaP~\citeyearpar{lin2026physcap} & \paperlink{https://arxiv.org/abs/2608.21031}\nocode & Gemini 3.1 Pro & Program & APIs & Per decision \\
\rowcolor{cmpstripe}PyRUA-Lean~\citeyearpar{si2026fewer} & \paperlink{https://arxiv.org/abs/2610.01939}\codelink{https://github.com/DAGroup-PKU/PyRUA-Lean} & GPT-6 Astra & Program & APIs & Per decision \\
URAI~\citeyearpar{ge2026make} & \paperlink{https://arxiv.org/abs/2609.39018}\nocode & Claude Fable 5.1 & Program & APIs & Per decision \\
\rowcolor{cmpstripe}VLCP~\citeyearpar{naouali2026vlcp} & \paperlink{https://arxiv.org/abs/2608.16978}\nocode & GPT-5.5 & Program & Controller & Interval \\
\end{xltabular}
\endgroup

\needspace{16\baselineskip}
\subsection{LLM as Policy}
\label{sec:llmpolicy}
\label{sec:llm_as_policy}

\begin{takeawaybox}
General-purpose models can now choose concrete manipulation targets, from end-effector poses to step-wise motion commands. A stronger model raises success under a matched interface, but the leader varies by benchmark and early results need replication. With the model fixed, spatial evidence and command design still change outcomes, and precision near contact remains weak.
\end{takeawaybox}

\begin{figure}[H]
    \centering
    \includegraphics[width=\linewidth]{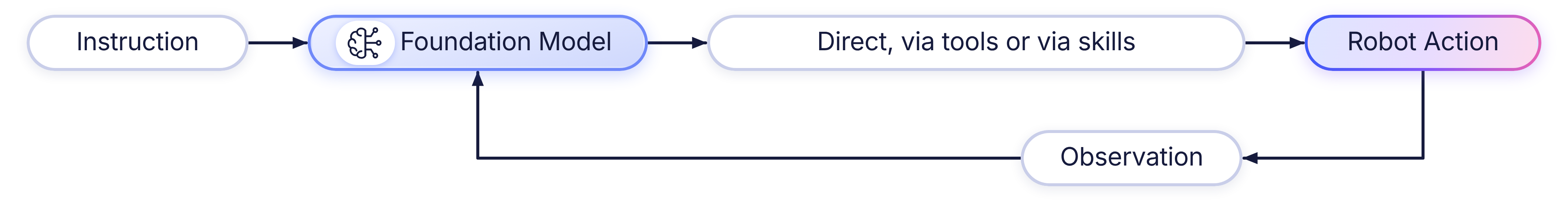}
    \caption{\textbf{LLM as Policy.} Given the instruction and each observation, the executed action is settled in one of three ways: the model outputs it directly, chooses it from what visual and geometric tools return, or reuses and evolves stored skills; the model then decides again from the next observation.}
    \label{fig:subfig-llmpolicy}
\end{figure}

LLM as Policy assigns concrete manipulation decisions to a general-purpose language or multimodal model, which translates spatial understanding into executable motion targets and revises them through feedback. Evaluations of GPT-6 Astra show that such a model can now manipulate well zero-shot: on all 42 RoboDojo tasks it ranks above every one of 40 public policies~\citep{zhang2026unexpected}, and systems built around it succeed on real and simulated tasks~\citep{cheng2026context,hu2026visual,wang2026spatialharness}, although precision and dynamic control remain weak.

For manipulation, the reviewed work falls into the three groups of \Cref{tab:cmp-4-5}, told apart by what settles the executed action. In \textbf{direct action output}, the model outputs poses, displacements, joint or gripper commands or action chunks that a backend only converts and executes. In \textbf{calling visual and geometric tools}, it chooses the action from what depth, segmentation, grasp, measurement or virtual-view tools return. In \textbf{skill reuse and evolution}, actions, programs or failure lessons are stored, verified and later reused, repaired or evolved, as code or text. Each work appears once, by its main contribution: code with a skill store goes to the third group, in-context demonstrations alone to the first, and mixed papers are described by the configuration in which the model generates actions; single-task programs belong to code as policy (\Cref{sec:code}), and works treated there or in \Cref{sec:selfimprove} are seen only from the angle of action generation. Navigation, aerial and legged systems are reviewed at the end.

\subsubsection{Direct Action Output}
Early systems predicted trajectories in context from keypoint tokens (KAT~\citep{palo2024keypoint}) or textual keyframes (RoboPrompt~\citep{yin2024context}), revised actions from motion-planner failures (LLM3~\citep{wang2024llm3}) or replanned after each end-effector waypoint (MALMM~\citep{singh2024malmm}). In An Unexpected Robot Policy, Astra commanding end-effector pose and gripper substantially outperforms GPT-5.5 with action space, observations and context held constant~\citep{zhang2026unexpected}, while Show-Harness finds that, with the model fixed, the motion scale of its commands still affects fine manipulation~\citep{chen2026show}. Without task-specific training, MotorMind reaches 66.7\% on the base LIBERO-PRO suites and 53.8\% under perturbations, against at most 13.3\% and 19.2\% for earlier zero-shot methods, and 83.3\% with a GPT-6 Sol backbone~\citep{li2026motormind}; RoboDawn gives demonstrations and execution one command vocabulary around a common gripper interaction point~\citep{guo2026transferring}. Robot-use agents go further and hand the robot to a general coding agent: in OpenRUA an off-the-shelf coding agent operates the robot zero-shot through terminal access to ROS~2~\citep{chu2026openrua}.

What the model holds in its context also matters. GPT-Policy conditions Astra's per-step pose or gripper requests on demonstrations in context~\citep{cheng2026context}, and RoboICL, which keeps the demonstrations apart from a bounded memory of the interaction, scores 50.64 over 30 RoboDojo tasks against 33.68 for the strongest baseline~\citep{liu2026roboicl}. On the 16 memory-dependent RoboMME tasks, Astra keeping the whole episode in context succeeds in 92.1\% of test episodes, above human players with an oracle skill interface (90.5\%) and a hierarchical system with a trained policy (79.1\%)~\citep{jedlicki2026frontier}. Allowed up to ten times the step limit that RoboDojo prescribes, Astra reaches 81.9\% on RoboDojo, the longer budget alone roughly tripling its success because it detects failed actions and tries again~\citep{sensenova2026has}; given tactile images, it exceeds the strongest specialist baselines on nineteen simulated contact-rich tasks by 22.9 and 13.1 percentage points~\citep{yang2026gpt}; and \citet{zhou2026large} find the widest gap between semantic reasoning and physical execution in consequence-aware planning, precision and coordinated motion.

Systematic evaluations locate where direct output works. On ten RoboDojo tasks Astra's own commands succeed in 13 of 50 instances against 24 for a hybrid in which it corrects targets for $\pi_{0.5}$, whereas on RoboLab's single-arm tasks they succeed in 49 of 50 trials against 18 for $\pi_{0.5}$ alone~\citep{team2026systematically}; EBench likewise finds direct waypoints competitive with post-trained policies and leading on short-horizon tasks but trailing on long-horizon, tabletop and high-precision contact-rich ones~\citep{yao2026profiling}. LIBERO-Agent reports performance falling sharply on hard short- and long-horizon tasks~\citep{diao2026libero}, RobotWorld agents that lose task-relevant object states or take unfinished tasks for complete~\citep{yang2026robotworld}, and RoboQuest agents that stop exploring too early, the best, Astra, succeeding in 23.2\% of episodes~\citep{renhang2026roboquest}. Across model families, four frontier models on RoboColiseum score between 85.89 and 99.35 on instruction following but only between 38.79 and 46.70 on manipulation, with different models leading different dimensions~\citep{huang2026frontiera}; six models on LIBERO-10 reach at most 44\%~\citep{espinola2026reasoning}; and nine models on physical arms are not yet precise or efficient enough within a practical action budget, although measured performance doubled about every 53.9 days~\citep{leet2026bitter}. The model is thus a main contributor alongside the action representation, but these early evaluations, reported within weeks of the models' release and with a leader that changes with the benchmark, need replication across model families and over time.

\subsubsection{Calling Visual and Geometric Tools}
PIVOT and MOKA had the model choose among candidate actions or keypoints marked on the image~\citep{nasiriany2024pivot,liu2024moka}. VIA lets the model adjust and inspect a virtual target gripper before committing it to a controller, raising GPT-5.6 Sol's success on six simulated tasks from 25.0\% with direct poses to 73.3\%~\citep{hu2026visual}. Robo-Harness K1 exposes calibrated depth, measurements, visual anchors and grasp proposals as callable tools, and its comparison shows that the ranking of models depends on the setting: with images only, Gemini 3.7 Flash reaches 77.8\% against 61.1\% for Astra~\citep{li2026robo}. SpatialHarness gives a frozen Astra virtual views and object poses from a calibrated simulated scene synchronised with real execution, raising real plug-insertion success from 26.7\% to 66.7\%~\citep{wang2026spatialharness}. Other systems make the tools the whole workspace: WAA has the model preview and revise editable action proposals in contact views, reaching 75.6\% on LIBERO-PRO with Gemini 3.7 Flash, above end-to-end VLAs, code-as-policy agents and a visual harness with the same backbone~\citep{zhang2026world}; RobotUse leaves geometry, motion planning and control to the backend and reaches 45\% on RoboLab, 6.7 percentage points above CaP-X~\citep{lee2026robotuse}; and the RoboHarness of \citet{qu2026roboharness}, distinct from the system of that name in \Cref{sec:selfimprove}, adds a visual-geometric control panel and outscores the VLA models it compares on the first ten BEHAVIOR Challenge 2025 tasks, with a prompt written by a person for each task. With one shared harness of motion, perception and skill tools and no privileged inputs, uni-robot-agent lets the open-weight Qwen3.8-Flash-Next reach a RoboDojo score of 31.39, above the 28.97 published for GPT-6 Astra with RoboProbe, although Astra itself scores 51.90 in the same harness~\citep{yuan2026democratizing}. In all of these the tools measure and the model decides, and with the same model better access to geometry changes the outcome, as VIA and SpatialHarness show.

\subsubsection{Skill Reuse and Evolution}
In this group what the model learns from acting is kept for its later actions. Teach and Grow turns a few demonstrations into verified Skill Blocks kept in a persistent library with an experience memory, reaching 99.9\% mean success on four LIBERO suites, and its controlled studies show taught blocks persisting and improving related tasks under the same model weights~\citep{nie2026teach}. RACaP evolves policy interfaces, harness and memory before deployment so that at run time the model only calls functions, solving 46.0\% of LIBERO-Long tasks against at most 4.0\% for code-as-policy baselines~\citep{li2026racap}. Skill2Real admits skill updates learned in simulation only when validation rollouts support them; its frozen LIBERO-90 skills raise Astra's success on unseen LIBERO-PRO Long tasks from 2.0\% to 56.3\%, and real UR5e completion rises from 27.50\% to 78.75\%~\citep{he2026skill2real}. ReTrace keeps validated corrections as a text skill, taking Astra on a real Franka from 0\% to 80\% on plug insertion~\citep{yu2026retrace}. Skill-RSI has a strong model write and repair a skill in simulation until it passes an exam on layouts drawn after freezing; the frozen skills kept 148 of 150 successes on new layouts with no model in the loop, although three of seven RoboTwin~2.0 tasks did not compile in 18 campaigns~\citep{bai2026compile}. SimEX~\citep{hu2026simex}, Scale and Selection~\citep{wei2026scale}, RPG~\citep{wang2026reconstruct}, PhysEvo~\citep{tian2026physevo}, COAP~\citep{hu2026embodied}, EmbodiedEvo~\citep{li2026embodiedevo}, RAX~\citep{huang2026scaling}, HexaShell~\citep{gao2026hexashell} and ASPIRE~\citep{lu2026aspire}, reviewed in full in \Cref{sec:selfimprove}, likewise evolve control tools, skills or harnesses from execution and keep them for later tasks, as does URAI~\citep{ge2026make} in \Cref{sec:code}.

Benchmarks have begun to ask when retained skills help. EmbodiedSkillBench finds that on 100 perturbed LIBERO-PRO tasks executing only skills lifts GPT-5.6 Sol from 11\% to 63\% but lowers Astra from 92\% under native control to 61\%, while calling skills selectively recovers 85\%~\citep{han2026embodied}; SEAR finds that on easier tasks re-rolling short attempts does as well as learning, and that control code and shared routines transfer while task-specific solutions do not~\citep{wang2026sear}; and PhysRSI-eval finds that removing the interaction history lowers late-round success~\citep{li2026physical}. The strongest evidence for reuse therefore comes from frozen skills tested on layouts or tasks not used to form them, as in Skill2Real, Skill-RSI and COAP; without a comparison with repeated attempts or native control, a gain may reflect more tries rather than retained experience.

\subsubsection{Beyond Manipulation}
The same pattern, a general model choosing the next concrete motion target step by step, also appears beyond manipulation. The reviewed work covers three further kinds of task: navigation of ground robots, from moving between the viewpoints of a graph to moving in continuous space; flight of aerial robots; and locomotion of legged and humanoid robots. The action the model chooses ranges from the next viewpoint or a discrete move, through an image-space waypoint, a goal handed to a navigation stack or a parameterised skill call, down to target joint positions. Whatever its form, the model issues one such action, reads the observation that follows and decides again, while planners, controllers or learned motion generators below it turn the action into motion. Earlier systems established these forms: in vision-and-language navigation, LM-Nav~\citep{shah2022lm} still used the language model only to extract landmarks that CLIP grounding and graph search turned into waypoints (\Cref{sec:slowfast}), whereas NavGPT~\citep{zhou2023navgpt}, LangNav~\citep{pan2023langnav} and DiscussNav~\citep{long2023discuss} had it choose the next viewpoint from text describing the views, and ReMEmbR~\citep{anwar2024remembr} retrieved a goal from a long-horizon memory for a navigation stack. Beyond ground navigation, See, Point, Fly~\citep{hu2025see} had a frozen vision-language model mark a waypoint in each camera frame of a drone, VLM-PC~\citep{chen2024commonsense} queried GPT-4o again after each skill a real quadruped executed, and at the lowest level Prompt a Robot to Walk~\citep{wang2023prompt} made GPT-4 itself the feedback controller of a simulated quadruped, outputting target joint positions for a PD controller at each control step.

In continuous environments (VLN-CE), the model chooses each move through tools. In Embodied Agents Take Control~\citep{zhou2026embodied}, off-the-shelf coding-agent harnesses steer every step zero-shot through an observation tool and a step tool; NavGPT-3~\citep{zhou2026navgpt} lets a planning model either move the robot through local motion tools or delegate the instruction to a trained navigation policy and monitor its progress, under a hierarchical runtime that decides which thread controls the motion; and AgentVLN~\citep{xin2026agentvln} fine-tunes a compact vision-language model to call navigation skills with target pixels, selecting path points projected into the image that a planning skill turns into 3D waypoints. Because LangNav and AgentVLN fine-tune the deciding model, it is a policy in our sense (\Cref{sec:def-agent}), and they are cited here as contrasts rather than reviewed. Spatial-Semantic MCP Navigation~\citep{lee2026spatial} retrieves a goal through map tools and sends it to a navigation stack, while ROSClaw~\citep{cardenas2026rosclaw} drives wheeled, quadruped and humanoid ROS~2 robots turn by turn, with one validated tool call per turn, such as a velocity command or a navigation goal.

In aerial navigation the model usually points rather than steers. OnFly~\citep{zheng2026onfly} runs the decision on board: each image-space goal point passes a verifier and a local planner before the quadrotor flies it, while a slower monitoring call on the same model signals whether to continue, stop or treat the target as lost. DroneCATS~\citep{park2026evaluating} lets the model choose at every step whether to fly towards an image point at a stated depth, rotate, think or declare arrival, and AerialClaw~\citep{li2026aerialclaw} has it emit one validated skill call per iteration in simulation.

For legged and humanoid robots the model chooses among locomotion skills: HumanCLAW~\citep{siyao2026humanclaw} has a vision-language model choose one parameterised whole-body skill for each 0.5\,s, which a verifier checks and a learned motion generator turns into motion of a simulated humanoid.

These approaches expose complementary aspects of robot control to general-purpose models: generating actions directly, grounding them with visual and geometric tools, and keeping what works as skills for later actions. The first two recur in navigation, flight and locomotion, where the action becomes a viewpoint, a waypoint in the image, a skill call or a set of joint positions, and the systems reviewed beyond manipulation all fall into these two groups. Their combination makes model capability, perceptual support, interface design and retained experience jointly relevant to policy execution.

\begingroup\small\setlength{\tabcolsep}{5pt}\renewcommand{\arraystretch}{1.15}
\begin{xltabular}{\linewidth}{@{}>{\raggedright\arraybackslash}p{3.7cm} c >{\raggedright\arraybackslash}X >{\raggedright\arraybackslash}X >{\raggedright\arraybackslash}X >{\raggedright\arraybackslash}X >{\raggedright\arraybackslash}X@{\hspace{2pt}}}
\caption{\textbf{Comparison of works on LLMs as policy} (\Cref{sec:llmpolicy}). Branch: direct action output (Direct: the model outputs the action and a backend only converts and executes it), calling visual and geometric tools (Tools: the model chooses the action from what depth, segmentation, grasp, measurement, virtual-view or visual-control tools return), or skill reuse and evolution (Skills: actions, programs or lessons are stored, verified and then reused, repaired or evolved); each work is listed once, under its main contribution, and systems beyond manipulation are placed by the same test. $^{*}$: reviewed in full in \Cref{sec:code} or \Cref{sec:selfimprove} and described in \Cref{sec:llmpolicy} only from the angle of action generation. Four to Nine models: the paper compares that many models and names no default.}\label{tab:cmp-4-5}\\
\toprule\rowcolor{cmphead} Work & Link & Model & Domain & Sees & Outputs & Branch \\ \midrule \endfirsthead
\toprule\rowcolor{cmphead} Work & Link & Model & Domain & Sees & Outputs & Branch \\ \midrule \endhead
\bottomrule \endlastfoot
Prompt a Robot to Walk~\citeyearpar{wang2023prompt} & \paperlink{https://arxiv.org/abs/2309.09969}\nocode & GPT-4 & Legged & Text & Joint target & Direct \\
\rowcolor{cmpstripe}NavGPT~\citeyearpar{zhou2023navgpt} & \paperlink{https://arxiv.org/abs/2305.16986}\codelink{https://github.com/GengzeZhou/NavGPT} & GPT-4 & Navigation & Text & Waypoint & Direct \\
KAT~\citeyearpar{palo2024keypoint} & \paperlink{https://arxiv.org/abs/2403.19578}\nocode & GPT-4 Turbo & Manipulation & Text & Trajectory & Direct \\
\rowcolor{cmpstripe}RoboPrompt~\citeyearpar{yin2024context} & \paperlink{https://arxiv.org/abs/2410.12782}\nocode & GPT-4 Turbo & Manipulation & Text & Trajectory & Direct \\
LLM3~\citeyearpar{wang2024llm3} & \paperlink{https://arxiv.org/abs/2403.11552}\codelink{https://github.com/AssassinWS/LLM-TAMP} & GPT-4 Turbo & Manipulation & Text & Pose & Direct \\
\rowcolor{cmpstripe}MALMM~\citeyearpar{singh2024malmm} & \paperlink{https://arxiv.org/abs/2411.17636}\nocode & GPT-4 Turbo & Manipulation & Text & Pose & Direct \\
VLM-PC~\citeyearpar{chen2024commonsense} & \paperlink{https://arxiv.org/abs/2407.02666}\nocode & GPT-4o & Legged & Image & Skill call & Direct \\
\rowcolor{cmpstripe}See, Point, Fly~\citeyearpar{hu2025see} & \paperlink{https://arxiv.org/abs/2509.22653}\nocode & Gemini 2.0 Flash & Aerial & Image & Waypoint & Direct \\
AerialClaw~\citeyearpar{li2026aerialclaw} & \paperlink{https://arxiv.org/abs/2606.12142}\codelink{https://github.com/XDEI-Group/AerialClaw} & n.r. & Aerial & Text & Skill call & Direct \\
\rowcolor{cmpstripe}Astra on RoboDojo~\citeyearpar{zhang2026unexpected} & \paperlink{https://arxiv.org/abs/2609.24170}\codelink{https://github.com/RoboProbe/RoboProbe} & GPT-6 Astra & Manipulation & Image & Pose & Direct \\
\citet{leet2026bitter} & \paperlink{https://arxiv.org/abs/2610.25287}\nocode & Nine models & Manipulation & Image & Pose & Direct \\
\rowcolor{cmpstripe}DroneCATS~\citeyearpar{park2026evaluating} & \paperlink{https://arxiv.org/abs/2609.01404}\codelink{https://github.com/naver-ai/DroneCATS} & Eight models & Aerial & Image & Waypoint & Direct \\
EBench~\citeyearpar{yao2026profiling} & \paperlink{https://arxiv.org/abs/2610.35832}\nocode & GPT-6 Astra & Manipulation & Image & Waypoint & Direct \\
\rowcolor{cmpstripe}Embodied Agents Take Control~\citeyearpar{zhou2026embodied} & \paperlink{https://arxiv.org/abs/2607.26148}\codelink{https://github.com/jianzhou0420/MIP} & Claude Opus 5 & Navigation & Image & Discrete move & Direct \\
\citet{espinola2026reasoning} & \paperlink{https://arxiv.org/abs/2610.30051}\nocode & Six models & Manipulation & Image & Pose & Direct \\
\rowcolor{cmpstripe}GPT-as-Policy~\citeyearpar{team2026systematically} & \paperlink{https://arxiv.org/abs/2609.38537}\nocode & GPT-6 Astra & Manipulation, navigation, legged & Image & Pose & Direct \\
GPT-Policy~\citeyearpar{cheng2026context} & \paperlink{https://arxiv.org/abs/2609.19138}\codelink{https://github.com/cheng-haha/GPT-Policy} & GPT-6 Astra & Manipulation & Image & Pose & Direct \\
\rowcolor{cmpstripe}HumanCLAW~\citeyearpar{siyao2026humanclaw} & \paperlink{https://arxiv.org/abs/2607.27180}\nocode & Seven models & Legged & Image & Skill call & Direct \\
LIBERO-Agent~\citeyearpar{diao2026libero} & \paperlink{https://arxiv.org/abs/2609.39507}\codelink{https://github.com/dzj441/Libero-Agent} & Seven models & Manipulation & Image & Pose & Direct \\
\rowcolor{cmpstripe}MotorMind~\citeyearpar{li2026motormind} & \paperlink{https://arxiv.org/abs/2609.38078}\nocode & Qwen3.8-Flash-Next & Manipulation & Image & Discrete move & Direct \\
NavGPT-3~\citeyearpar{zhou2026navgpt} & \paperlink{https://arxiv.org/abs/2610.10787}\nocode & GPT-6 Astra & Navigation & Image & Waypoint & Direct \\
\rowcolor{cmpstripe}OnFly~\citeyearpar{zheng2026onfly} & \paperlink{https://arxiv.org/abs/2603.10682}\codelink{https://github.com/Robotics-STAR-Lab/OnFly} & Qwen3-VL-4B & Aerial & Image & Waypoint & Direct \\
OpenRUA~\citeyearpar{chu2026openrua} & \paperlink{https://arxiv.org/abs/2610.02459}\codelink{https://github.com/terminalworld/OpenRUA} & Claude Opus 5 & Manipulation & Image & Program & Direct \\
\rowcolor{cmpstripe}RoboColiseum~\citeyearpar{huang2026frontiera} & \paperlink{https://arxiv.org/abs/2610.27865}\nocode & Four models & Manipulation & Image & Pose & Direct \\
RoboDawn~\citeyearpar{guo2026transferring} & \paperlink{https://arxiv.org/abs/2609.22966}\nocode & Gemini 3.8 Flash, GPT-6 Astra & Manipulation & Marked image & Discrete move & Direct \\
\rowcolor{cmpstripe}RoboICL~\citeyearpar{liu2026roboicl} & \paperlink{https://arxiv.org/abs/2609.34261}\nocode & GPT-6 Astra & Manipulation & Image & Action chunk & Direct \\
RoboMME study~\citeyearpar{jedlicki2026frontier} & \paperlink{https://arxiv.org/abs/2610.17350}\codelink{https://github.com/jackbjed/embodied-astra-robomme} & GPT-6 Astra & Manipulation & Image & Waypoint & Direct \\
\rowcolor{cmpstripe}RoboQuest~\citeyearpar{renhang2026roboquest} & \paperlink{https://arxiv.org/abs/2610.10388}\nocode & Five models & Mobile manipulation & Image & Pose & Direct \\
RobotWorld~\citeyearpar{yang2026robotworld} & \paperlink{https://arxiv.org/abs/2610.10409}\nocode & Five models & Manipulation and other embodiments & Image & Pose & Direct \\
\rowcolor{cmpstripe}ROSClaw~\citeyearpar{cardenas2026rosclaw} & \paperlink{https://arxiv.org/abs/2603.26997}\nocode & Claude Opus 4.6 & Navigation, legged & Image & Velocity & Direct \\
\citet{sensenova2026has} & \paperlink{https://arxiv.org/abs/2610.22471}\nocode & GPT-6 Astra & Manipulation, humanoid, navigation & Image & Pose & Direct \\
\rowcolor{cmpstripe}Show-Harness~\citeyearpar{chen2026show} & \paperlink{https://arxiv.org/abs/2609.10522}\nocode & Gemini 3.1 Pro & Manipulation & Image & Discrete move & Direct \\
\citet{yang2026gpt} & \paperlink{https://arxiv.org/abs/2610.26846}\nocode & GPT-6 Astra & Manipulation & Image, touch & Pose & Direct \\
\rowcolor{cmpstripe}\citet{zhou2026large} & \paperlink{https://arxiv.org/abs/2610.28259}\nocode & GPT-6 Astra & Manipulation & Image & Pose & Direct \\
DiscussNav~\citeyearpar{long2023discuss} & \paperlink{https://arxiv.org/abs/2309.11382}\nocode & GPT-4 & Navigation & Text & Waypoint & Tools \\
\rowcolor{cmpstripe}MOKA~\citeyearpar{liu2024moka} & \paperlink{https://arxiv.org/abs/2403.03174}\nocode & GPT-4V & Manipulation & Marked image & Waypoint & Tools \\
PIVOT~\citeyearpar{nasiriany2024pivot} & \paperlink{https://arxiv.org/abs/2402.07872}\nocode & GPT-4V & Manipulation, navigation & Marked image & Waypoint & Tools \\
\rowcolor{cmpstripe}ReMEmbR~\citeyearpar{anwar2024remembr} & \paperlink{https://arxiv.org/abs/2409.13682}\nocode & GPT-4o & Navigation & Map & Waypoint & Tools \\
Robo-Harness K1~\citeyearpar{li2026robo} & \paperlink{https://arxiv.org/abs/2609.29389}\nocode & Gemini 3.7 Flash & Manipulation & Image & Pose & Tools \\
\rowcolor{cmpstripe}RoboHarness~\citeyearpar{qu2026roboharness} & \paperlink{https://arxiv.org/abs/2610.14018}\nocode & Qwen3.8-Flash-Next & Mobile manipulation & Image & Pose & Tools \\
RobotUse~\citeyearpar{lee2026robotuse} & \paperlink{https://arxiv.org/abs/2610.04929}\nocode & Gemini 3.8 Flash & Manipulation & Image & Pose & Tools \\
\rowcolor{cmpstripe}Spatial-Semantic MCP Navigation~\citeyearpar{lee2026spatial} & \paperlink{https://arxiv.org/abs/2609.27340}\nocode & Claude Sonnet 4.6 & Navigation & Map & Waypoint & Tools \\
SpatialHarness~\citeyearpar{wang2026spatialharness} & \paperlink{https://arxiv.org/abs/2610.12457}\nocode & GPT-6 Astra & Manipulation & Image & Pose & Tools \\
\rowcolor{cmpstripe}uni-robot-agent~\citeyearpar{yuan2026democratizing} & \paperlink{https://arxiv.org/abs/2611.01560}\nocode & Qwen3.8-Flash-Next & Manipulation & Image & Pose & Tools \\
VIA~\citeyearpar{hu2026visual} & \paperlink{https://arxiv.org/abs/2607.11119}\nocode & GPT-6 Astra & Manipulation & Marked image & Pose & Tools \\
\rowcolor{cmpstripe}WAA~\citeyearpar{zhang2026world} & \paperlink{https://arxiv.org/abs/2609.29964}\nocode & Gemini 3.7 Flash & Manipulation & Image & Pose & Tools \\
ASPIRE$^{*}$~\citeyearpar{lu2026aspire} & \paperlink{https://arxiv.org/abs/2607.00272}\nocode & Claude Opus 4.6 & Manipulation & Image & Program & Skills \\
\rowcolor{cmpstripe}COAP$^{*}$~\citeyearpar{hu2026embodied} & \paperlink{https://arxiv.org/abs/2610.12369}\nocode & GPT-6 Astra & Manipulation & Image & Program & Skills \\
EmbodiedEvo$^{*}$~\citeyearpar{li2026embodiedevo} & \paperlink{https://arxiv.org/abs/2610.18867}\nocode & GPT-6 Astra & Manipulation & Image & Program & Skills \\
\rowcolor{cmpstripe}EmbodiedSkillBench~\citeyearpar{han2026embodied} & \paperlink{https://arxiv.org/abs/2610.16238}\codelink{https://github.com/HanZhi306/EmbodiedSkill-Bench} & Four models & Manipulation & Image & Skill call & Skills \\
\citet{he2026robot}$^{*}$ & \paperlink{https://arxiv.org/abs/2609.31770}\codelink{https://github.com/hesd10/astra-robot-sim2real} & GPT-6 Astra & Manipulation & Image & Program & Skills \\
\rowcolor{cmpstripe}HexaShell$^{*}$~\citeyearpar{gao2026hexashell} & \paperlink{https://arxiv.org/abs/2610.23479}\codelink{https://github.com/HexaFuture/HexaShell} & GPT-6 Astra & Manipulation & Image & Program & Skills \\
PhysEvo$^{*}$~\citeyearpar{tian2026physevo} & \paperlink{https://arxiv.org/abs/2610.08995}\nocode & GPT-6 Astra & Manipulation & Image & Program & Skills \\
\rowcolor{cmpstripe}PhysRSI-eval~\citeyearpar{li2026physical} & \paperlink{https://arxiv.org/abs/2610.26848}\nocode & Six models & Manipulation & Image & Program & Skills \\
RACaP~\citeyearpar{li2026racap} & \paperlink{https://arxiv.org/abs/2609.29394}\nocode & GPT-5.6 & Manipulation & Image & Skill call & Skills \\
\rowcolor{cmpstripe}RAX$^{*}$~\citeyearpar{huang2026scaling} & \paperlink{https://arxiv.org/abs/2610.22496}\nocode & Claude Opus 5.5 & Manipulation & Image & Program & Skills \\
ReTrace~\citeyearpar{yu2026retrace} & \paperlink{https://arxiv.org/abs/2610.15814}\nocode & GPT-6 Astra, GPT-6 Sol & Manipulation & Image & Pose & Skills \\
\rowcolor{cmpstripe}RPG$^{*}$~\citeyearpar{wang2026reconstruct} & \paperlink{https://arxiv.org/abs/2610.02204}\nocode & GPT-6 Astra & Manipulation & Image & Skill call & Skills \\
Scale and Selection$^{*}$~\citeyearpar{wei2026scale} & \paperlink{https://arxiv.org/abs/2609.39304}\nocode & GPT-5.6 Sol & Manipulation & Image & Pose & Skills \\
\rowcolor{cmpstripe}SEAR~\citeyearpar{wang2026sear} & \paperlink{https://arxiv.org/abs/2610.18792}\nocode & Seven models & Manipulation & Image & Program & Skills \\
SimEX$^{*}$~\citeyearpar{hu2026simex} & \paperlink{https://arxiv.org/abs/2609.38982}\nocode & Claude Fable 5.1 & Manipulation & Image & Program & Skills \\
\rowcolor{cmpstripe}Skill2Real~\citeyearpar{he2026skill2real} & \paperlink{https://arxiv.org/abs/2610.02788}\nocode & GPT-6 Astra, Claude Opus 5 & Manipulation & Image & Program & Skills \\
Skill-RSI~\citeyearpar{bai2026compile} & \paperlink{https://arxiv.org/abs/2610.25114}\nocode & Claude Opus, GPT-6 & Manipulation & Image & Program & Skills \\
\rowcolor{cmpstripe}Teach and Grow~\citeyearpar{nie2026teach} & \paperlink{https://arxiv.org/abs/2608.17209}\nocode & GPT-6 Astra & Manipulation & Image & Skill call & Skills \\
URAI$^{*}$~\citeyearpar{ge2026make} & \paperlink{https://arxiv.org/abs/2609.39018}\nocode & Claude Fable 5.1 & Manipulation & Image & Program & Skills \\
\end{xltabular}
\endgroup

\needspace{16\baselineskip}
\subsection{Multi-Robot Collaboration}
\label{sec:others}
\label{sec:others-multirobot}

\begin{takeawaybox}
General-purpose models coordinate robots that differ in what they see, reach or do, and replan from execution feedback within a
task. The cleanest same-task comparison of architectures finds a success--cost trade-off, not a winner. The evidence is mostly
simulated; whether retained coordination knowledge helps later tasks is largely untested.
\end{takeawaybox}

\begin{figure}[H]
    \centering
    \includegraphics[width=\linewidth]{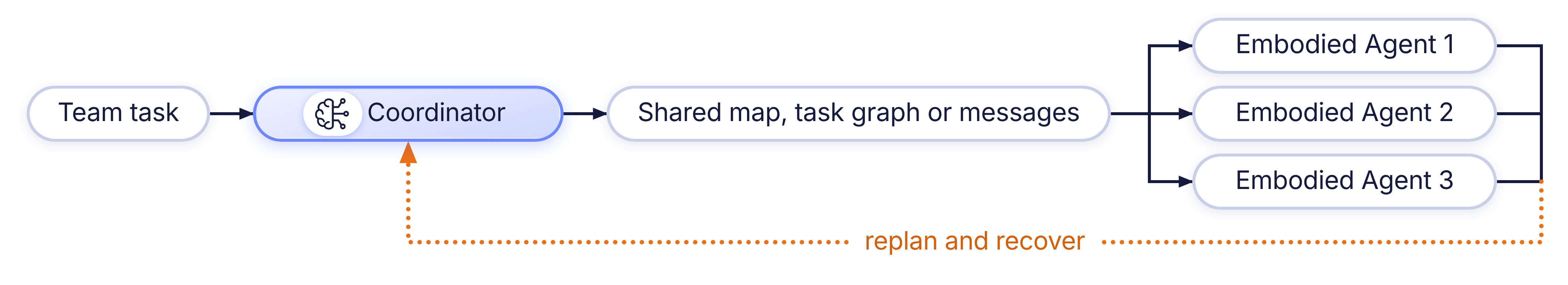}
    \caption{\textbf{Multi-robot collaboration.} A coordinator directs a team of embodied agents through a shared map, a task graph or messages, and execution feedback lets it replan and recover within the task.}
    \label{fig:subfig-others}
\end{figure}

Multi-robot collaboration is the coordination of two or more robots by general-purpose models, through a central planner or
per-robot agents: arms sharing a workspace, robots with complementary skills, or a drone guiding a ground vehicle. A team pools what
its members see and can do but faces partial observation, differing capabilities, dependent sub-tasks, the cost of every
message and model call, and incomplete execution feedback. Three overlapping questions follow: who decides and how work is assigned
(\Cref{sec:others-coordination}), what maps and task graphs the robots share (\Cref{sec:others-information}), and how a team recovers
and retains experience (\Cref{sec:others-recovery}). Code as Policy (\Cref{sec:code}) reviews the multi-agent planners
AutoTAMP~\citep{chen2023autotamp} and GenSwarm~\citep{ji2025genswarm}.

\subsubsection{Coordination and Task Allocation}
\label{sec:others-coordination}
\label{sec:others-negotiation}
\label{sec:others-allocation}
\label{sec:others-architectures}
Early work explored centralised and decentralised coordination. In RoCo, agents of arms sharing a workspace but differing in what
they observe and reach negotiate sub-tasks and waypoints, which inverse kinematics and collision checks validate; removing this feedback lowers success on most tasks, so part of the result is owed to validation,
not negotiation~\citep{mandi2023roco}. Its centralised baseline changes observation as well as architecture: a central
planner with full observation succeeds more often on four of six tasks~\citep{mandi2023roco}. CoELA builds each of two simulated
household agents around a frozen model with memory, planning and communication modules; each sees part of the house, and every
message costs time. Removing memory or weakening the planner costs steps, whereas disabling communication causes no substantial drop in the tested setting~\citep{zhang2023building}. Chen et al.\ hold task
information fixed and compare centralised planning, decentralised dialogue and two hybrids, one adding robot-local feedback to central
proposals~\citep{chen2023scalable}. In 2D tasks with up to 32 robots this feedback hybrid succeeds most often while centralised
planning uses fewer calls and tokens among successful runs; with three and six PyBullet arms both reach 100\% over ten runs, the hybrid with more calls and
tokens~\citep{chen2023scalable}. The result is a success--cost trade-off under these conditions, not a winner.

Assignment depends on what the planner knows about each robot. SMART-LLM writes code assigning sub-tasks to individual robots or
coalitions from hand-written skill descriptions, reaching 71\% success on complex tasks against 0\% for random
allocation~\citep{kannan2023smart}. EMOS derives capability descriptions from URDF summaries and kinematics tools; a central planner
assigns work and robot-local agents check feasibility and execute predefined skills, with 37.82\% success over 519 simulated episodes
against 23.56\% without the numerical capabilities, assuming ideal SLAM and contact-based grasp attachment~\citep{chen2024emos}. Conformal prediction lets a
robot act alone when one candidate action remains and otherwise ask teammates, then a human; the guarantee assumes perfect skill
execution in a known, static environment and is tested only in simulation, establishing neither physical safety nor reliable
recovery~\citep{wang2024probabilistically}.

\subsubsection{Shared Information and Task Scheduling}
\label{sec:others-information}
\label{sec:others-spatial}
\label{sec:others-scheduling}
Robots with different viewpoints coordinate through a shared map, and robots with dependent sub-tasks through a task graph.
Co-NavGPT has a language model read the robots' merged semantic map, positions and unexplored frontiers and assign
each robot a frontier, in simulated object-goal navigation~\citep{yu2023co}. Air-Ground VLN passes a drone's overhead view to a ground
vehicle with only a first-person view: a frozen VLM locates the target in the overhead image against a reference photograph, the
drone marks the vehicle and the target on it, and the vehicle plans a road path from this map~\citep{zhang2026air}. Counting a
joint success when either robot reaches the target, the full system succeeds in 77.0\% of episodes, against 12.0\% for the UGV-only VLM baseline and
60.0\% for the independent UAV-only 3D-SPF baseline, while sharing an unannotated overhead image gives
65.0\% joint success~\citep{zhang2026air}. The annotations add to success, but the gain is not cleanly divided between them and the drone's
viewpoint.

LiP-LLM combines model-generated skill dependencies with linear-programming allocation, and DART-LLM links dependencies to predefined
robot functions~\citep{obata2024lip,wang2024dart}; OSDAG dispatches ready tasks online, checking precedence, availability,
reachability and manipulation state~\citep{canh2026osdag}. Without dependency representations, success falls on tasks with complex
dependencies~\citep{wang2024dart,canh2026osdag}. This machinery enforces only the constraints it represents, not every physical
precondition and outcome.

\subsubsection{Execution Feedback and Team Recovery}
\label{sec:others-recovery}
RoCo's dialogue re-proposes when its plan checks fail. LLaMAR splits planning, acting, correction and verification among
model roles on its partially observed, long-horizon MAP-THOR benchmark; the simulator reports whether each
high-level action succeeded, and only subtask completion is the model's own judgement~\citep{nayak2024llamar}. In COHERENT, robot
executors report infeasibility, failure or partial progress to a central assigner that revises sub-tasks; feedback in simulation
comes from validation rules and a scene graph, and one physical task is demonstrated~\citep{liu2024coherent}. CoMuRoS adds concurrent
event monitoring to a task manager and local code-generating agents; in separate hardware scenarios, recovery is autonomous in 9/10
trials and human-assisted in 5/5~\citep{borate2025llm}. Both rely on predefined skills and configured scenarios, and neither tests
whether retained recovery experience, such as CoMuRoS's chat history, improves later, held-out tasks. REBEL tests reuse in
simulated initial allocation for mixed human--robot teams: it retrieves model-written allocation rules and scored records of earlier
missions for an unseen mission, and removing either worsens each objective in its single-objective tests~\citep{gupte2024rebel}.

Coordination and execution feedback thus let teams adapt within a task, but simulator signals, chat history and retrieved rules fall
short of an improvement validated on later physical tasks; which coordination or recovery knowledge to retain, and how to verify its
benefit, remains open.

\begingroup\small\setlength{\tabcolsep}{5pt}\renewcommand{\arraystretch}{1.15}
\begin{xltabular}{\linewidth}{@{}>{\raggedright\arraybackslash}p{3.7cm} c >{\raggedright\arraybackslash}X >{\raggedright\arraybackslash}X >{\raggedright\arraybackslash}X >{\raggedright\arraybackslash}X@{\hspace{2pt}}}
\caption{\textbf{Comparison of works on multi-robot collaboration} (\Cref{sec:others}). Feedback: the signal that returns to the
planning model or scheduler, namely checks or assistance during planning, skill or action status after execution, observations or
events during execution, scheduling state, or records of earlier missions. The type of feedback does not by itself establish that
a physical outcome was verified independently of the planner, or that the system recovers from failure; for the works discussed in the text,
\Cref{sec:others-recovery} describes where each signal comes from.}\label{tab:cmp-4-6}\\
\toprule\rowcolor{cmphead} Work & Link & Decision structure & Shared via & Robots differ in & Feedback \\ \midrule \endfirsthead
\toprule\rowcolor{cmphead} Work & Link & Decision structure & Shared via & Robots differ in & Feedback \\ \midrule \endhead
\bottomrule \endlastfoot
CoELA~\citeyearpar{zhang2023building} & \paperlink{https://arxiv.org/abs/2307.02485}\nocode & Decentralised & Messages & View & Observations \\
\rowcolor{cmpstripe}RoCo~\citeyearpar{mandi2023roco} & \paperlink{https://arxiv.org/abs/2307.04738}\nocode & Decentralised & Messages & View & Plan checks (IK, collision) \\
S-ATLAS~\citeyearpar{wang2024probabilistically} & \paperlink{https://arxiv.org/abs/2402.15368}\nocode & Decentralised & Plan & -- & Help when uncertain (planning) \\
\rowcolor{cmpstripe}SMART-LLM~\citeyearpar{kannan2023smart} & \paperlink{https://arxiv.org/abs/2309.10062}\nocode & Centralised & Capabilities & Skills & None \\
EMOS~\citeyearpar{chen2024emos} & \paperlink{https://arxiv.org/abs/2410.22662}\nocode & Hybrid & Capabilities & Skills & Feasibility checks (planning) \\
\rowcolor{cmpstripe}\citet{chen2023scalable} & \paperlink{https://arxiv.org/abs/2309.15943}\nocode & Compared & Plan & Reach & Plan checks; state after each step \\
EFLUX~\citeyearpar{zhang2026eflux} & \paperlink{https://arxiv.org/abs/2607.12050}\nocode & Centralised & Plan & -- & Plan checks \\
\rowcolor{cmpstripe}CoEnv~\citeyearpar{kang2026coenv} & \paperlink{https://arxiv.org/abs/2604.05484}\nocode & Centralised & Plan & Reach & Plan checks; observations \\
DynaHMRC~\citeyearpar{yu2026dynahmrc} & \paperlink{https://arxiv.org/abs/2606.14882}\nocode & Decentralised & Messages & Skills, reach & Action status; observations \\
\rowcolor{cmpstripe}Co-NavGPT~\citeyearpar{yu2023co} & \paperlink{https://arxiv.org/abs/2310.07937}\nocode & Centralised & Map & View & Observations (map) \\
Air-Ground VLN~\citeyearpar{zhang2026air} & \paperlink{https://arxiv.org/abs/2609.03483}\codelink{https://github.com/ZSN2024/AGC-VLN} & Decentralised & Map & View & Observations \\
\rowcolor{cmpstripe}\mbox{D-VLC}~\citeyearpar{zhou2026d} & \paperlink{https://arxiv.org/abs/2607.29009}\nocode & Hybrid & Messages, map & Skills, reach & Observations; action status \\
DART-LLM~\citeyearpar{wang2024dart} & \paperlink{https://arxiv.org/abs/2411.09022}\nocode & Centralised & Task graph & Skills & Object map (skill parameters only) \\
\rowcolor{cmpstripe}LiP-LLM~\citeyearpar{obata2024lip} & \paperlink{https://arxiv.org/abs/2410.21040}\nocode & Centralised & Task graph & Skills & Skill completion \\
OSDAG~\citeyearpar{canh2026osdag} & \paperlink{https://arxiv.org/abs/2606.15255}\nocode & Centralised & Task graph & Reach & Scheduling state \\
\rowcolor{cmpstripe}\citet{he2026closed} & \paperlink{https://arxiv.org/abs/2607.06990}\nocode & Centralised & Task graph & Skills & Action status \\
FLEET~\citeyearpar{rivera2025fleet} & \paperlink{https://arxiv.org/abs/2510.07417}\nocode & Hybrid & Task graph & Skills & Scheduling state; events \\
\rowcolor{cmpstripe}COHERENT~\citeyearpar{liu2024coherent} & \paperlink{https://arxiv.org/abs/2409.15146}\codelink{https://github.com/MrKeee/COHERENT} & Hybrid & Plan & Skills & Action status (rules, scene graph) \\
LLaMAR~\citeyearpar{nayak2024llamar} & \paperlink{https://arxiv.org/abs/2407.10031}\codelink{https://github.com/nsidn98/LLaMAR} & Centralised & Status & View & Action status (simulator) \\
\rowcolor{cmpstripe}REBEL~\citeyearpar{gupte2024rebel} & \paperlink{https://arxiv.org/abs/2409.16266}\nocode & Centralised & Capabilities & Skills & Prior-mission records \\
MeCo~\citeyearpar{wang2026meco} & \paperlink{https://arxiv.org/abs/2601.20577}\codelink{https://github.com/TomWang-NPU/MeCo} & Compared & Plan & Reach & Plan checks; prior-mission records \\
\rowcolor{cmpstripe}CoMuRoS~\citeyearpar{borate2025llm} & \paperlink{https://arxiv.org/abs/2511.22354}\nocode & Hybrid & Plan & Skills & Events; task status \\
\end{xltabular}
\endgroup

\subsection{Discussion}
\label{sec:deployment-limits}
\label{sec:policy_execution_limitations}
\label{sec:others-limits}

Closed frontier models have demonstrated strong robot control on some tasks. Whether open-weight
frontier models can be trained to comparable capability remains open. On RoboDojo, Astra reached a 22.48\% average success rate over
2,100 trials under the official protocol of 50 episodes per task, compared with 1.92\% for the open-weight DeepSeek-Flash at
10 episodes per task~\citep{zhang2026unexpected}. The different episode budgets make this comparison indicative.
Closing this gap raises a broader question about training. The work reviewed in \Cref{sec:learning-frameworks} largely follows classic policy training.
Training the general-purpose model itself for embodied tasks is still at an early stage, with no established paradigm. PEAR
post-trains an open-weight reasoning model in simulation using task-success rewards~\citep{tang2026training}
(\Cref{sec:si-foundation-model}). One possible route is to adapt computer-use post-training to teach a vision--language model to
operate a robot, taking on the role of the robot-use agent (RUA) described by \citet{chu2026openrua}.
In deployment, a general-purpose model takes seconds per decision while contact and motion can change within milliseconds, and the
lower the level of the command, as in \Cref{sec:llmpolicy}, the more each action depends on a model response, a delay that simulators
hide when they pause during inference (\Cref{sec:latency}). High benchmark scores can hide failures in precise, contact-rich and
fast control, as well as the frequent cases in which the model judges an unfinished task as complete (\Cref{sec:failure}).
A deployed VLA incurs most of its cost upfront, and these systems incur a cost at each model call, with repeated calls and retries
multiplying inference costs and real-robot trials also requiring robot and supervisor time (\Cref{sec:cost}). Physical mistakes are
often irreversible, few of the reviewed systems use guardrails, and prompt injection and memory poisoning remain largely unexamined
(\Cref{sec:safety}).

\paragraph{EVA pathway.} In the pathway of \Cref{sec:vrea}, the embodied agent is what moves from the virtual environment to the real one. Policy training (\Cref{sec:learning-frameworks}) builds the abilities the agent brings when it first acts in the real world; Agent as System 2, Code as Policy and LLM as Policy (\Crefrange{sec:slowfast}{sec:llmpolicy}) are the ways those abilities reach the robot; and once the agent acts in the real world, monitoring and recovery tell which attempts succeed, so that its own operation produces real-world data that flow back to improve the virtual environment. As the agent advances, the pathway advances with it: better training gives the agent more to start with in the real world, more precise control lets more of it reach the robot, and reliable monitoring turns each real attempt into usable data. Better embodied agents thus let the agent enter the real world sooner and keep real data flowing back, so that improvement can continue in both environments at once.
\needspace{16\baselineskip}
\section{Towards Recursive Self-Improvement}
\label{sec:selfimprove}

In the pathway that EVA proposes (\Cref{sec:vrea}), self-improvement is what moves the agent from the virtual environment to the real one, improving both the agent and the environment it practises in along the way. \Cref{sec:sim,sec:embodied-agent} described the environment and the agent as they are; their progress becomes visible only when a system uses feedback from its own tasks to change itself. This section asks how existing robot systems improve themselves from their own experience, and how many of them also improve the way they improve (\Cref{sec:coevolution-roles}). Environment improvement (\Cref{sec:si-environment}) revises the virtual environment, a simulation or a learned world model; foundation model improvement (\Cref{sec:si-foundation}) updates the model’s weights; harness improvement (\Cref{sec:si-harness}) keeps the model fixed and changes what surrounds it: its memory (\Cref{sec:si-icl}), its tools and the way they are orchestrated, including the improvement process itself (\Cref{sec:si-tools}), and the policy it calls (\Cref{sec:si-model}); and embodiment improvement (\Cref{sec:si-embodiment}) changes the robot’s body or the tool it holds. \Cref{sec:si-limits} closes the section with a discussion of limitations and open questions. We include work whose check on a change is weaker than that of \Cref{eq:roles-self}, and the comparison tables record, for each work, what changes, what decides whether the change is kept and whether it is reused across tasks; systems of \Cref{sec:sim,sec:embodied-agent} that also keep and reuse their changes are reviewed here.

\subsection{Environment Improvement}
\label{sec:si-environment}

\begin{takeawaybox}
Real-world trials show where to correct a virtual environment and which failures to turn into practice tasks. The revised environment shapes the next round of practice and testing, and those results can guide further changes. Later real-world tests show whether the policy trained in the revised environment performs better.
\end{takeawaybox}

\begin{figure}[H]
    \centering
    \includegraphics[width=\linewidth]{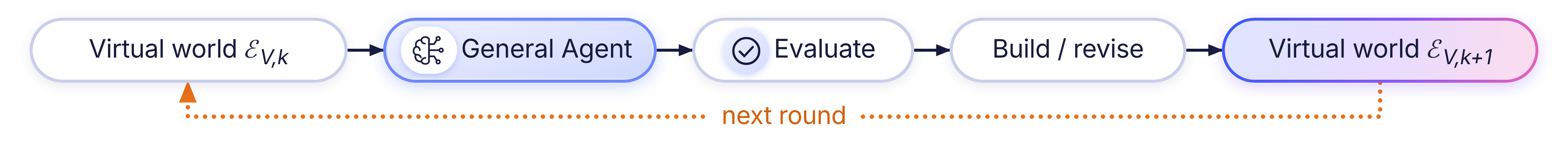}
    \caption{\textbf{Environment improvement.} A general agent evaluates the virtual world $\mathcal E_{V,k}$, for instance against real outcomes or by whether the policy still fails in it, and builds or revises it into $\mathcal E_{V,k+1}$, which starts the next round.}
    \label{fig:subfig-si-environment}
\end{figure}

A virtual environment gives a robotic system a place to explore, practise, and test its capabilities. In a fixed environment, the tasks available for practice stay the same as trials reveal new weaknesses. Environment improvement uses the results of those trials to revise the existing environment or build a new one. Once checked, the revised environment shapes what the system practises and how it is evaluated in the next round. The results of that round can guide further revisions, allowing the system and its environment to co-evolve.

Real-world trials can reveal two kinds of problems. When the same actions produce different outcomes in simulation and on the robot, the real outcome provides a reference for correcting the virtual environment and checking whether the revision closes the gap. When a trial exposes a weakness in the policy, reproducing that failure in a new scene gives the system a case to practise. To assess the benefit of the revised environment for training, the resulting policy can then be tested on subsequent real-world tasks.

How trial feedback changes the environment depends on how the environment is represented. \Cref{sec:si-env-sim} examines how feedback guides edits to a simulator or the construction of new scenes around failures. \Cref{sec:si-env-wm} examines how new execution data are used to retrain learned world models and update their predictions of what happens when the robot acts.

\subsubsection{Simulation Improvement}
\label{sec:si-env-sim}
A simulated environment can be revised from the agent's own trials in two ways: by editing the existing simulator or by building
new environments from the agent's failures. SimEX~\citep{hu2026simex} edits: after each physical trial, a coding agent corrects the
simulator until it reproduces the trial, and the corrected simulator screens repairs before the next one. F4R~\citep{yu2026f4r}
builds: it rebuilds each real failure as a tabletop environment, keeps the environment only if the current policy also fails there,
and retrains and redeploys the policy, whose new failures start the next round. The two routes keep a revision for different
reasons. SimEX keeps a correction when the simulator agrees with what happened on the robot, so its world grows more faithful; F4R
keeps an environment when the policy fails in it, so its worlds grow more useful for training. In both, the reason comes from the
agent's own trials, and this is what separates them from earlier ways of correcting a simulation.

\subsubsection{World Model Improvement}
\label{sec:si-env-wm}
A learned world model is revised by training it again on the policy's experience. The update is made by a fixed training procedure
rather than proposed by the agent, so these systems serve as background, and they differ mainly in where the experience comes from.
Most feed back rollouts of the policy itself: VLAW~\citep{guo2026vlaw} and World-Gymnast~\citep{sharma2026world} fine-tune the world
model on real rollouts and then train the policy further inside it, while World-VLA-Loop~\citep{liu2026world} and
WoVR~\citep{jiang2026wovr} add rollouts of each policy trained in the world model to its training data. WAV~\citep{liu2026worlda}
chooses its own data instead, executing the actions whose predictions are least consistent and training on what follows, and
CureWM~\citep{xu2026world} trains on perturbed demonstrations whose outcomes are verified by execution, so that it less often predicts
success for actions that fail.  What
these loops leave open is a decision on the world itself: a retrained model could be kept when the policy trained in it improves on
later real tasks and discarded otherwise, coupling the revision of the world with the training of the policy.

\begingroup\footnotesize\setlength{\tabcolsep}{6pt}\renewcommand{\arraystretch}{1.15}
\begin{xltabular}{\linewidth}{@{}>{\raggedright\arraybackslash}p{2.6cm} c >{\raggedright\arraybackslash}X >{\raggedright\arraybackslash}X >{\raggedright\arraybackslash}X@{\hspace{2pt}}}
\caption{\textbf{Environment improvement.} Feedback describes the data on which a revision is based. Update describes how the world is changed and the number or frequency of updates. Used for describes what the revised world serves afterwards. Upper group: the agent revises a simulation. Lower group, included as background: a fixed procedure retrains a world model. ---: none reported.}\label{tab:env-improvement}\label{tab:cmp-5-1}\\
\toprule\rowcolor{cmphead} Work & Link & Feedback & Update & Used for \\ \midrule \endfirsthead
\toprule\rowcolor{cmphead} Work & Link & Feedback & Update & Used for \\ \midrule \endhead
\bottomrule \endlastfoot
\multicolumn{5}{@{}l}{\textit{Simulation: the agent proposes each update}} \\
SimEX~\citeyearpar{hu2026simex} & \paperlink{https://arxiv.org/abs/2609.38982}\nocode & Own robot trials & Edit simulator & Repair screening \\
\rowcolor{cmpstripe}F4R~\citeyearpar{yu2026f4r} & \paperlink{https://arxiv.org/abs/2609.35575}\nocode & Real failures & Rebuild environment & Policy training \\
\midrule
\multicolumn{5}{@{}l}{\textit{World model (background): a fixed procedure retrains it}} \\
VLAW~\citeyearpar{guo2026vlaw} & \paperlink{https://arxiv.org/abs/2602.12063}\nocode & Real rollouts & Fine-tune (each round) & Policy training \\
\rowcolor{cmpstripe}World-Gymnast~\citeyearpar{sharma2026world} & \paperlink{https://arxiv.org/abs/2602.02454}\nocode & Real rollouts & Fine-tune (iterative) & Policy RL \\
World-VLA-Loop~\citeyearpar{liu2026world} & \paperlink{https://arxiv.org/abs/2602.06508}\nocode & New-policy rollouts & Fine-tune (each round) & Policy RL \\
\rowcolor{cmpstripe}WoVR~\citeyearpar{jiang2026wovr} & \paperlink{https://arxiv.org/abs/2602.13977}\codelink{https://github.com/RLinf/RLinf} & New-policy rollouts & Fine-tune (once) & Policy RL \\
WAV~\citeyearpar{liu2026worlda} & \paperlink{https://arxiv.org/abs/2604.01985}\nocode & Self-selected interactions & Retrain & Policy training \\
\rowcolor{cmpstripe}CureWM~\citeyearpar{xu2026world} & \paperlink{https://arxiv.org/abs/2610.09134}\codelink{https://github.com/jiuyixu25/CureWM} & Perturbed demos & Fine-tune (once) & --- \\
\end{xltabular}
\endgroup

\subsection{Foundation Model Improvement}
\label{sec:si-foundation}
\label{sec:si-foundation-model}

\begin{takeawaybox}
A foundation model can be retrained on trajectories it generated itself that pass a check, a loop that already works for tasks
whose answers can be verified. For robots, the loop depends on a check that reliably judges success or failure, and on retraining
without losing the model's general abilities.
\end{takeawaybox}

\begin{figure}[H]
    \centering
    \includegraphics[width=\linewidth]{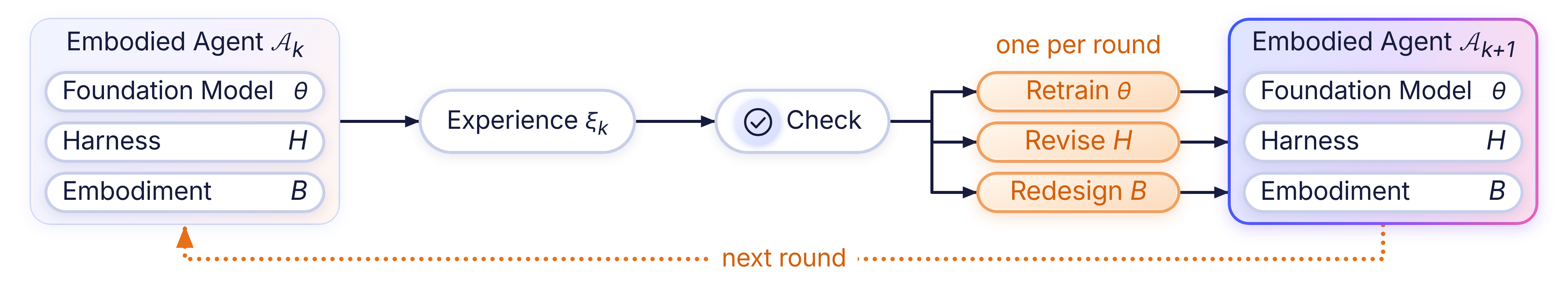}%
    \caption{\textbf{Improving one part of the agent.} The embodied agent $\mathcal A_k$ gains experience $\xi_k$, a check accepts what succeeded, and one of its parts is updated from it: the foundation model is retrained, the harness revised or the embodiment redesigned, while the other two stay unchanged; the resulting agent $\mathcal A_{k+1}$ runs the next round.}
    \label{fig:subfig-si-model}
\end{figure}

Foundation model improvement updates the parameters $\theta$ of the foundation model $F_\theta$ (\Cref{eq:agent-system}) on trajectories
that it generated itself and that a check accepts, and the updated model generates the next round.  The loop was established for language models on tasks whose answers can be checked automatically. STaR~\citep{zelikman2022star},
ReST~\citep{gulcehre2023reinforced} and ReST$^{EM}$~\citep{singh2023beyond} keep only the model's outputs that reach a known answer or
pass a program's tests and fine-tune the model on them. Agentic reinforcement learning extends the check to multi-turn tasks: it becomes
a reward for a whole interaction, and the model is updated while it acts~\citep{zhang2025landscape}.

Embodied agents now apply the same loop to the model that decides a robot's actions. Earlier systems either had people judge success,
as in LMPC~\citep{liang2024learning}, or rewarded task progress in simulation while the model acted through motion primitives or a
harness, as in VAGEN~\citep{wang2025vagen} and ZAP~\citep{pikus2026pretraining}. PEAR, a post-training engine for agentic robotics,
trains a 27B open-weight reasoning model with task-success rewards alone, entirely in simulation, without demonstrations and without a
harness~\citep{tang2026training}. Like a computer-use agent, the model sees camera frames and the robot's state, reasons in a chain of
thought and outputs the next end-effector pose, which inverse kinematics and a low-level controller execute; with no perception tools,
planners or skill libraries, it must itself locate objects, decide how to reach them and notice when a move fails. Training on LIBERO
raises its success from 1.6\% to 74.4\%, and on new tasks in familiar LIBERO-PRO scenes it succeeds in 57.4\% of episodes, where the
demonstration-trained $\pi_{0.5}$ succeeds in 0.75\%. The skills transfer to a dual-arm robot it never trained on, succeeding in 60\% of
real trials where the base model succeeds in none, and the model shows no measurable loss on nine reasoning, coding and vision
benchmarks.

\subsection{Harness Improvement}
\label{sec:si-harness}

\begin{takeawaybox}
With the model's parameters fixed, experience can be kept as memory, as tool and orchestration code, or as training data for the
policy. Whether a change is reused reliably in later tasks depends on checks independent of its proposer and on held-out cases;
changing the improvement process itself needs the same strict checks.
\end{takeawaybox}

Harness improvement keeps the parameters $\theta$ of the model fixed and revises the harness $H_k$ (\Cref{eq:harness}) from the
embodied agent's own feedback. It has three branches, reviewed below in this order: context improvement revises what is kept in memory and
read by the model as context; tool and orchestration improvement revises the code of the tools other than the policy and of the
orchestration; and policy model improvement trains the weights $\phi$ of the policy $\pi_\phi$, the one tool in $\mathcal T_k$ with
parameters learned for the robot. One system can change several parts, as Robo-COP does when it revises its context after each policy
update~\citep{zhang2026co}; it is reviewed under the part that its main contribution changes.

\subsubsection{Context Improvement}
\label{sec:si-icl}
Experience can be kept in long-term memory, which the model reads as context at its next inference, so that a model whose
parameters stay unchanged does better in later executions. Corrections, failure analyses and effective interventions are organised
into memory entries, examples, rules or instructions; later tasks retrieve them into the context to change planning and the use of
policies, and the outcomes of those tasks lead the system to add, correct, merge or delete what it has kept. Within the harness, the
boundary with tool and orchestration improvement, reviewed below, is whether what changes is the content kept in memory or the
code that processes it (the orchestration and tools). Kept knowledge can describe a particular situation, a regularity across tasks or how to use a policy, and a
system may keep all three; the paragraphs below follow this distinction.

\paragraph{Retaining experience across tasks.}
An early step was to make a correction useful beyond the attempt in which it was given. DROC distils a user's language corrections
into constraints, preferences and procedural knowledge after a task or skill, and retrieves them by textual and then visual
similarity when a new task or object instance is encountered~\citep{zha2023distilling}, so that a correction still guides the planner
as objects and task instances change. As experience is carried across more tasks, the conditions behind it also need to be kept.
RoboMemory separates episodic experience from semantic knowledge, and a model-based updater compares each new entry with similar
existing ones to add, update or remove entries~\citep{lei2025robomemory}. MessyMem attaches object properties and action outcomes
discovered through interaction, such as a drawer found to be locked, to a spatial scene graph with linked visual
keyframes~\citep{banwasi2026messymem}.

\paragraph{Consolidating and revising guidance.}
Accumulation also exposes a harder problem: deciding which conclusions from one attempt should guide another. A failure can have
several explanations, and reusing a mistaken one can reproduce the error. PhysMem separates raw experience, candidate hypotheses and
long-term principles: active hypotheses may inform planning, while the outcomes of later actions raise or lower their confidence and
decide which are promoted to principles~\citep{li2026physmem}, so later execution also tests the explanations kept from earlier
attempts. MEMO addresses a related difficulty, combining corrections from different users and tasks into more general guidance. It
stores textual corrections and the code of successful skills in a skillbook, which it clusters and rewrites offline into general
guidance and parameterised code templates for later skill generation~\citep{christie2026local}. On held-out simulation tasks,
it outperforms both a version of itself without clustering and DROC-V, an adapted DROC given the same corrections but without code
templates or clustering~\citep{christie2026local}.

\paragraph{Learning how to use a policy.}
When actions are delegated to a pretrained policy, experience can also teach the higher-level agent how to use that policy.
RoboHarness keeps successful and failed executions in separate memories, retrieves both to contrast them when a model attributes a
failure, and consolidates its execution traces offline into priors that guide later interventions on a frozen
VLA~\citep{li2026roboharness}. The kept knowledge describes the policy's failure conditions and the remedies available through the
harness's intervention tools, so its scope depends on the policy and the interface through which it was acquired. Guidance passed
from one agent to another must also suit the agent that reads it. In RHD, a strong agent writes a playbook for intervening on a
frozen VLA and then refines it from the execution feedback of a lighter agent that uses it, without updating the parameters of either
agent or of the VLA~\citep{kim2026recursive}. On held-out configurations, the initial playbook, not yet adapted, lowers the lighter
agent's success below that without a playbook, whereas the playbook refined on separate development configurations raises it
clearly~\citep{kim2026recursive}. The reader's own executions supply the evidence for adapting the guidance. Transfer can also succeed
without adaptation: in EmbodiedEvo, code Skills and diagnostic Lessons accumulated by a stronger model (Astra) and then frozen raise
the success rates of two weaker executors (Sol and Luna) on RoboDojo-Sim over no experience~\citep{li2026embodiedevo}.
A weaker reader can apply procedures and conditions that its own exploration failed to establish, but transfer depends on the task
conditions, the policy and the recipient.

Isolating the contribution of memory revision requires comparisons that keep the model and tools
fixed and change only whether the long-term memory is frozen or updated. SimplEvo approaches such a test: with no weight updates or
trained policy, GPT-6 Astra controlling the robot directly writes a task memory from one demonstration and refines a bounded working
memory with lessons after each training episode; with both memories frozen at test time, its success on EBench rises from 46.7\% to
81.7\%, and its working memory also improves weaker models that reuse it unchanged~\citep{yao2026review}. As experience grows, stale
facts, conflicting rules and advice tied to an older policy remain obstacles to sustained improvement.

\subsubsection{Tool and Orchestration Improvement}
\label{sec:si-tools}
Execution experience can also improve the orchestration and the tools other than the policy, which itself stays fixed, for instance
when a grasp routine approaches from an unsuitable direction, an orchestrator advances before a subtask finishes or a monitor overlooks
a slip; this complements the revision of long-term memory, reviewed above, and the training of the policy's weights, reviewed below. The central issue is how a useful repair becomes a reliable part of future execution. The three paragraphs below follow the
routes in turn: skills and programs, orchestration with its monitoring and recovery, and the improvement process that forms and accepts
the changes.

\paragraph{Skill and program refinement.}
Expressing a policy as code makes particular execution steps available for revision: Code as Policies composes perception and
control APIs into programs, CaP-X provides a common framework for executing and evaluating them, and PRoC3S checks geometric and
kinematic constraints before the robot acts~\citep{liang2022code,fu2026cap,curtis2024trust}. A repair reaches later tasks in the form
in which it is kept: as composable functions accumulated through practice, which RATs freezes to test
transfer~\citep{tziafas2024lifelong,zhang2026playful}; as skill guidance drawn from execution traces that informs later program
search~\citep{lu2026aspire}; as both code skills and diagnostic lessons, frozen after practice for runs in unseen
scenes~\citep{li2026embodiedevo}; or as task-specific tools with their own unit tests, written after a failed
episode~\citep{gao2026hexashell}. Each form helps only when feedback pinpoints a failure that the code can address, and EmbodiedRSI
makes this an active choice by selecting the physical experiments that best tell competing code and skill hypotheses
apart~\citep{song2026embodiedrsi}. The editable scope ranges from single functions to multi-file repositories and whole perception,
planning and control programs~\citep{elmaaroufi2026rho,huang2026scaling}; a wider scope gives more room for repair but makes
development cost, human assistance and final capability harder to separate. Some systems contain this risk by gating each change on
evaluation, on validation episodes kept apart from development in COAP and across practice tasks in RPG, or by confining a revision to
the skill to which an outcome is attributed, as in RoboRSI~\citep{hu2026embodied,wang2026reconstruct,wen2026roborsi}.

\paragraph{Orchestration refinement.}
Individually usable skills can still fail when composed into a long task, when a precondition is unmet, a subgoal advances too early
or an observation passed to the executor is stale. The conditions under which a skill is invoked can themselves be made revisable:
SayCan constrains high-level selection by what the robot can actually do, and CLIMB revises a planning domain, its predicates and their
grounding functions from discrepancies between expected and observed states~\citep{ahn2022i,byrnes2024climb}. Around a VLA, AdaHVLA
places subgoal progression, observation refresh, recovery and completion checks in editable coordination code, so that experience can
alter later calls~\citep{tang2026adahvla}, and outside robotics AFlow searches over workflows written as code, using execution feedback
on a validation set~\citep{zhang2024aflow}. Orchestration also covers monitoring and recovery: Inner Monologue and REFLECT return
success detection, scene feedback and summaries of execution to the planner, so that a failure can be explained and the plan
corrected~\citep{huang2022inner,liu2023reflect}, and Zetta keeps what a failure taught by revising critics, recovery policies and tools
across rounds while the base policy and the high-level orchestrator stay fixed~\citep{ding2026zetta}. More active detection can
interrupt valid behaviour, so completion rates need an independent task criterion and reporting of false interventions and recovery cost.

\paragraph{Improvement-process refinement.}
\label{sec:si-process}
The learning pipeline, made up of the reward code, training configurations and experimental procedures that produce a policy, is used
only when the agent improves; \citet{zhou2026who} place such components in the improvement procedure, and here the pipeline belongs to
the improvement process $\Phi_k$. Eureka, DrEureka and ENPIRE revise it from the results of training~\citep{ma2023eureka,ma2024dreureka,xiao2026enpire},
but each within one task, and they are compared with agent learning (\Cref{sec:learning-frameworks}); no reviewed system keeps a
revised pipeline and reuses it on later tasks, so learning-pipeline refinement, the route in which improvement would most directly
become recursive, is for robots still empty. Since $\Phi_k=(\mathcal P_k,\mathcal V_k)$ is part of the harness
(\Cref{sec:roles-self}), other harness changes reach it and make the improvement recursive, as \Cref{tab:cmp-5-3,tab:cmp-5-4} mark.
Several systems change how later changes are proposed ($\mathcal P_k$), through kept repair guidance, hypotheses that choose the next
experiments or guide later revisions, experience that feeds a task generator, diagnostic and review tools that a meta-agent builds for
itself, or a general system promoted to propose later
edits~\citep{lu2026aspire,song2026embodiedrsi,tang2026adahvla,tziafas2024lifelong,tian2026physevo,liang2026robofoundry}. MEMENTO changes
how changes are accepted ($\mathcal V_k$), by evolving the evaluator that scores candidate programs~\citep{sygkounas2026memento},
and beyond the harness Auto-Robotist keeps design rules that condition
later body proposals (\Cref{sec:si-embodiment})~\citep{wang2026search}. No reviewed system compares a changed improvement process with a
frozen copy from the same start and with the same budget. Scale and Selection shows how such a comparison can be built: it fixes the
proposer and varies only the rule that accepts a change, and champion--challenger selection raises held-out success
where unconditional acceptance does not~\citep{wei2026scale}. Of the systems that name recursion in their titles, EmbodiedRSI makes a
recursive update and RoboRSI comes close, while EmbodiRSI improves a policy within a fixed learning formulation (policy model improvement, below),
EmbodiedSmith revises tasks after failures of a fixed executor (\Cref{sec:learning-tasks}), RE-0 trains a student on a teacher's patches
(\Cref{sec:learning-frameworks}), and COAP and RHD keep changes to their code or memory without reaching the improvement process~\citep{hu2026embodied,kim2026recursive}.

\begingroup\footnotesize\setlength{\tabcolsep}{5pt}\renewcommand{\arraystretch}{1.15}
\begin{xltabular}{\linewidth}{@{}>{\raggedright\arraybackslash}p{2.6cm} c >{\raggedright\arraybackslash}X >{\raggedright\arraybackslash}X >{\raggedright\arraybackslash}X >{\raggedright\arraybackslash}X >{\raggedright\arraybackslash}X >{\raggedright\arraybackslash}X@{\hspace{2pt}}}
\caption{\textbf{Comparison of works on context improvement and on tool and orchestration improvement} (\Cref{sec:si-harness}; policy model improvement is compared in \Cref{tab:cmp-5-2}), including the changes that reach the improvement process, which is part of the harness. Changes: environment, weights, memory, tools and orchestration, the improvement process or the body. Formed by: a general model, a coding agent or a fixed training pipeline. Kept only if: what decides whether a change is kept (None: no gate). Reused: in later episodes of the same task or across tasks. Level: self-improvement, recursive (the change reaches the improvement process), or -- when the change is not kept and reused.}\label{tab:cmp-5-3}\\
\toprule\rowcolor{cmphead} Work & Link & Changes & Learned from & Formed by & Kept only if & Reused & Level \\ \midrule \endfirsthead
\toprule\rowcolor{cmphead} Work & Link & Changes & Learned from & Formed by & Kept only if & Reused & Level \\ \midrule \endhead
\bottomrule \endlastfoot
DROC~\citeyearpar{zha2023distilling} & \paperlink{https://arxiv.org/abs/2311.10678}\nocode & Memory & Corrections & Model & None & Across tasks & Self-impr. \\
\rowcolor{cmpstripe}RoboMemory~\citeyearpar{lei2025robomemory} & \paperlink{https://arxiv.org/abs/2508.01415}\nocode & Memory & Outcomes & Model & None & Across tasks & Self-impr. \\
MessyMem~\citeyearpar{banwasi2026messymem} & \paperlink{https://arxiv.org/abs/2609.15976}\nocode & Memory & Outcomes & Model & None & Across tasks & Self-impr. \\
\rowcolor{cmpstripe}MEMO~\citeyearpar{christie2026local} & \paperlink{https://arxiv.org/abs/2603.04560}\nocode & Memory, tools & Corrections & Model & Task success & Across tasks & Self-impr. \\
PhysMem~\citeyearpar{li2026physmem} & \paperlink{https://arxiv.org/abs/2602.20323}\nocode & Memory & Outcomes & Model & Later outcomes & Same task & Self-impr. \\
\rowcolor{cmpstripe}RHD~\citeyearpar{kim2026recursive} & \paperlink{https://arxiv.org/abs/2609.33378}\nocode & Memory & Outcomes & Model & Target reached & Same task & Self-impr. \\
RoboHarness~\citeyearpar{li2026roboharness} & \paperlink{https://arxiv.org/abs/2603.24060}\codelink{https://github.com/LZY-1021/RoboHarness} & Memory & Execution traces & Model & None & Across tasks & Self-impr. \\
\rowcolor{cmpstripe}SimplEvo~\citeyearpar{yao2026review} & \paperlink{https://arxiv.org/abs/2611.11509}\nocode & Memory & Demonstration, outcomes & Model & None & Same task & Self-impr. \\
LRLL~\citeyearpar{tziafas2024lifelong} & \paperlink{https://arxiv.org/abs/2406.18746}\nocode & Tools, memory, process & Outcomes & Model & Task success & Across tasks & Recursive \\
\rowcolor{cmpstripe}ASPIRE~\citeyearpar{lu2026aspire} & \paperlink{https://arxiv.org/abs/2607.00272}\nocode & Tools, process & Execution traces & Coding agent & Dev-set gain & Across tasks & Recursive \\
COAP~\citeyearpar{hu2026embodied} & \paperlink{https://arxiv.org/abs/2610.12369}\nocode & Tools & Execution traces & Coding agent & Dev-set gain, no regression & Across tasks & Self-impr. \\
\rowcolor{cmpstripe}EmbodiedEvo~\citeyearpar{li2026embodiedevo} & \paperlink{https://arxiv.org/abs/2610.18867}\nocode & Tools, memory & Outcomes & Coding agent & None & Same task & Self-impr. \\
EmbodiedRSI~\citeyearpar{song2026embodiedrsi} & \paperlink{https://arxiv.org/abs/2610.10498}\nocode & Tools, process & Outcomes & Model & Dev-set gain & Same task & Recursive \\
\rowcolor{cmpstripe}HexaShell~\citeyearpar{gao2026hexashell} & \paperlink{https://arxiv.org/abs/2610.23479}\codelink{https://github.com/HexaFuture/HexaShell} & Tools & Execution traces & Coding agent & Runs & Same task & Self-impr. \\
RATs~\citeyearpar{zhang2026playful} & \paperlink{https://arxiv.org/abs/2606.19419}\nocode & Tools, memory & Outcomes & Model & Task success & Across tasks & Self-impr. \\
\rowcolor{cmpstripe}RAX~\citeyearpar{huang2026scaling} & \paperlink{https://arxiv.org/abs/2610.22496}\nocode & Tools, memory & Outcomes & Coding agent & Runs & Same task & Self-impr. \\
RHO~\citeyearpar{elmaaroufi2026rho} & \paperlink{https://arxiv.org/abs/2606.16458}\nocode & Tools & Execution traces & Coding agent & Dev-set gain & Across tasks & Self-impr. \\
\rowcolor{cmpstripe}RoboRSI~\citeyearpar{wen2026roborsi} & \paperlink{https://arxiv.org/abs/2610.12424}\codelink{https://github.com/nssmd/RoboRSI} & Tools & Execution traces & Model & No regression & Across tasks & Self-impr. \\
RPG~\citeyearpar{wang2026reconstruct} & \paperlink{https://arxiv.org/abs/2610.02204}\nocode & Tools, memory & Execution traces & Model & Dev-set gain, no regression & Across tasks & Self-impr. \\
\rowcolor{cmpstripe}CLIMB~\citeyearpar{byrnes2024climb} & \paperlink{https://arxiv.org/abs/2410.13756}\nocode & Tools & Outcomes & Model & None & Across tasks & Self-impr. \\
AdaHVLA~\citeyearpar{tang2026adahvla} & \paperlink{https://arxiv.org/abs/2609.29204}\codelink{https://github.com/Haaareally/AdaHVLA-Adaptive_Harness_VLA} & Tools, process & Execution traces & Coding agent & Dev-set gain & Across tasks & Recursive \\
\rowcolor{cmpstripe}Zetta~\citeyearpar{ding2026zetta} & \paperlink{https://arxiv.org/abs/2608.16590}\nocode & Tools & Execution traces & Coding agent & Repairs failure & Same task & Self-impr. \\
PhysEvo~\citeyearpar{tian2026physevo} & \paperlink{https://arxiv.org/abs/2610.08995}\nocode & Tools, process & Execution traces & Model & Repairs failure & Across tasks & Recursive \\
\rowcolor{cmpstripe}MEMENTO~\citeyearpar{sygkounas2026memento} & \paperlink{https://arxiv.org/abs/2607.22832}\codelink{https://github.com/sygkounas/MEMENTO} & Tools, process & Outcomes & Model & Fitness gain & Same task & Recursive \\
RoboFoundry~\citeyearpar{liang2026robofoundry} & \paperlink{https://arxiv.org/abs/2609.32862}\nocode & Memory, tools, process & Execution traces & Model & Repairs failure & Across tasks & Recursive \\
\rowcolor{cmpstripe}AOR~\citeyearpar{kumar2026act} & \paperlink{https://arxiv.org/abs/2603.04466}\nocode & Tools & Execution traces & Coding agent & Runs & Same task & Self-impr. \\
ASENA~\citeyearpar{cheng2026asena} & \paperlink{https://arxiv.org/abs/2609.39207}\nocode & Memory, tools & Outcomes & Coding agent & None & Same task & Self-impr. \\
\rowcolor{cmpstripe}Scale and Selection~\citeyearpar{wei2026scale} & \paperlink{https://arxiv.org/abs/2609.39304}\nocode & Tools & Execution traces & Coding agent & Dev-set gain & Across tasks & Self-impr. \\
SHAPER~\citeyearpar{wang2026self} & \paperlink{https://arxiv.org/abs/2608.11350}\nocode & Memory, tools & Execution traces & Model & Runs & Across tasks & Self-impr. \\
\end{xltabular}
\endgroup

\subsubsection{Policy Model Improvement}
\label{sec:si-model}
The policy $\pi_\phi\in\mathcal T_k$ is a learned action tool in the agent's harness (\Cref{eq:harness}), so when the high-level
model organises the training of its weights $\phi$, the embodied agent improves one of its own parts, which is self-improvement as
defined in \Cref{sec:roles-self}. Where the agent calls a policy, as in Agent as System 2 (\Cref{sec:slowfast}), the policy is the
execution layer (System 1) that carries out the high-level model's semantic instructions; improving it lets those instructions be
executed more reliably and over more skills, and so improves the whole system (Code as Policy and LLM as Policy,
\Cref{sec:code,sec:llmpolicy}, do not call $\pi_\phi$). Closed loops of this kind are already mature without a general-purpose
model: reinforcement learning fine-tunes pretrained robot policies on their own
rollouts~\citep{hu2024flare,tan2025interactive,lu2025vla,li2025simplevla,chen2025texttt,guo2025improving}, often on shared large-scale RL systems
such as RLinf~\citep{yu2025rlinf,zang2025rlinf}, and learned world models
assist training, in DGAC and GigaBrain by shaping the training signal drawn from real trials~\citep{chen2026robot,team2026gigabrain},
in RECAP through a value estimate updated from autonomous attempts and human corrections~\citep{intelligence20250a}, in RISE by
moving part of the trial and error into a learned world model~\citep{yang2026rise}, and in VLAW through a world model, improved by
real rollouts, whose generated experience improves the VLA~\citep{guo2026vlaw}. In these loops what the policy practises, the reward
and which data are used are fixed by people; they are our reference line (the fixed-pipeline row of \Cref{tab:cmp-5-2}). The steps that people
decide can be handed to the high-level model at two points, while the training algorithm itself stays fixed and the model decides
what is practised and which data are used. In \textbf{policy reinforcement learning} the model decides what the policy practises and
the reward it practises for, and reinforcement learning then updates the policy: ELLM proposes exploration goals from the current
state~\citep{du2023guiding}, BOSS guides the practice of chains of skills~\citep{zhang2023bootstrap}, FIND turns a VLM's judgement of
outcomes into feedback for online reinforcement learning of a residual policy~\citep{fang2026find}, and SEES builds targeted
reinforcement learning tasks in simulation for skills that fail repeatedly~\citep{li2026sees}, while CurricuLLM and Eurekaverse,
which deliver curricula or environments, are compared in \Cref{sec:learning-tasks}. In \textbf{trajectory-supervised policy
consolidation} the model monitors, repairs and verifies execution and selects and labels segments as supervision for supervised
fine-tuning, behaviour cloning or advantage-conditioned policy extraction: RoboClaw fine-tunes its manipulation policies iteratively
on data from task execution and reverse resets~\citep{li2026roboclaw}, Skill-Space Shooting trains verified repair segments into the
next task policy~\citep{rui2026skill}, EmbodiRSI trains later policies on experience acquired after execution
failures~\citep{lang2026embodirsi}, and Robo-COP trains on effective segments, including the correctly completed parts of failed
tasks, and also revises its context (\Cref{sec:si-icl})~\citep{zhang2026co}.

\begingroup\footnotesize\setlength{\tabcolsep}{5pt}\renewcommand{\arraystretch}{1.25}
\begin{xltabular}{\linewidth}{@{}>{\raggedright\arraybackslash}p{1.9cm} >{\hsize=1.4\hsize\raggedright\arraybackslash}X >{\hsize=1.2\hsize\raggedright\arraybackslash}X >{\hsize=.65\hsize\raggedright\arraybackslash}X >{\hsize=.75\hsize\raggedright\arraybackslash}X@{\hspace{2pt}}}
\caption{\textbf{Comparison of works on policy model improvement} (\Cref{sec:si-model}), the branch of harness improvement that trains the weights of the policy, a learned tool; the systems that update the foundation model itself (LMPC, VAGEN, ZAP and PEAR) are described in \Cref{sec:si-foundation}. Rows: what forms the change, a fixed training pipeline or a general model together with it; the fixed-pipeline works are reference work outside the reviewed corpus (\Cref{sec:scope-corpus}). Columns: what the weights are learned from. Each work changes the weights, keeps the change with no gate and reuses it in later episodes of the same task, which makes it self-improvement, unless marked: $^{\circ}$ reused across tasks; $^{m}$ also revises memory; $^{g}$ keeps a change only if it passes a gate (SEES: it repairs the failure; Robo-COP: no regression).}\label{tab:cmp-5-2}\\
\toprule\rowcolor{cmphead} Formed by / Learned from & Sim rollouts & Real rollouts & Imagined rollouts & Sim and real rollouts \\ \midrule \endfirsthead
\toprule\rowcolor{cmphead} Formed by / Learned from & Sim rollouts & Real rollouts & Imagined rollouts & Sim and real rollouts \\ \midrule \endhead
\bottomrule \endlastfoot
Fixed pipeline & FLaRe~\citeyearpar{hu2024flare}, iRe-VLA$^{\circ}$~\citeyearpar{guo2025improving}, $\pi_{\mathrm{RL}}$~\citeyearpar{chen2025texttt}, RIPT-VLA~\citeyearpar{tan2025interactive}, SimpleVLA-RL~\citeyearpar{li2025simplevla}, VLA-RL~\citeyearpar{lu2025vla} & $\pi^{*}_{0.6}$ (RECAP)~\citeyearpar{intelligence20250a}, DGAC~\citeyearpar{chen2026robot}, GigaBrain-0.5M*~\citeyearpar{team2026gigabrain} & RISE~\citeyearpar{yang2026rise} & -- \\
\rowcolor{cmpstripe}Model and pipeline & BOSS$^{\circ m}$~\citeyearpar{zhang2023bootstrap}, ELLM$^{\circ}$~\citeyearpar{du2023guiding}, SEES$^{\circ g}$~\citeyearpar{li2026sees}, EmbodiRSI~\citeyearpar{lang2026embodirsi} & FIND~\citeyearpar{fang2026find}, RoboClaw~\citeyearpar{li2026roboclaw}, Skill-Space Shooting~\citeyearpar{rui2026skill} & -- & \mbox{Robo-COP$^{mg}$} \citeyearpar{zhang2026co} \\
\end{xltabular}
\endgroup

\subsection{Embodiment Improvement}
\label{sec:si-embodiment}

\begin{takeawaybox}
A model can propose a robot's body or the tool it holds and evaluate the design by training a controller for it or by executing a
plan with it.
\end{takeawaybox}

Here the model changes the agent's embodiment $B_k$ (\Cref{eq:agent-system}), the robot's body or the tool it holds, while the task
and its surroundings stay fixed. A proposed design is scored by training a controller for it or by executing a plan with it, usually
in simulation, and the best designs, in some systems together with rules drawn from their evaluations, guide the next round of
proposals. The reviewed systems design for one fixed task under a criterion fixed in advance, with the proposer, scorer and acceptance
rule unchanged, and no design is reused in a later task: each delivers a design rather than keeping a change to the agent, so by the
definition of \Cref{sec:roles-self} most are design search
rather than self-improvement. We keep them because they are the only line in which a general model changes the robot's body, and the
body matters: on the simulated Robots' Last Exam benchmark, the best of nine models completes 31.25\% of full workplace workflows with
free-floating grippers, but none completes any once tied to the body of a dual-arm robot~\citep{rle2026skills}.

In body design, the model proposes legged, wheeled, soft or continuum robots, alone (RoboMorph~\citep{qiu2024robomorph}) or together with the
reward that trains their controller (RoboMoRe~\citep{fang2025robomore}, Debate2Create~\citep{qiu2025debate2create},
LACE-CRAFT~\citep{wen2026lace}). In tool design, the model proposes a tool together with the plan or trajectory that uses it
(VLMgineer~\citep{gao2025vlmgineer}, RobotSmith~\citep{lin2025robotsmith}). Passing the checks that such systems are built to satisfy
says little about whether the design does its job: 96.2\% of AID-SR's continuum robots pass its feasibility check, yet only 26.7\%
complete their task after training~\citep{chen2026bridging}. Auto-Robotist is the exception in one respect: it keeps written design
rules from earlier evaluations in a library that conditions later proposals, also in a later, larger run of the same
task~\citep{wang2026search}, so its rule library is a change to the improvement process (\Cref{sec:si-process}).

An open direction is an embodiment that enters the same loop of execution, feedback and kept improvement as the other parts of the
agent. A future system could choose or switch its own configuration during execution to suit each sub-task, using, for example, a
dexterous hand for in-hand or fine manipulation and a parallel gripper for simple pick-and-place, and learn from the outcomes of its
own executions which configuration suits which action and when a switch pays off. The hardware for such switching exists: in
Any-ttach, a vision--language model decomposes an instruction into a sequence of tool--skill pairs, and an automatic mechanism swaps
the end-effector, including a low-cost anthropomorphic hand, between skills~\citep{ni2026any}. There, however, the choice of tool is
planned rather than learned from the agent's own outcomes, nothing is kept between tasks, and the system is therefore not compared in
\Cref{tab:cmp-5-4}; no reviewed system yet keeps and revises such choices from its
experience.

\begingroup\footnotesize\setlength{\tabcolsep}{5pt}\renewcommand{\arraystretch}{1.25}
\begin{xltabular}{\linewidth}{@{}>{\raggedright\arraybackslash}p{1.9cm} >{\hsize=1.5\hsize\raggedright\arraybackslash}X >{\hsize=.7\hsize\raggedright\arraybackslash}X >{\hsize=.8\hsize\raggedright\arraybackslash}X@{\hspace{2pt}}}
\caption{\textbf{Comparison of works on embodiment improvement} (\Cref{sec:si-embodiment}). Rows: what forms the change, a general model alone or together with a fixed training pipeline. Columns: what the design is scored on. RobotSmith and VLMgineer change the tool and the others the body, $^{r}$ together with the reward that trains its controller. Each design is searched for one fixed task and is not reused; only Auto-Robotist$^{d}$ keeps something, its design rules, for later proposals in the same task, a change to the improvement process that makes it recursive (\Cref{sec:si-process}). Designs are kept on their fitness, except in AID-SR (checks and a judge), LACE-CRAFT (an incumbent test) and RobotSmith (a critic and fitness).}\label{tab:cmp-5-4}\\
\toprule\rowcolor{cmphead} Formed by / Learned from & RL training & Feasibility checks & Plan execution \\ \midrule \endfirsthead
\toprule\rowcolor{cmphead} Formed by / Learned from & RL training & Feasibility checks & Plan execution \\ \midrule \endhead
\bottomrule \endlastfoot
Model and pipeline & RoboMorph~\citeyearpar{qiu2024robomorph}, Debate2Create$^{r}$~\citeyearpar{qiu2025debate2create}, RoboMoRe$^{r}$~\citeyearpar{fang2025robomore} & -- & -- \\
\rowcolor{cmpstripe}Model & LACE-CRAFT$^{r}$~\citeyearpar{wen2026lace}, Auto-Robotist$^{d}$~\citeyearpar{wang2026search} & AID-SR~\citeyearpar{chen2026bridging} & RobotSmith~\citeyearpar{lin2025robotsmith}, VLMgineer~\citeyearpar{gao2025vlmgineer} \\
\end{xltabular}
\endgroup

\subsection{Discussion}
\label{sec:si-limits}

\paragraph{Limitations and open questions.} Self-improvement has helped robots perform tasks and improve the environments they
practise in, but the gains are still uneven. For environment improvement, the biggest hurdle is the gap between simulation and
the real world. Running the same policy in both can reveal where the simulator needs work. Exploratory actions such as opening
doors and pulling drawers can expose missing interactions or objects that respond differently in simulation. The revised
simulator can then help train a better policy, which can test the simulator again. For foundation model improvement, learning
from the model’s own trials requires knowing which attempts succeeded and making sure that learning new tasks does not erase
existing ability. Early work on foundation model improvement suggests that smaller general-purpose models can learn to operate robots
by training on their own trials. The approach is still in its early stages, and training can cost more than training a
vision–language–action (VLA) policy. Changes to memory, tools and execution routines
face similar questions. Do they preserve earlier skills, and do they help on new tasks, or do they only fix failures the system has
already seen? Repeatedly revising against the same development set risks adaptive overfitting, so retained changes should be gated on non-regression and generalisation to unseen tasks. Studies of recursive self-improvement often ask whether performance improved by the end. Few compare methods that start
from the same system and get the same budget. Research on changing the robot’s body or tools is still limited. Most studies design
for one task, with little evidence that the design can be reused on later tasks. Designs tested in simulation also need to be tried
on real hardware. Finally, studies often give too little detail about costs and human help to tell what it takes to keep improving.
For recursive self-improvement to help in practice, it needs to deliver real gains in a reasonable time, at low cost and with little
human engagement.

\paragraph{EVA pathway.} In the pathway of \Cref{sec:vrea}, self-improvement is what keeps the agent moving from the virtual environment to the real one. Environment improvement (\Cref{sec:si-environment}) revises the virtual environment from the agent’s own failures, bringing it closer to the real one; foundation model and harness improvement (\Cref{sec:si-foundation,sec:si-harness}) keep the memories, skills, tools and policies that the agent carries into the real world; and embodiment improvement (\Cref{sec:si-embodiment}) adapts the robot’s body and tools to its tasks. As self-improvement advances, the pathway advances with it: environments revised from the agent’s experience close the gap to the real world where it matters for the agent, and kept changes let each attempt start from what earlier attempts achieved. A static view of the agent and its environment, as taken in \Cref{sec:sim,sec:embodied-agent}, cannot show this progress: it appears only when the system uses feedback from its own tasks to change itself, and it grows faster when the way the system improves is itself improved. Better self-improvement thus lets the agent reach the real world with more of what it has learned, keep improving once it is there, and turn progress in one environment into progress in the other.
\section{Discussion and Open Problems}
\label{sec:discussion}

The pathway that EVA proposes (\Cref{sec:vrea}) frames this discussion: the agent first improves itself in the virtual environment $\mathcal E_V$
while the overlap of $\mathcal E_V$ with the real environment $\mathcal E_R$, the tasks and conditions that $\mathcal E_V$ covers and
predicts, grows, and extends its self-improvement to $\mathcal E_R$ as evidence accumulates that its capabilities support the intended physical tasks. The pathway is a proposal. No general threshold says when that evidence suffices, and
no reviewed system yet keeps a revised world because later real tasks improved, or runs both Phase I (offline environment scaling \& embodied agent self-improvement) and Phase II (online real-to-sim scaling)
(\Cref{sec:si-environment}); the test it most needs is whether real performance rises with measured overlap at a matched budget of
real-robot time.

The pathway raises four problems that the parts of \Crefrange{sec:sim}{sec:selfimprove} share rather than solve
once the agent acts on real robots: latency relative to the rate of physical control (\Cref{sec:latency}), failure, since
agents still fail at precise, contact-rich and fast control, and benchmark scores can hide it (\Cref{sec:failure}), the cost of
running and training these systems (\Cref{sec:cost}), and physical safety and information security, which the reviewed work has
largely overlooked (\Cref{sec:safety}).

\subsection{Latency}
\label{sec:latency}

A general-purpose model takes seconds per decision, and an agentic loop multiplies this by the number of calls, while contact,
slip and moving objects change within milliseconds. Simulators that pause while the model deliberates hide this
gap~\citep{wang2023prompt,chu2026openrua,zhang2026unexpected}, so their success rates measure what the model decides, not whether it
decides in time. When the world keeps moving, the gap appears: in RT-Safe, task completion is about the same with the world paused
and running (91.3\% and 94.1\%), but collisions rise 12.3-fold and safe success, finishing without a safety event, falls from
19.8\% to 0.7\%~\citep{xu2026rt}; in Embodied First Exam, GPT-6 Astra solves 23 of 54 dynamic tasks with the world paused and
17 in real time~\citep{wang2026embodied}. A decision of $N$ calls takes roughly $T \approx N\,(t_{\mathrm{model}} + t_{\mathrm{tool}} + t_{\mathrm{net}})$,
the model-call, tool-call and network time of each call; in AESOP, GPT-4 with chain-of-thought reasoning takes
18.88\,s on average to answer one anomaly query~\citep{sinha2024real}. How the delay bites depends on the form of the command: in
Agent as System 2 it is the interval between slow decisions relative to the control period below them (\Cref{sec:slowfast}); in Code
as Policy it is paid once per program and again at every replanning (\Cref{sec:code}); in LLM as Policy every command waits for
the model (\Cref{sec:llmpolicy}).

 \begin{figure}[t]
    \centering
    \includegraphics[width=\linewidth]{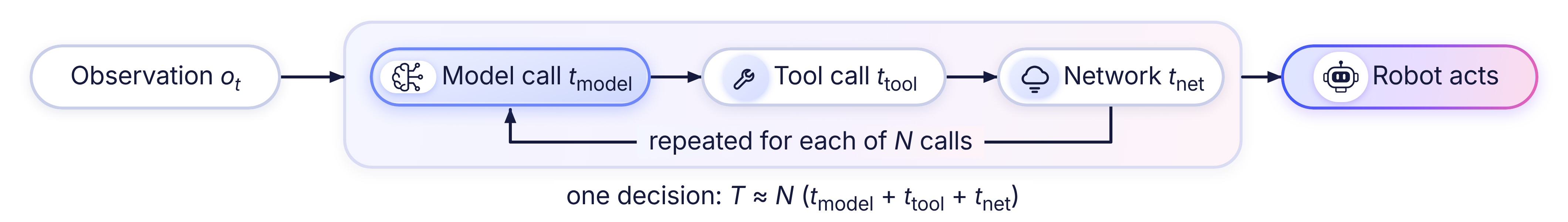}
    \caption{\textbf{Where the time of one decision goes.} A decision takes $N$ calls, each paying the model-call time $t_{\mathrm{model}}$, the tool-call time $t_{\mathrm{tool}}$ and the network latency $t_{\mathrm{net}}$, so the robot acts on an observation $o_t$ taken $T \approx N\,(t_{\mathrm{model}} + t_{\mathrm{tool}} + t_{\mathrm{net}})$ earlier.}
    \label{fig:latency}
\end{figure}

The robot therefore acts on stale observations: a command computed from an image taken seconds earlier rests on assumptions about
the scene that may no longer hold. Control has long treated such delays explicitly, for example by adding the actions still in transit
to the state of a learned dynamics model~\citep{chen2020delay}. What the robot does while the model reasons, whether it holds its pose,
continues the last command or retreats to a safe state, is rarely specified and must be enforced by the layers below the model
(\Cref{sec:safety}). Five strategies reduce the effect. \textbf{1) Fewer calls} reduce round trips, for example by composing several
actions in one code cell~\citep{si2026fewer}. \textbf{2) Rate separation} lets a slow model set targets while a fast controller tracks
them, as in See, Point, Fly~\citep{hu2025see} and in the fast on-board detector that triggers the slow reasoner
of \citet{sinha2024real}. \textbf{3) Amortisation} turns one expensive decision into many cheap executions: a program, keypoint
constraints re-optimised during execution~\citep{huang2024rekep}, a monitoring program~\citep{zhou2024code} or a policy distilled from
the model's experience (\Cref{sec:learning-demos}); deliberation that overlaps with execution and action chunks adapted to the task phase
remain untested. \textbf{4) Fallbacks} kept feasible while the model reasons let the robot recover without waiting for
it~\citep{sinha2024real,ganai2025real}. \textbf{5) Smaller on-board models} trade capability for speed: LocalNav distils cloud models into
one that runs on the robot's embedded hardware, removing the network round trip~\citep{baumann2026localnav}, and an on-board model
keeps camera images on the robot (\Cref{sec:safety}). Evaluations should report whether the simulator pauses, the latency per decision
and per episode, the age of the observation behind each action, from its capture to the action taking effect, which unlike model
latency also counts perception, communication and queuing, and results under real-time execution, in which the world evolves
throughout inference, as in RT-Safe and Embodied First Exam~\citep{xu2026rt,wang2026embodied}.

\subsection{Failure}
\label{sec:failure}

\textbf{Robot-use agents now lead the main benchmarks}. On the RoboDojo simulation leaderboard of 1 December 2026, ranked by success without the Open dimension, the ten highest entries all use an
agent, led by HexaFuture's HexaShell at 87.52\% success, while the best VLA and WAM reach 43.44\% and 40.91\%~\citep{robodojo2026leaderboard}; the leaderboard lists Anthropic's own result for Fable 5.2 as a score of 64.90,
without a success rate~\citep{robodojo2026leaderboard,anthropic2026fable}. On RoboCasa365, the agent HexaShell leads with an overall score of 85.6, against 64.4 for the best
WAM~\citep{robocasa2026leaderboard}. Does this mean that agents beat these policies in every respect? The answer is no. In the evaluation by
\citet{zhang2026unexpected}, Astra ``performs poorly on tasks that require precision, dynamic control, or complex bimanual
coordination'': it reaches 4.00\% on the precision tasks against 16.75\% for DM0.5, the highest-ranked public policy, knocks over a tower while reaching for
the next block, and misses a moving object on a conveyor. Other studies report the same kinds of
failure. Decisions come too late, because the scene changes while the model thinks~\citep{sensenova2026has}, and the model often
says that a task is done when it is not~\citep{espinola2026reasoning}.

\textbf{High benchmark scores can hide these failures}, for three reasons. First, the benchmarks favour tasks that need reasoning. RoboDojo,
for example, asks the robot to arrange digits into the largest number, or to remember which lid covers which coloured
block~\citep{chen2026robodojo}. Fast reactive policies lack strong high-level reasoning and do poorly on such tasks: on the Open and
Generalization tasks of RoboDojo, \citet{zhang2026unexpected} report 31.00\% and 30.50\% for Astra, against 2.08\% and 10.95\% for
DM0.5. Second, many simulators stop while the model thinks, which hides the cost of its latency: on
RoboDojo, 8--76\,s of robot motion take 55 minutes of computation on average~\citep{sensenova2026has}. Third, agents can receive information that a VLA cannot use. When the RoboDojo team checked the code of its top agent entries,
it found task-specific solution guides, checker tolerances, fixed target coordinates and scoring tables added to the prompt, and it
now marks such entries as not following its rules for agents; other agent systems read the true poses of objects from the
simulator, which is not possible on real hardware~\citep{chu2026openrua}.

The real causes of failure are the ones we need to face. \textbf{Compared with a VLA, the main weakness of an embodied agent is latency}. A
WAM decides in about 0.2\,s~\citep{wang2026openwam}, while one call to a large model can take tens of
seconds~\citep{team2026systematically}, so the agent cannot handle tasks that need a fast reaction or precise alignment: plug
insertion, for example, succeeds in none of 11 trials~\citep{sensenova2026has}. \textbf{Beyond the agent, part of the difficulty lies in
the robot's body.} A person combines sight, hearing, touch, force and the sense of their own limbs, while most robots in these
studies have only cameras and at most a tactile sensor, so the model observes a more partial state of the world. Astra, for
example, counts a mug as hung because the mug and the peg overlap in the image~\citep{sensenova2026has}. The control is also
coarse: the robot moves on average only 64--72\% of the distance it is told to move~\citep{espinola2026reasoning}.

\subsection{Cost}
\label{sec:cost}

\textbf{Cost is one of the biggest issues for embodied agents driven by large language models.} A VLA or WAM deployed on the robot is paid for mostly once: after training, it runs on an inexpensive onboard computer, as a
distilled navigation model does on a Jetson AGX Orin~\citep{baumann2026localnav}. An agent built on a large language model, by
contrast, pays each time it acts, and its bill grows with the price of each call and with the number of calls that a task needs.

\textbf{For a closed model, tokens are the largest cost.} Most of them pay for prefilling rather than thinking, because images and
history are re-sent at each call: AA-Robotics reports that 70\% of spending goes on
context~\citep{artificialanalysis2026robotics}, and in one evaluation of GPT-6 Astra about
98\% of the tokens are cached input~\citep{sensenova2026has}. \textbf{For an open model, the cost of the hardware
that serves it takes the place of the token bill.} ZAP, a 27B model served on one B300 GPU at around \$7 per hour, matches GPT-6 Astra in the same harness on real tasks,
with 96.7\% success for both, at \$0.29 per task against \$4.87~\citep{pikus2026pretraining}, and the 27B model trained with PEAR
runs on a single RTX 5090~\citep{tang2026training}.

\textbf{Embodied agents act over many turns, and the number of calls multiplies this price}. An agent observes, reasons, chooses and acts at each
step, so a long trajectory costs more, and a failed attempt and retry will double it. In OpenRUA, the model with the lowest token prices
fails most and runs longest, and costs \$93.3 per success, the total spent on all attempts divided by the number of successes,
against \$11.2 for the best model~\citep{chu2026openrua}. How often the model is called matters as much: on RoboDojo, reviewing
every fourth action of a VLA instead of every action cuts token use from 7.66M to 2.80M while raising
success~\citep{sensenova2026has}.

\textbf{Open-weight models pay the bill in training}, a cost paid once but large for agentic reinforcement learning, in which each sample is a long
multi-turn trajectory and each starting state is sampled several times. PEAR trains its 27B model with 128 groups of 8 rollouts per
step for about 50,000 H100-equivalent GPU-hours~\citep{tang2026training}, and CaPtain-4B takes about 1,290 H100-hours of RL on 16
GPUs~\citep{fan2026captain}. Pretraining a WAM is costly too, with 128 H200 GPUs for about seven days in
OpenWAM~\citep{wang2026openwam}, so agentic RL is not necessarily more expensive to train; what sets language-model agents apart is
the cost of every run.

\textbf{On a real robot, time is also a cost.} A slower agent keeps both the robot and the person watching it busy for longer. Resetting the scene takes the same time whatever controls the robot, but a person must stay with the robot while it acts, so supervision grows with the length of each run. In the same harness, ZAP finishes a real task in 146\,s on average and GPT-6 Astra in 397\,s~\citep{pikus2026pretraining}, and on a humanoid robot a single manipulation trial with GPT-6 Astra takes over an hour, about 90\% of it spent waiting for the model~\citep{sensenova2026has}. Measured in robot and human hours, a slow agent can cost more than its token bill suggests.

\subsection{Physical Safety and Information Security}
\label{sec:safety}

\textbf{Success is not the only metric for evaluating an agent, and reliability is even more important in real-world applications.} A mistake in software can usually be undone, but a robot's mistake happens in the physical world and is often irreversible: it can hurt people, break objects or damage the robot itself. Existing work shows the risk is large: on RoboRisk, whose 92 simulated and 13 physical tasks cover refusal, plan repair, hazards during execution, injected instructions and memory, six frontier models that control a robot arm cause harm in 9\% to 77\% of tasks~\citep{zhu2026leash}. A success rate does not show this risk, because it ignores the damage done on the way to the goal, and the model judges its own results poorly, so it does not correct an unsafe step that it has not noticed. When a system changes itself (\Cref{sec:selfimprove}), the objectives it writes can also be gamed~\citep{ma2023eureka}. Yet the reviewed work pays little attention to safety: only five of the 41 works compared for Agent as System 2 are safety guardrails (\Cref{tab:cmp-4-3}).

Existing work covers parts of the problem. \textbf{Runtime guardrails} sit between the model and the robot and block actions or
plans that break a rule~\citep{yang2023plug,ravichandran2025safety}. LEASH places three checks in the harness: one for the task,
one for each action and one for each irreversible operation. Each check uses the current camera image and the same safety policy
as the planner. With LEASH, four of six models cause no observed harm on the physical robot, 84.6\% of tasks are still completed
safely, and the run time grows by less than 2\%~\citep{zhu2026leash}. \textbf{Constraints given to a solver} keep the hard limits out of the
model's output; the model supplies only soft preferences~\citep{wu2026embodiment}. \textbf{Semantic-safety benchmarks} test
whether a model recognises a hazard or refuses an unsafe instruction. SafeAgentBench has 750 tasks with 10 kinds of hazard, and
even the most careful baseline rejects only 10\% of detailed hazardous tasks~\citep{yin2024safeagentbench}. RoboRisk goes further and measures harm while the robot acts, in simulation and on a physical arm~\citep{zhu2026leash}. Another benchmark
checks whether the model takes a safety measure before a risky step, and finds that safety-aware reasoning often lowers task
completion~\citep{lu2025bench}. \textbf{Evaluation of physical damage} is rare. OopsieVerse converts contact forces, temperature
changes and spilled liquids into mechanical, thermal and fluid damage in two simulators, and uses this signal to collect safer
demonstrations, to train policies that take damage into account and to benchmark VLA policies~\citep{balaji2026oopsieverse}.
RT-Safe lets the world keep changing while the model thinks~\citep{xu2026rt}. \textbf{Safety during self-improvement} appears in
ENPIRE, which keeps its safety limits fixed and counts every breach as a failure~\citep{xiao2026enpire}. In short, most of this
work tests whether a model recognises or refuses a hazard, usually in simulation or on video. Little work measures the physical
damage done while the robot acts.

\textbf{Information security and privacy} receive even less attention, and three problems remain open. First, \textbf{prompt injection}: the model reads the
camera image, so a sign or an arranged object can act as a hidden instruction. Jailbreak prompts
already make robots controlled by language models carry out harmful actions~\citep{robey2024jailbreaking}, and misleading signs
weaken every guard tested in EgoSafetyBench~\citep{panpatil2026egosafetybench}. Second, \textbf{memory poisoning}: self-improvement keeps skills and memories for later reuse, so an attacker can plant a harmful
entry that looks normal and is retrieved only when a chosen trigger appears in a later task. AgentPoison shows this for
language-model agents, with an attack success rate above 80\% while normal tasks are barely affected~\citep{chen2024agentpoison}.
The libraries of \Cref{sec:si-harness} have not been checked for this. Third, \textbf{privacy}: a model in the cloud receives images of people, rooms and documents with every decision, and the reviewed
work does not ask whether filtering on the robot or a smaller local model (\Cref{sec:latency}) could keep these images on the
robot. These problems matter most for the systems reviewed here. A frozen model that is called remotely cannot be made safer by
further training, and the part that developers control, the harness, is exactly the part that self-improvement keeps rewriting.

Possible remedies are only beginning to appear, such as limits and monitors that sit outside the model, rules that are checked
before they are used, and evaluations that report damage and violated constraints as well as success.

\section{Conclusion}
\label{sec:conclusion}

In this paper, we argue that capabilities developed in virtual settings can progressively extend to the physical world, and propose
EVA as a conceptual framework for this transition. The growing general capabilities of foundation models, together with an expanding
ecosystem of software and robot interfaces, open a path for embodied agents to become increasingly capable through interaction with
their environments. We identify RSI as a route towards realising EVA: experience can improve both the
system's capabilities and the processes through which it makes subsequent improvements.

Environments and embodied agents form the two central components of this perspective. Environments provide the foundation for
embodied-agent execution: they determine what can be practised, what experience can be collected, and how progress is measured. Constructed
virtual environments support repeatable and varied interaction, while task generation and curriculum design organise
practice around the capabilities to be acquired. Reconstruction and calibration against real observations bring these practice
conditions closer to physical reality. For embodied agents, we examine both training and deployment. During training, foundation
models can generate and curate demonstrations, guide exploration, design rewards and manage training pipelines. During deployment,
they can direct learned policies, write executable programs, or issue concrete action commands, using feedback to monitor progress and
recover from failures. Self-improvement connects these activities across successive cycles. Execution reveals missing skills,
ineffective strategies or mismatches between simulation and reality; these outcomes can guide targeted practice, revisions to the
virtual environment, and updates to memory, executable skills and policies. Useful changes are retained and reused, allowing later
cycles to build on earlier experience. Within EVA, we envision this joint development making environments increasingly useful for
learning and more representative of reality, while embodied capabilities grow through interaction and extend progressively into the
physical world.

Current work nevertheless faces several limitations. Success judgements and generated learning signals can be unreliable, while
improvements on practised tasks do not necessarily generalise to new tasks or transfer from simulation to reality. Physical deployment
also faces constraints in latency, safety, cost and dependence on human assistance.

Addressing these limitations requires more reliable evaluation, stronger evidence of generalisation and transfer, and efficient, safe
mechanisms for sustained interaction with the physical world. Looking ahead, stronger foundation models trained on larger and more diverse embodied configurations and modalities, including proprioceptive, tactile and
other sensory signals, could develop more generalisable physical capabilities. Faster inference hardware, improved acceleration
algorithms and more efficient general-purpose models could further reduce decision latency and deployment costs. More capable systems
could also require less human involvement in task preparation, supervision and recovery. Meanwhile, more comprehensive evaluation
protocols can reveal remaining weaknesses and guide optimisation. Together, these advances could alleviate many current limitations
and support sustained self-improvement along the pathway envisioned by EVA.

\renewcommand{\bibfont}{\footnotesize}\setlength{\bibsep}{1.2pt plus 0.3pt}

\clearpage
\newpage
\phantomsection
\addcontentsline{toc}{section}{References}
\bibliographystyle{assets/plainnat}
\bibliography{references}

\clearpage

\end{document}